\documentclass{article}

\usepackage{microtype}
\usepackage{graphicx}
\usepackage{booktabs} 
\usepackage{enumitem}

\usepackage{hyperref}

\usepackage[accepted]{icml2024}

\usepackage{amsmath}
\usepackage{amssymb}
\usepackage{mathtools}
\usepackage{amsthm}
\usepackage{wrapfig}
\usepackage{multirow}

\usepackage{url}
\usepackage{mathtools}
\usepackage{makecell}

\usepackage{caption}
\usepackage{subcaption}
\usepackage[T1]{fontenc}
\usepackage{lscape} 

\usepackage[capitalize,noabbrev]{cleveref}

\usepackage{tikz,tikz-cd}
\usetikzlibrary{arrows,positioning,automata} 

\theoremstyle{plain}

\theoremstyle{definition}

\theoremstyle{remark}

\usepackage[textsize=tiny]{todonotes}

\definecolor{darkgreen}{rgb}{0., 0.5, 0.}
\newcommand{\problem}{forgetting of pre-trained capabilities}
\newcommand{\abbrv}{FPC}
\newcommand{\problemUpper}{Forgetting of pre-trained capabilities}
\newcommand{\roboticsequence}{\texttt{RoboticSequence}}

\newcommand{\diff}[1]{\textcolor{blue}{#1}}
\newcommand{\newdiff}[1]{\textcolor{blue}{#1}}

\newcommand{\camera}[1]{#1}

\renewcommand{\diff}[1]{#1}
\renewcommand{\newdiff}[1]{#1}

\newcommand{\envshift}{state coverage gap}
\newcommand{\envshiftUpper}{State coverage gap}
\newcommand{\imitationgap}{imperfect cloning gap}
\newcommand{\imitationgapUpper}{Imperfect cloning gap}

\DeclarePairedDelimiterX{\infdivx}[2]{(}{)}{%
  #1\;\delimsize\|\;#2%
}

\icmltitlerunning{Fine-tuning Reinforcement Learning Models is Secretly a Forgetting Mitigation Problem}

\begin{document}

\twocolumn[
\icmltitle{Fine-tuning Reinforcement Learning Models is Secretly a Forgetting Mitigation Problem}



\icmlsetsymbol{equal}{*}

\begin{icmlauthorlist}
\icmlauthor{Maciej Wołczyk}{equal,ideas}
\icmlauthor{Bartłomiej Cupiał}{equal,ideas,uw}
\icmlauthor{Mateusz Ostaszewski}{wut}
\icmlauthor{Michał Bortkiewicz}{wut}
\icmlauthor{Michał Zając}{ju}
\icmlauthor{Razvan Pascanu}{deepmind}
\icmlauthor{Łukasz Kuciński}{ideas,uw,pan}
\icmlauthor{Piotr Miłoś}{ideas,uw,pan,deepsense}
\end{icmlauthorlist}

\icmlaffiliation{ideas}{IDEAS NCBR}
\icmlaffiliation{uw}{University of Warsaw}
\icmlaffiliation{wut}{Warsaw University of Technology} 
\icmlaffiliation{ju}{Jagiellonian University} 
\icmlaffiliation{deepmind}{Google DeepMind} 
\icmlaffiliation{pan}{Institute of Mathematics, Polish Academy of Sciences}
\icmlaffiliation{deepsense}{deepsense.ai}

\icmlcorrespondingauthor{Maciej Wołczyk}{maciej.wolczyk@gmail.com}

\icmlkeywords{Fine-tuning Reinforcement Learning Models is Secretly a Forgetting Mitigation Problem}

\vskip 0.3in
]



\printAffiliationsAndNotice{\icmlEqualContribution} 

\begin{abstract}
    Fine-tuning is a widespread technique that allows practitioners to transfer pre-trained capabilities, as recently showcased by the successful applications of foundation models. However, fine-tuning reinforcement learning (RL) models remains a challenge. This work conceptualizes one specific cause of poor transfer, accentuated in the RL setting by the interplay between actions and observations: \emph{\problem{}}. Namely, a model deteriorates on the state subspace of the downstream task not visited in the initial phase of fine-tuning, on which the model behaved well due to pre-training. This way, we lose the anticipated transfer benefits. We identify conditions when this problem occurs, showing that it is common and, in many cases, catastrophic. Through a detailed empirical analysis of the challenging NetHack and Montezuma's Revenge environments, we show that standard knowledge retention techniques mitigate the problem and thus allow us to take full advantage of the pre-trained capabilities. In particular, in NetHack, we achieve a new state-of-the-art for neural models, improving the previous best score from $5$K to over $10$K points in the Human Monk scenario.
\end{abstract}

\section{Introduction}
Fine-tuning neural networks is a widespread technique in deep learning for knowledge transfer between datasets \citep{yosinski2014transferable,girshick2014rich}. Its effectiveness has recently been showcased by spectacular successes in the deployment of foundation models in downstream tasks, including natural language processing \citep{chung2022scaling}, computer vision \citep{sandler2022fine}, automatic speech recognition~\citep{zhang2022bigssl}, and cheminformatics~\citep{chithrananda2020chemberta}. These successes are predominantly evident in supervised and self-supervised learning domains. However, achievements of comparable significance have not yet fully found their way to reinforcement learning (RL) \citep{wulfmeier2023foundations}.

In this study, we explore the challenges and solutions for effectively transferring knowledge from a pre-trained model to a downstream task in the context of RL fine-tuning. We find that the interplay between actions and observations in RL leads to a changing visitation of states during the fine-tuning process with catastrophic consequences. Intuitively, the agent may lose pre-trained abilities in parts of the downstream task not covered in early fine-tuning, diminishing the expected transfer benefits.

\begin{figure*}
    \centering
    \includegraphics[width=0.95\textwidth]{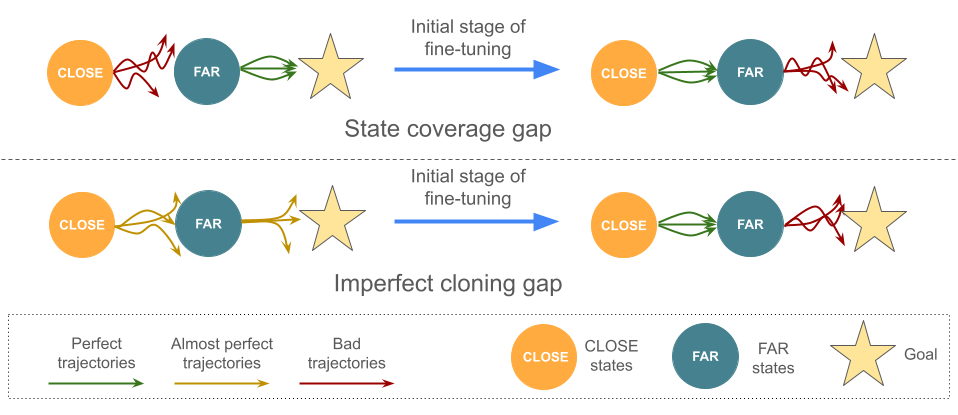}
    \caption{\textbf{\problemUpper{}}. For illustration, we partition the states of the downstream task into \textsc{Close} and \textsc{Far}, depending on the distance from the starting state; the agent must master \textsc{Far} to reach the goal.
    In the \envshift{} (top), the pre-trained policy performs perfectly on \textsc{Far} but is suboptimal on \textsc{Close}.  During the initial stage of fine-tuning, while mastering \textsc{Close}, the policy deteriorates, often catastrophically, on \textsc{Far}. In \imitationgap{} (bottom), the pre-trained policy is decent both on \textsc{Close} and \textsc{Far}; however, due to compounding errors in the initial stages of fine-tuning, the agent rarely visits \textsc{Far}, and the policy deteriorates on this part. In both cases, the deteriorated policy on \textsc{Far} is hard to recover and thus necessitates long training to solve the whole task.}
    \label{fig:intro}
\end{figure*}

We refer to this issue as \emph{\problem{}} (\abbrv{}).  We identify two important instances of \abbrv{}: \emph{\envshift{}} and \emph{\imitationgap{}}, illustrated in Figure~\ref{fig:intro} and defined in Section \ref{sec:preliminaries}. We show empirically that the problem is severe, as these instances are often encountered in practice, leading to poor transfer to downstream tasks. These findings are in contrast to the conventional wisdom that emerged from the supervised learning setting, where the data distribution is i.i.d. and forgetting is not a factor if one cares only about the performance on the downstream task; see \citep[Sec 3.5]{wulfmeier2023foundations} and \citep{radford2018improving,Devlin2019BERTPO,dosovitskiy2020image}.



Finally, we show that phrasing \emph{\envshift{}} and \emph{\imitationgap{}} as instances of \emph{forgetting} is meaningful as typical retention techniques \citep{kirkpatrick2017overcoming,rebuffi2017icarl,wolczyk2021continual} can alleviate these problems. We demonstrate this effect on NetHack, Montezuma's Revenge, and tasks built out of Meta-World, an environment simulating tasks for robotic arms. Applying knowledge retention enhances the fine-tuning performance on all environments and leads to a $2$x improvement in the state-of-art results for neural models on NetHack. Further analysis shows that \problem{} is at the heart of the problem, as vanilla fine-tuning rapidly forgets how to perform in parts of the state space not encountered immediately in the downstream task. 

As such, the main recommendation of our work is that methods targeting catastrophic forgetting should be routinely used in transfer RL scenarios. 
In summary, our contributions are as follows: 
\begin{itemize} 
    \item We pinpoint \problem{} as a critical problem limiting transfer from pre-trained models in RL and provide a conceptualization of this phenomenon, along with its two common instances: {\envshift{}} and {\imitationgap{}}.

    \item We propose knowledge retention techniques as a tool that mitigates \abbrv{} and allows us to transfer from the pre-trained model efficiently.

    \item We thoroughly examine our approach on Nethack, Montezuma's Revenge, and sequential robotic tasks, improving the state-of-the-art for neural models on NetHack by $2$x.
\end{itemize}

 \section{\problemUpper{}} \label{sec:preliminaries}
 
\begin{figure*}
    \centering
    \hfill 
    \begin{subfigure}[t]{0.38\textwidth}
        \centering
        \includegraphics[width=\textwidth]{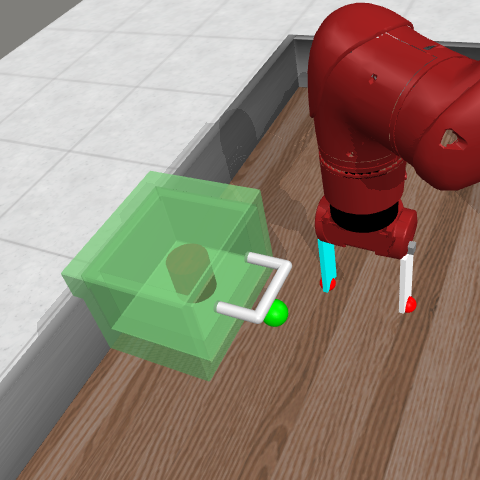}
    \end{subfigure}
    \hfill 
    \begin{subfigure}[t]{0.54\textwidth}
        \centering
        \includegraphics[width=\textwidth]{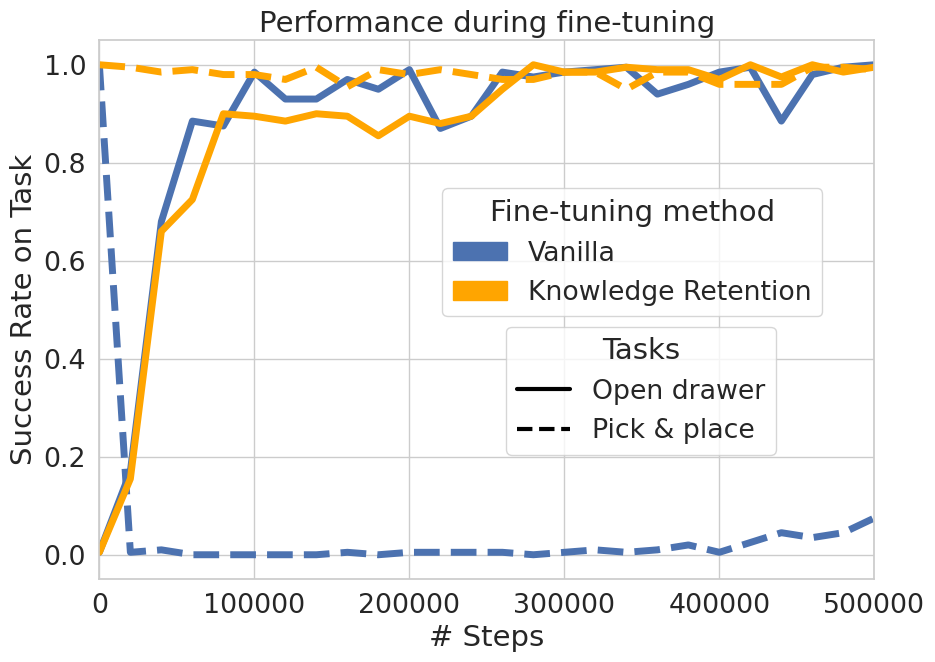}
    \end{subfigure}
    \hfill 
    
    \caption{Example of \textbf{\envshift{}}. (Left) We assume that a pre-trained model is able to pick and place objects (e.g., the cylinder). However, it does not know how to open drawers. Consider a new task in which the agent needs first to open the drawer (\textsc{Close} states) and then pick and place the object (\textsc{Far} states). (Right) During fine-tuning, the model rapidly forgets how to manipulate objects before learning to open the drawer and struggles to reacquire this skill (dashed blue line). Knowledge retention techniques alleviate this issue (dashed orange line). At the same time, in both cases, the model learns how to open the drawer (solid lines). 
    }
    \label{fig:state_coverage_gap}
\end{figure*}

To illustrate the \problem{}, let us consider a Markov Decision Problem (MDP) where the state space can be approximately split into two sets: \textsc{Close} and \textsc{Far}, see Figure~\ref{fig:intro}. 
The states in \textsc{Close} are easily reachable from the starting state and the agent frequently visits them. The states in \textsc{Far} are reachable only by going through \textsc{Close}; hence, they are infrequently visited as they can be reached only once some learning on \textsc{Close} happens. For example, an agent learning to play a video game might only see the first level of the game (\textsc{Close}) at the start of the training before it learns how to get to the subsequent levels (\textsc{Far}).

\textbf{\problemUpper{}} happens when a model performing well on \textsc{Far} loses this ability due to interference in the function approximator when training on \textsc{Close}. We believe this problem has not yet been studied thoroughly and has a major significance for transfer RL since it is commonly present in standard RL settings and often leads to substantial performance deterioration. The subsequent experimental sections provide multiple examples of its occurrence, and in Appendix~\ref{app:toy_mdp}, we show that it can be observed already in simple two-state MDPs as well as gridworlds. To facilitate further study of this problem, we highlight two specific scenarios where \problem{} occurs: the \envshift{} and \imitationgap{}.

In \textbf{\envshift{}}, we consider a pre-trained agent that is performing well mostly on \textsc{Far}
and does not know how to behave on \textsc{Close}. 
However, when fine-tuned on \textsc{Close}, its behavior on \textsc{Far} will deteriorate considerably due to  
forgetting\footnote{\camera{For a more thorough discussion on the nature of interference in RL we refer the reader to~\citet{schaul2019ray}}}
and will have to be re-acquired. 
This setting is representative of common transfer RL scenarios 
\citep{parisotto2015actor,rusu2016progressive,rusu2022probing}, 
see also the top row of Figure \ref{fig:intro} and Figure \ref{fig:state_coverage_gap} for illustration. 

The \textbf{\imitationgap{}} occurs when the pre-trained agent is a perturbed version of an agent that is effective in the current environment. Even if the difference is small, this discrepancy can lead to a substantial imbalance with the agent visiting states in \textsc{Close} much more often than \textsc{Far}. While trying to correct the slightly suboptimal policy on \textsc{Close}, the policy on \textsc{Far} can get worse due to 
forgetting,
see the depiction in Figure \ref{fig:intro}.
Such scenarios frequently arise due to slight changes in the reward structure 
between pre-training and fine-tuning or approximation errors when cloning an expert policy, and, more generally, when using models pre-trained on offline static datasets~\citep{nair2020awac,baker2022video,zheng2023adaptive}.

\textbf{Knowledge retention} 
In this paper, we argue that to benefit from fine-tuning pre-trained RL models, we need to mitigate \abbrv{}. 
To this end, we consider the following popular methods for knowledge retention: 
Elastic Weight Consolidation (EWC), replay by behavioral cloning (BC), kickstarting (KS), and episodic memory (EM). \textbf{EWC} is a regularization-based approach that applies a penalty on parameter changes by introducing an auxiliary loss: $\mathcal L_{aux}(\theta) = \sum_i F^i (\theta_{\text{*}}^i  - \theta^i )^2$,  where $\theta$ (resp $\theta_{\text{*}}$) are the weights of the current (resp. pre-trained) model, and $F$ is the diagonal of the Fisher matrix. We also use \textbf{behavioral cloning}, an efficient replay-based approach \citep{rebuffi2017icarl,wolczykdisentangling}. We implement BC in the following way. Before the training, we gather a subset of states $\mathcal{S}_{BC}$ on which the pre-trained model $\pi_{*}$ was trained, and we construct a buffer $\mathcal{B}_{BC}:= \{(s, \pi_{*}(s)): s\in \mathcal{S}_{BC}\}$. For the fine-tuning phase, we initialize the policy with $\theta_{*}$ and we apply an auxiliary loss of the form $\mathcal L_{BC}(\theta) = \mathbb{E}_{s \sim \mathcal{B}_{BC}} [ D_{KL}\infdivx{\pi_*(s)}{\pi_\theta(s)} ]$ alongside the RL objective. \textbf{Kickstarting} applies KL of a similar form, but the expectation is over data sampled by the current policy, i.e., $\mathcal L_{KS}(\theta) = \mathbb{E}_{s \sim \pi_\theta} [ D_{KL}\infdivx{\pi_*(s)}{\pi_\theta(s)} ]$. For \textbf{episodic memory}, we can easily use it with off-policy methods by simply keeping the examples from the pre-trained task in the replay buffer when training on the new task. Following previous best practices~\cite {wolczykdisentangling}, we do not apply knowledge retention to the parameters of the critic. See Appendix~\ref{app:cl_methods} for more details. 

\textbf{Relation to continual reinforcement learning}
The main focus of this paper is the efficient fine-tuning of a pre-trained RL agent.
We consider forgetting only as far as it impacts the transfer and we are \emph{solely} interested in the performance on the downstream task, disregarding the performance of the pre-trained tasks.  
This is in contrast to continual reinforcement learning \cite{khetarpal2022towards, wolczyk2021continual,kirkpatrick2017overcoming}, where one of the goals is to retain the performance on the pre-trained tasks.
Interestingly, we show that contrary to prior knowledge~\cite{wulfmeier2023foundations}, forgetting might severely hinder the transfer capabilities in standard transfer RL settings with a stationary downstream task.



\section{Experimental setup}\label{sec:exp_setup}

We perform experiments on three environments: NetHack, Montezuma's Revenge, and \roboticsequence{}. 
Below, we describe them in detail and show concrete instances of concepts from Section~\ref{sec:preliminaries} such as 
pre-trained policy $\pi_*$ or \textsc{Far} and \textsc{Close} sets.
In each environment, we run vanilla fine-tuning and training from scratch as baselines, and we test fine-tuning with different knowledge retention methods (e.g., Fine-tuning + BC). 

\textbf{NetHack Learning Environment}~\citep{kuttler2020nethack} is a complex game consisting of procedurally generated multi-level dungeons. Since their layouts are randomly generated in each run, the player has to learn a general strategy rather than memorize solutions.
NetHack is stochastic and requires mastering diverse skills, such as maze navigation, searching for food, fighting, and casting spells. It has been a popular video game for decades that recently has become a challenging testbed at the forefront of RL research~\cite{hambro2022insights,piterbarg2023nethack,klissarov2023motif}.
Due to computational constraints, we focus solely on a single setting in our experiments, i.e., Human Monk. \camera{The code is available at~\url{https://github.com/BartekCupial/finetuning-RL-as-CL}.} 


We take the current state-of-the-art neural model~\citep{tuyls2023scaling} as our pre-trained policy $\pi_*$. It was trained using behavioral cloning on $115$B environment transitions sampled from AutoAscend, a rule-based agent that is currently the best-performing bot. The policy $\pi_*$ scores over 5K points.

\begin{figure*}[t]
    \centering
    \includegraphics[width=0.95\linewidth]{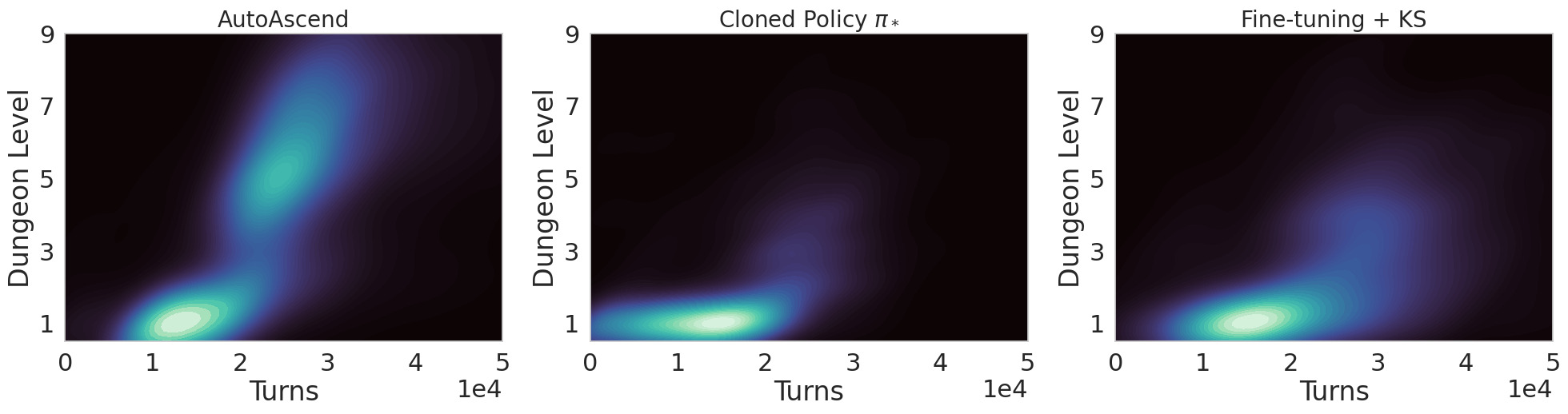}
    \caption{ 
    Density plots showing maximum dungeon level achieved compared to the total number of turns (units of in-game time)  for expert AutoAscend (left), pre-trained policy $\pi_*$ (center), and fine-tuning + KS (right) Brighter colors indicate higher visitation density. Level visitation of $\pi_*$ differs significantly from the level visitation of the AutoAscend expert. This is an example of \imitationgap{} as the agent will not see further levels at the start of fine-tuning. The knowledge retention-based method manages to perform well and explore different parts of the state space.
    }
    \label{fig:nethack_level_visitation}
\end{figure*}

Since the policy $\pi_*$ rarely leaves the first level of the game (see Figure \ref{fig:nethack_level_visitation}), 
we conceptualize \textsc{Close} as the set of states corresponding to this initial level. 
Accordingly, \textsc{Far} represents states from subsequent levels.
During fine-tuning, we use asynchronous PPO (APPO)~\citep{Petrenko2020SampleFE}. More technical details, including the neural network architecture, can be found in Appendix~~\ref{app:nethack}. 

%



%




\textbf{Montezuma's Revenge} is a popular video game that requires the player to advance through a sequence of rooms filled with traps and enemies while collecting treasures and keys~\cite{bellemare2013arcade}. 
The environment has sparse rewards and is a well-known exploration challenge in RL.



We pre-train a policy $\pi_*$ on a part of the environment that includes only rooms from a certain room onward (see the layout of the game in Figure \ref{fig:montezuma_layout} in Appendix \ref{app:montezuma}). In particular, in the main text, we start pre-training from Room 7 and we verify other room choices in Appendix~\ref{app:montezuma_results}. During fine-tuning, the agent has to solve the whole game, starting from the first room. As such, Room~7 and subsequent ones represent the \textsc{Far} states, and the preceding rooms represent \textsc{Close} states.   
We conduct experiments using PPO with Random Network Distillation~\citep{burda2018exploration} to boost exploration, which is essential in this sparse reward environment. 
More technical details, including the neural network architecture, can be found in Appendix~\ref{app:montezuma}.

\textbf{\roboticsequence{}} is a multi-stage robotic task based on the Meta-World benchmark~\cite {yu2020meta}. 
The robot is successful only if during a single episode, it completes sequentially the following sub-tasks: use a hammer to hammer in a nail (\texttt{hammer}), push an object from one specific place to another (\texttt{push}), remove a bolt from a wall (\texttt{peg-unplug-side}), push an object around a wall (\texttt{push-wall}).

We use a pre-trained policy $\pi_*$ that can solve the last two stages, \texttt{peg-unplug-side} and \texttt{push-wall} (\textsc{Far}), but not the first two, \texttt{hammer} and \texttt{push} (\textsc{Close}). See Figure~\ref{fig:state_coverage_gap} for an example of another, two-stage instantiation of \roboticsequence{}. We use Soft Actor-Critic (SAC)~\cite{haarnoja2018soft} in all robotic experiments. More technical details, including the neural network architecture, can be found in Appendix~\ref{app:robotic_details}.

\section{Main result: knowledge retention mitigates \problem{}}\label{sec:main_results}

\begin{figure*}[t]
    \centering
    \includegraphics[width=\linewidth]{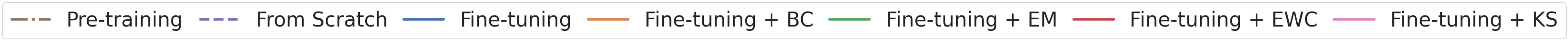}
    \begin{subfigure}[t]{0.32\textwidth}
        \centering
        \includegraphics[width=\linewidth]{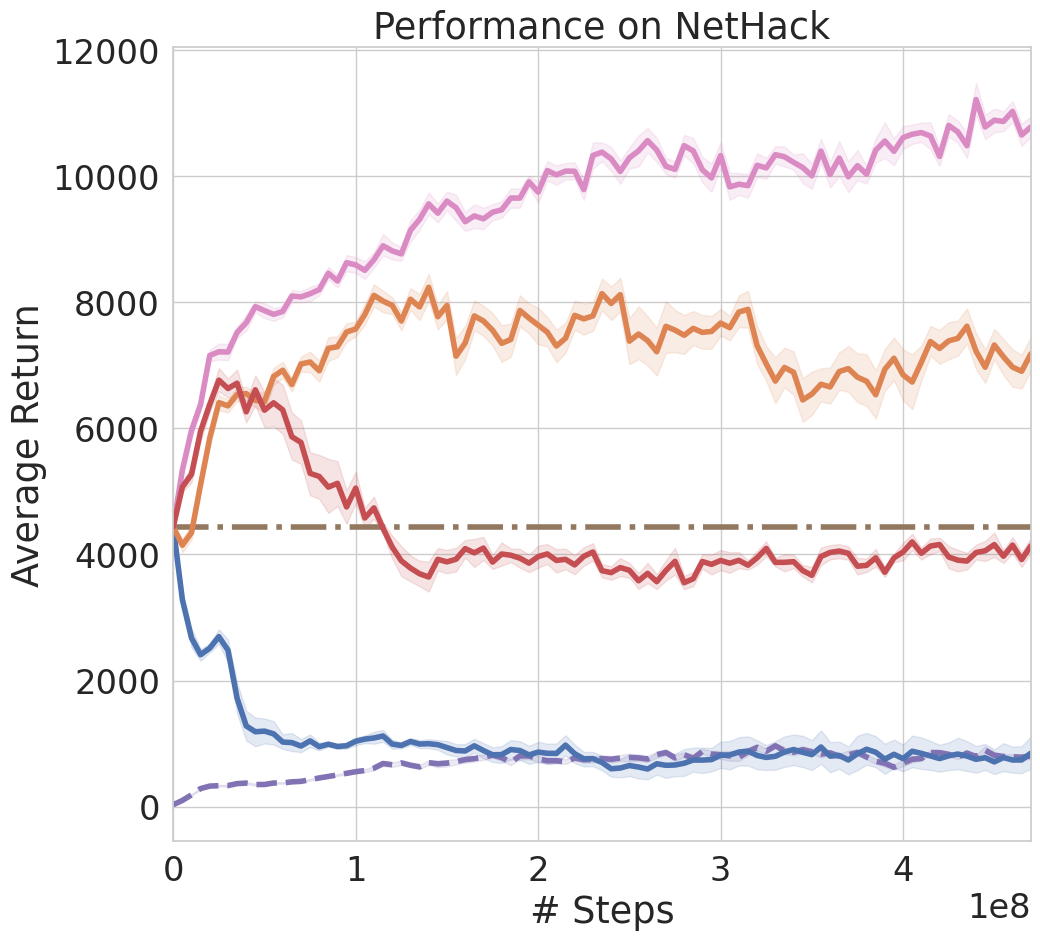}
        \caption{NetHack}
        \label{fig:main_result_nethack}
    \end{subfigure}
    \hfill
    \begin{subfigure}[t]{0.32\textwidth}
        \centering
        \includegraphics[width=\textwidth]{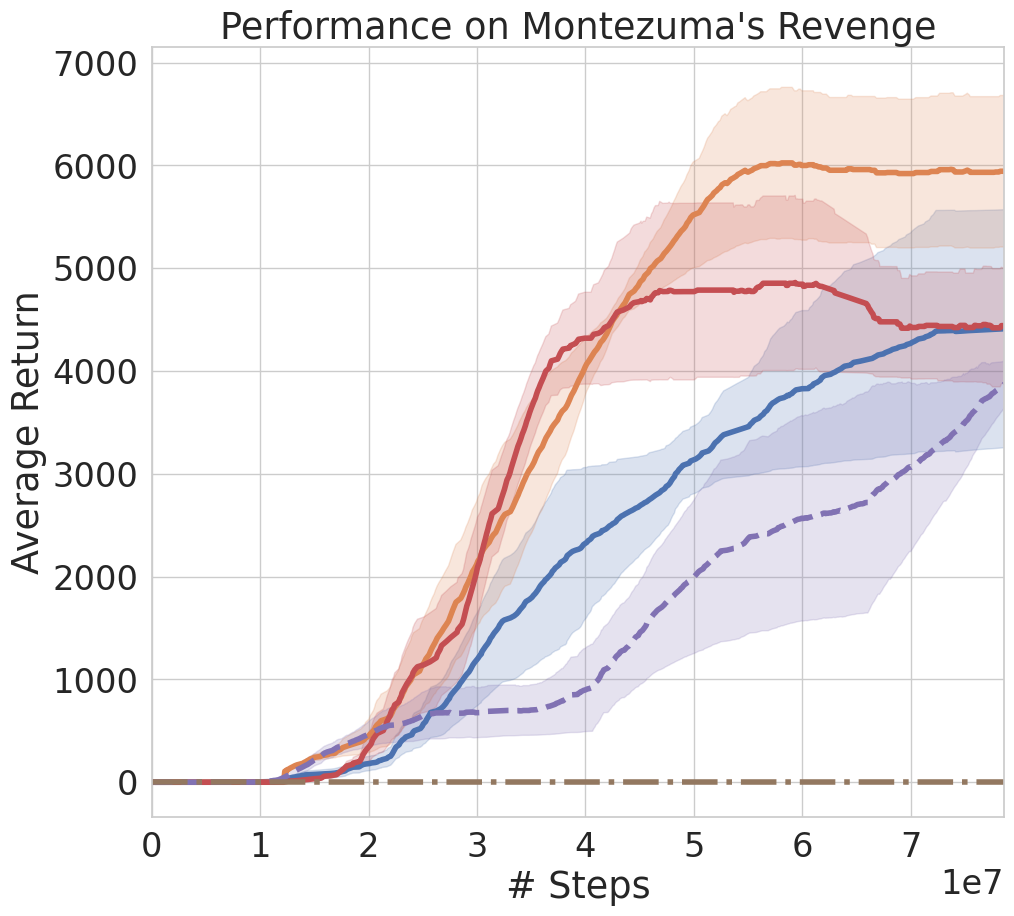}
        \caption{Montezuma's Revenge}
        \label{fig:main_result_montezuma}
    \end{subfigure}
    \hfill
    \begin{subfigure}[t]{0.32\textwidth}
        \centering
        \includegraphics[width=\textwidth]{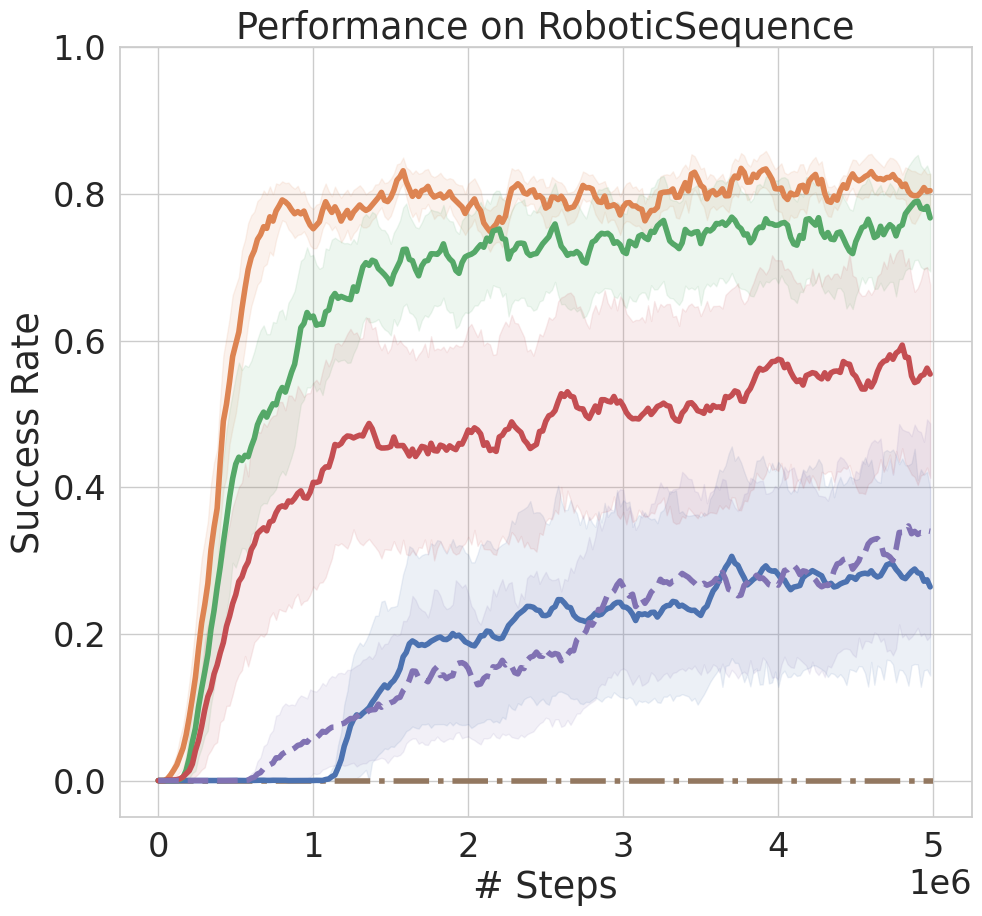}
        \caption{\roboticsequence{}}
        \label{fig:main_result_robotic}
    \end{subfigure}
    \caption{Performance on (a) NetHack, (b) Montezuma's Revenge, and (c) \roboticsequence{}. For NetHack, the \abbrv{} is driven by \imitationgap{}, while for the remaining two by \envshift{}. In all cases, knowledge retention techniques improve the performance of fine-tuning. 
    We omit KS in Montezuma's Revenge and \roboticsequence{} as it underperforms.
    }
    \label{fig:main_result}

\end{figure*}

In this section, we present empirical results showing that across all environments (1) vanilla fine-tuning often fails to leverage pre-trained knowledge, and, importantly, (2) the knowledge retention methods fix this problem, unlocking the potential of the pre-trained model and leading to significant improvements. Here, we focus on performance and defer a detailed analysis of the \problem{} phenomenon to Section~\ref{sec:analysis}. 

\textbf{NetHack} 
We demonstrate that fine-tuning coupled with knowledge retention methods surpasses the current state-of-the-art \citep{tuyls2023scaling} by $2$x, achieving $10$K points when compared to the previous $5$K, see Figure \ref{fig:main_result_nethack}. Interestingly, vanilla fine-tuning alone proves insufficient, as the agent's performance deteriorates, losing pre-trained capabilities and failing to recover from this loss.

We discover that retaining the prior knowledge unlocks the possibility of improving the policy during fine-tuning, see Figure~\ref{fig:main_result_nethack}. However, choosing an effective method for knowledge retention is nuanced, as discussed in the commentary at the end of this section. In the context of NetHack, KS works best, followed by BC, both surpassing the state-of-the-art. Conversely, EWC shows poor performance, deteriorating after some training. Importantly, implementing knowledge retention within existing frameworks is straightforward, distinguishing our method from the more intricate approaches used for NetHack~\citep{piterbarg2023nethack, klissarov2023motif}, which utilize large language models or hierarchical reinforcement learning. We note that our best agent performs well not only in terms of the overall score but other metrics that are relevant in NetHack, such as the number of visited levels or amount of gold gathered, see Appendix~\ref{app:nethack_results_big}.


\textbf{Montezuma's Revenge} 
We show that adding a knowledge retention technique in the form of BC improves not only the speed of learning but also the performance when compared to vanilla fine-tuning or training from scratch, see Figure \ref{fig:main_result_montezuma}. 
EWC also outperforms training from scratch and converges faster than vanilla fine-tuning, although it saturates on the lower average return.
The performance of the BC version starts to diverge from vanilla fine-tuning at around $20$M steps when the agent starts to enter Room~7, which is the first room observed in pre-training. 
This is where the beneficial effects of  \envshift{} mitigation come into play.


\begin{figure}[t]
\centering
    \includegraphics[width=0.95\linewidth]{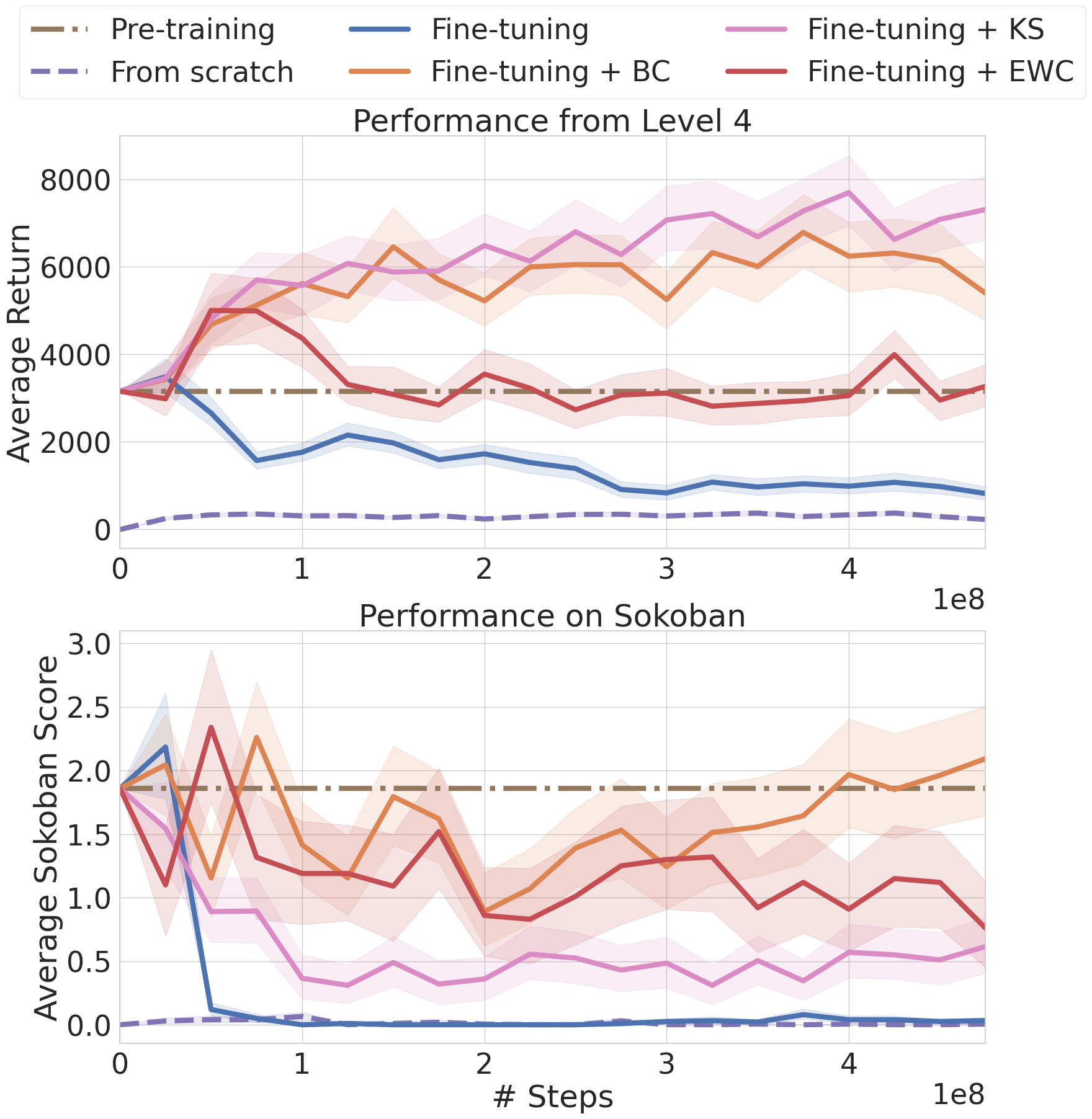} 
    \caption{
    The average return throughout the fine-tuning process on two NetHack tasks: level 4 (top), and Sokoban level (bottom). 
               The result is averaged over 200 episodes, each starting from where the expert (AutoAscend) ended up upon first entering level.}
    \label{fig:nethack_per_level}     
\end{figure}

\textbf{\roboticsequence{}} 
We show that the knowledge retention methods mitigate the \envshift{} problem and allow the model to benefit from 
pre-trained capabilities, see Figure \ref{fig:main_result_robotic}.
In terms of performance and speed of learning, BC is the most effective, followed by EM and EWC, respectively. 
Notably, BC successfully solves all four stages of \roboticsequence{} $80\%$ of the time, a strong result considering the challenges posed by compounding failure probabilities; see Figure \ref{fig:analysis_robotic} for success rates of individual stages.
Importantly, vanilla fine-tuning or training from scratch are virtually indistinguishable, and both significantly fall behind BC, EM, and EWC. 

\textbf{Discussion of knowledge retention methods} Although knowledge retention methods improve the performance of fine-tuning, the choice of the right approach is crucial. We observe that the choice between KS and BC depends on the nature of the problem and, when in doubt, it might be prudent to test both. 
For NetHack and \imitationgap{} case, where the agent is initialized to closely mimic the expert, it might be sufficient to prevent forgetting on states visited online by the fine-tuned policy, hence use KS.
%
%
On the other hand, we found that BC is successful in mitigating \envshift{}, a phenomenon appearing in Montezuma's Revenge and \roboticsequence{}, as it allows the fine-tuned policy to learn on \textsc{Close} and prevents it from forgetting on \textsc{Far}. KS fails completely in this setting, as it tries to match the pre-trained model's outputs also on \textsc{Close} states, which were not present in pre-training. As such, we do not report metrics for KS in these environments. 

Episodic memory (EM) performs well on \roboticsequence{}, where we use SAC. However, it can be only applied with algorithms that employ an off-policy replay buffer. Since NetHack and Montezuma's Revenge leverage, respectively, APPO and PPO, it cannot be trivially applied in their case. Finally, although EWC exceeds vanilla fine-tuning in all settings, it is consistently outperformed by the other approaches. 

\section{Analysis: \problem{} hinders RL fine-tuning}\label{sec:analysis}

In this section, we investigate \problem{} in detail, shedding additional light on the reasons for the poor performance of vanilla fine-tuning demonstrated in Section \ref{sec:main_results}. 
One of the findings is that the results on the \textsc{Far} states rapidly decline as we start fine-tuning. Even after re-learning, the final policy is significantly different than the pre-trained one, suggesting that the agent learned a new solution instead of benefiting from the previous one. On the other hand, fine-tuning with knowledge retention techniques is robust to these issues.

\textbf{NetHack}
Although $\pi_*$ is a relatively big model pre-trained on a large amount of data, it fails to capture 
some of AutoAscend's complex behaviors and long-term strategy, a vivid example of \imitationgap{}. 
Indeed, in Figure~\ref{fig:nethack_level_visitation} we can see \emph{a distribution shift between the expert and the pre-trained model} 
hindering fine-tuning efficiency. We also show that fine-tuning with knowledge retention (KS in this case) manages to overcome this problem and explores the dungeon in a manner more comparable to AutoAscend.


We study the extent to which knowledge retention techniques mitigate the negative effects of \imitationgap{} on two levels representing \textsc{Far} states:  
level $4$ and Sokoban level\footnotemark{}, see Figure~\ref{fig:nethack_per_level}.
\footnotetext{In NetHack, the Sokoban level is a branch of the dungeon modeled and named after an NP-hard game where the goal is to push boxes on target locations, 
see \href{run:https://nethackwiki.com/wiki/Sokoban}{NetHack wiki} and Appendix~\ref{app:nethack}.} The performance of fine-tuning on level 4 can be temporarily enhanced by EWC and consistently improved by KS or BC, which is in line with the results presented in Figure \ref{fig:main_result_nethack}. 
%
 \camera{Solving the Sokoban level does not yield immediate rewards so the vanilla fine-tuning agent pursues other strategies that are more beneficial in the short term. As such, it is not surprising that this particular behavior is forgotten. However, forgetting this skill will be hurtful in the long term, since completing the Sokoban levels unlocks a variety of equipment that is crucial for high performance during the rest of the game. Differentiating between behaviors that should be forgotten and those that should be kept is an important future direction for knowledge retention methods.}
%
%

The Sokoban results allow us to get some insights into the qualitative differences between the KS and BC. 
Namely, KS struggles with sustaining the performance on Sokoban, as uses trajectories \emph{gathered by the online policy}. These do not contain any Sokoban levels at the start of the fine-tuning, as Sokoban is only encountered in the deeper parts of the dungeon.
Conversely, BC uses data \emph{gathered by the expert} and, as a result, constantly rehearses the correct way of solving this puzzle. As such, we note that both BC and KS have their specific advantages. We identify designing methods that combine these improvements as important future work.
See Appendix \ref{app:nethack_results_big} for additional NetHack metrics.

\begin{figure}[t]
    \centering
    \includegraphics[width=\linewidth]{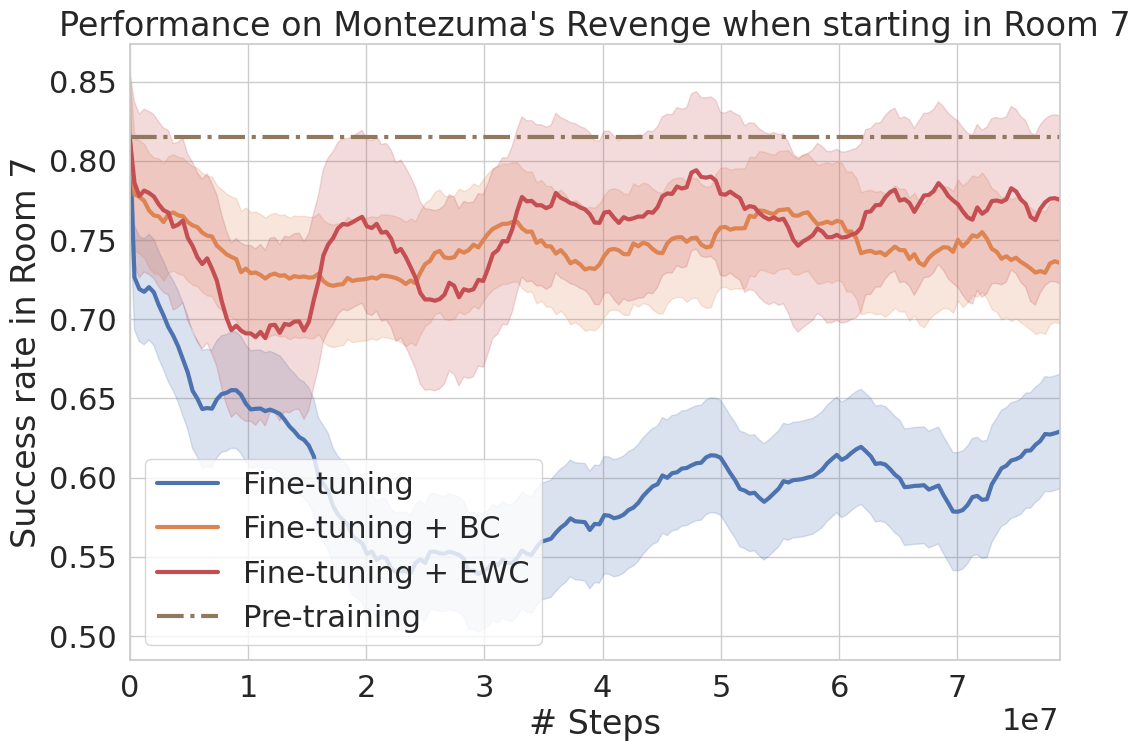}
    \caption{Montezuma's Revenge, success rate in Room~7 which represents a part of the \textsc{Far} states. } 
    \label{fig:montezuma_room10}
\end{figure}

\textbf{Montezuma's Revenge} 
We assess the scope of the \envshift{} problem by evaluating agents in Room~7, throughout the learning process, see Figure \ref{fig:montezuma_room10}. This is the first room present in pre-training and as such marks the transition between \textsc{Close} and \textsc{Far} states. Verifying the agent's performance here allows us to measure how much knowledge was lost. 
The vanilla \mbox{fine-tuning} success rate\footnotemark{} drops considerably as the training progresses. 
\footnotetext{We use this metric since the reward signal in Montezuma's Revenge is too sparse to provide reliable measurements.}
While it starts improving when the agent revisits Room~7, i.e., after $20$M environment steps, it does not reach the performance of $\pi_*$.
In contrast to this behavior, both BC and EWC 
maintain a stable success rate, closely resembling the performance of the expert policy $\pi_*$ that was pre-trained to solve the game restricted only to the levels following Room~7. 

In Appendix~\ref{app:montezuma_results} we confirm these findings with different definitions of \textsc{Far} and \textsc{Close} sets. Additionally, we study how forgetting impacts exploration, showing that with knowledge retention the agent manages to visit a larger number of rooms than with vanilla fine-tuning. 

    \begin{figure}[t]
        \includegraphics[width=\linewidth]{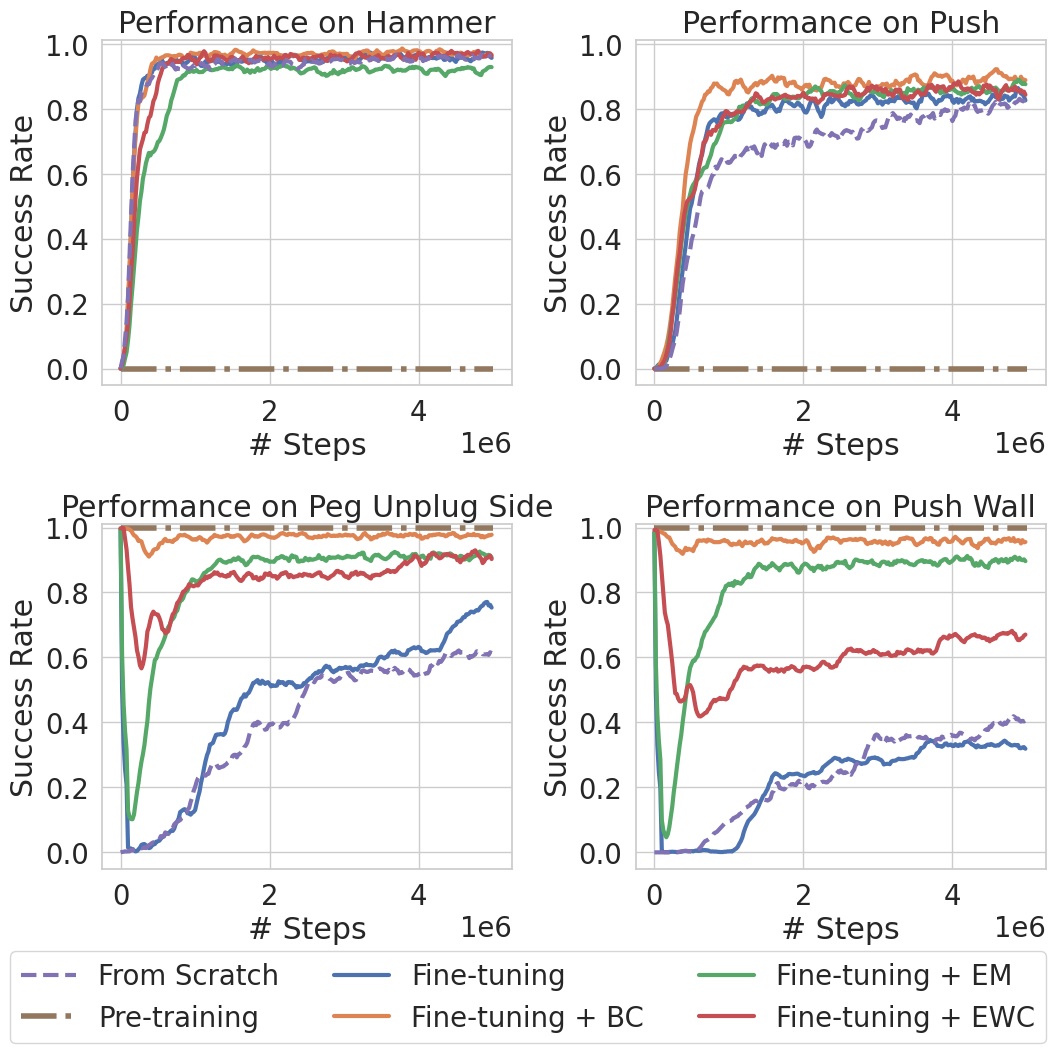}
        \caption{Success rate for each stage of \roboticsequence{}. The fine-tuning experiments start from a pre-trained policy $\pi_*$ that performs well on \texttt{peg-unplug-side} and \texttt{push-wall}.}
        \label{fig:analysis_robotic}
    \end{figure}

\begin{figure*}
    \centering
    \includegraphics[width=\textwidth]{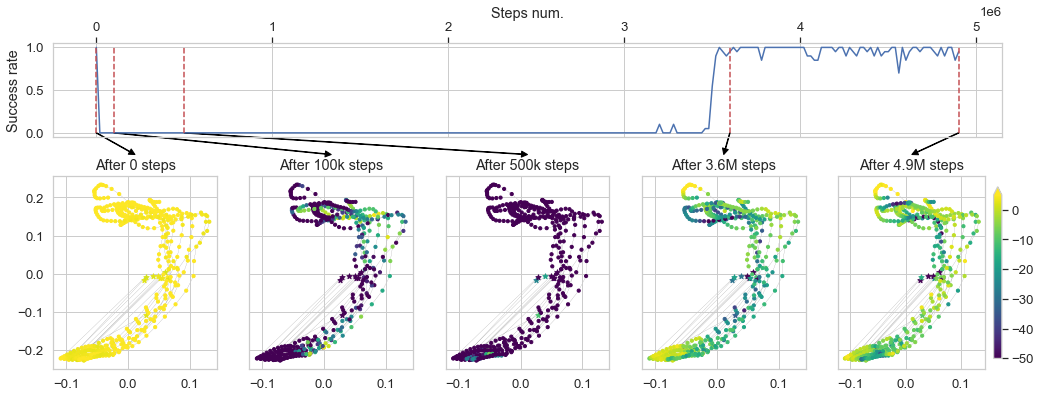}
    \caption{
    Log-likelihood under the fine-tuned policy of trajectories collected using $\pi_*$ on \texttt{push-wall}, i.e., 
    state-action pairs $(s, a^*), a^*\sim\pi_*(s)$. 
    The top row contains success rates, while the bottom row visualizes $2$D PCA projections, color-coded according to the log-likelihood.
    %
    As fine-tuning progresses the model forgets the initial solution and is unable to recover it. 
    }
    \label{fig:pca}
\end{figure*}

\textbf{\roboticsequence{}} 
Figure \ref{fig:analysis_robotic} shows that the vanilla fine-tuned agent forgets on \textsc{Far} states (stages \texttt{peg-unplug-side} and \texttt{push-wall}), again showcasing \envshift{}. 
While we observed in Section \ref{sec:main_results} that the knowledge retention methods mitigate this problem, here we can see the behavior broken down into individual stages. 
In particular, it is apparent that while learning on \texttt{hammer} or \texttt{push} (\textsc{Close}), the agent initially forgets how to perform on \textsc{Far}. 
Moreover, this deterioration is severe, i.e., when the training finally reaches these stages, the performance grows slowly. 
However, BC, EM, and EWC are able to maintain or to a certain degree regain performance (exact results vary by the method).
%
%
This pinpoints that \textit{the standard fine-tuning does not exhibit positive transfer} of the knowledge of the last two stages. 

%
%

We supplement this analysis by studying the log-likelihoods assigned by the fine-tuned policy to trajectories collected using the expert policy, i.e., the state-action pairs $(s, a^*)$, where $a^*\sim\pi_*(s)$. 
%
This is visualized on Figure \ref{fig:pca}
where we show how the policy deteriorates in certain parts of the state space (projected to 2D using PCA) in the \texttt{push-wall} environment. After $100K$ steps, the model assigns high probability to some of the correct actions on the part of the state space, but its overall success rate has already collapsed to $0$.  
As such, even partial forgetting in the initial stages significantly impacts performance. After the $500K$ steps, the likelihood values collapse on all expert trajectories. 
The situation changes when the agent relearns how to behave on \texttt{push-wall} but the log-likelihoods do not reach the original values,  
showing that the fine-tuned agent learned a different policy.

We expand this analysis in Appendix \ref{app:robotic_analysis}, showing that the hidden representation of the policy network is irreversibly changed in the early stages of fine-tuning and that \problem{} persists irrespective of the network size and aggravates as the size of~\textsc{Close} set increases.

\textbf{Other considerations}
\camera{Additionally, we note that choosing the most appropriate knowledge retention method for the problem at hand may depend on other constraints. In particular, if no prior data can be utilized during the fine-tuning, then BC and EM cannot be used, and one should instead leverage EWC, or apply Kickstarting that distills the knowledge on the online data. On the other hand, if one wishes to minimize computational complexity, EWC might be the best choice, as the other methods require processing more examples per training step. Finally, if there are restrictions on memory, one should weigh the cost of keeping the data (needed for BC, EM) against the cost of keeping the parameters of the pre-trained model (needed for EWC, KS).}

\section{Related Work}
\label{sec:related_work}

\paragraph{Transfer in RL} Due to high sample complexity and computation costs, training reinforcement learning algorithms from scratch is expensive~\citep{ceron2021revisiting,vinyals2019grandmaster,machado2018revisiting}. 
As such, transfer learning and reusing prior knowledge as much as possible~\citep{agarwal2022reincarnating} are becoming more attractive. However, the fine-tuning strategy massively popular in supervised learning~\citep{bommasani2021opportunities,yosinski2014transferable,girshick2014rich} is relatively less common in reinforcement learning. 
Approaches that are often used instead include
kickstarting without transferring the parameters~\citep{schmitt2018kickstarting,lee2022spend}, and reusing offline data~\citep{lee2022multi,kostrikov2021offline}, skills~\citep{pertsch2021guided} or the feature representations~\citep{schwarzer2021pretraining,stooke2021decoupling}, see~\citet{wulfmeier2023foundations} for a thorough discussion.

Fine-tuning in RL is often accompanied by knowledge retention mechanisms, even though they are sometimes not described as such. In particular, \citet{baker2022video} includes a regularization term to limit forgetting, \citet{kumar2022pre} mixes new data with the old data, and \citet{seo2022reinforcement} introduces modularity to the model. 
Here, we focus on the characterization and the experimental analysis of this issue in fine-tuning RL models, and pinpointing some specific settings when forgetting might occur, such as \imitationgap{}.

\camera{\paragraph{Offline to Online Reinforcement Learning} Recent work explored techniques for efficiently transitioning from offline to online reinforcement learning. \citet{ball2023efficient} use symmetric sampling of offline and online data and combine it with layer normalization and ensembles in an off-policy setting. \citet{lee2022offline} propose using a network for measuring “online-ness” of data and prioritizing samples in a replay buffer according to that measure. \citet{nakamoto2024cal} modify Conservative Q-Learning to train on a mixture of the offline data and the new online data, weighted in some proportion during fine-tuning. We highlight that mixing new data with old data can be viewed as a knowledge retention technique similar to Episodic Memory. Although these approaches are relevant to our study and we see testing them as important future work, we use behavioral cloning in pre-training for simplicity, especially as it has been shown to outperform offline RL methods in the NetHack domain~\cite{hambro2022dungeons}.
}

\camera{\paragraph{Impact of interdependence between \textsc{Far} and \textsc{Close}} The relation between \textsc{Far} and \textsc{Close} states has an important impact on the degree of forgetting, which might be understood through the lens of CL literature on task similarity. For example, \citet{lee2021continual} find that intermediate task similarity levels lead to the highest degrees of forgetting. \citet{evron2022catastrophic} reach a similar conclusion in the linear regression setting when a given task is seen only once, but also find that high similarity causes most forgetting when repeatedly revisiting tasks.
Furthermore, \citet{evron2024joint} suggest that this behavior might be explained by heavy overparameterization, since in non-overparameterized cases forgetting grows monotonically as the task difference increases.} 

\camera{\paragraph{Generalization to multi-task setting} While our work focuses on single-task fine-tuning, prior research has explored fine-tuning on multiple unseen tasks. \citet{yang2023essential} compared offline RL methods with imitation learning in a 2D goal-reaching aiming to test generalization to unseen goals. \citet{mandi2022effectiveness} showed that multi-task pre-training with fine-tuning often outperformed meta-reinforcement learning approaches in adaptation tasks with high task diversity and strictly unseen test tasks. 
At the same time, we believe studying single-task fine-tuning in NetHack provides valuable preliminary insights into this problem, as the game's procedural generation on each run requires flexibly applying learned skills to adapt to new contexts.}

\paragraph{Continual reinforcement learning} Continual RL deals with learning over a changing stream of tasks represented as MDPs
\citep{khetarpal2022towards,wolczyk2021continual,nekoei2021continuous,powers2022cora,huang2021continual,kessler2022surprising}. 
Several works propose methods for continual reinforcement learning based on replay and distillation~\citep{rolnick2019experience,traore2019discorl}, or modularity~\citep{mendez2022modular,gaya2022building}.
Although relevant to our study, these works usually investigate changes in the dynamics of non-stationary environments. In this paper, we switch the perspective and focus on the data shifts occurring during fine-tuning in a stationary environment. In fact, some of the standard techniques in RL, such as using the replay buffer, can be seen as a way to tame the non-stationarity inherent to RL~\citep{lin1992reinforcement,mnih2013playing}. For a further discussion about how our setup differs from continual reinforcement learning, see Section~\ref{sec:preliminaries}. 

\section{Limitations \& Conclusions}
This study shows that \problem{} is a crucial consideration for fine-tuning RL models. Namely, we verify in multiple scenarios, ranging from toy MDPs to the challenging NetHack domain, that fine-tuning a model on a task where the states from pre-training are not available at the beginning of the training might lead to a rapid deterioration of the prior knowledge. We highlight two specific cases: \envshift{} and \imitationgap{}. 

Although we aim to comprehensively describe~\problem{}, our study is limited in several ways. In our experiments, we used fairly simple knowledge retention methods to illustrate the forgetting problem. We believe that CL offers numerous more sophisticated methods that should achieve great results on this problem~\citep{mallya2018packnet,ben2022lifelong,mendez2022modular,khetarpal2022towards}. Additionally, we note that knowledge retention methods can be harmful if the pre-trained policy is suboptimal since they will stop the fine-tuned policy from improving. 
In some environments, it might not be easy to identify the part of the state space where the policy should be preserved. Furthermore, we focus on two specific transfer scenarios, while in the real world, there are many more settings exhibiting unique problems.  Finally, we do not study very large models (i.e. over $1B$ parameters) and efficient approaches to fine-tuning that tune only selected parameters~\cite{xu2023parameter,hu2021lora}. We see all these topics as important directions for future work.

\camera{While our study focuses on the RL setting, some of its findings might have a broader scope. Non-stationary dynamics might also emerge in the supervised learning i.i.d. setting when the model sequentially acquires increasingly sophisticated skills (e.g., LLMs first learn simple grammar and understand advanced skills only much later) \cite{evanson2023language, luo2023empirical}. This suggests that the principles of knowledge retention and forgetting we explored could be relevant beyond the specific RL scenarios we tested, potentially impacting a wide range of learning systems that evolve over time. A comprehensive examination of these dynamics across different learning models and environments remains a crucial area for future research.}

\section*{Impact statement}
Our main focus is improving the transfer capabilities of reinforcement learning models. We do not foresee any major societal impact of this study that we feel should be highlighted here. 


\section*{Acknowledgements}

The work of MO and MB was funded by National Science Center Poland under the grant agreement 2020/39/B/ST6/01511 and by Warsaw University of Technology within the Excellence Initiative: Research University (IDUB) programme. PM was supported by National Science Center Poland under the grant agreement 2019/35/O/ST6/03464. We gratefully acknowledge Polish high-performance computing infrastructure PLGrid (HPC Centers: ACK Cyfronet AGH) for providing computer facilities and support within computational grant no. PLG/2023/016286


\nocite{kessler2023effectiveness}
\bibliography{icml2024}

\begin{thebibliography}{101}
\providecommand{\natexlab}[1]{#1}
\providecommand{\url}[1]{\texttt{#1}}
\expandafter\ifx\csname urlstyle\endcsname\relax
  \providecommand{\doi}[1]{doi: #1}\else
  \providecommand{\doi}{doi: \begingroup \urlstyle{rm}\Url}\fi

\bibitem[Agarwal et~al.(2022)Agarwal, Schwarzer, Castro, Courville, and Bellemare]{agarwal2022reincarnating}
Agarwal, R., Schwarzer, M., Castro, P.~S., Courville, A., and Bellemare, M.~G.
\newblock Reincarnating reinforcement learning: Reusing prior computation to accelerate progress.
\newblock \emph{arXiv preprint arXiv:2206.01626}, 2022.

\bibitem[Aljundi et~al.(2018)Aljundi, Babiloni, Elhoseiny, Rohrbach, and Tuytelaars]{aljundi2018memory}
Aljundi, R., Babiloni, F., Elhoseiny, M., Rohrbach, M., and Tuytelaars, T.
\newblock Memory aware synapses: Learning what (not) to forget.
\newblock In \emph{Proceedings of the European conference on computer vision (ECCV)}, pp.\  139--154, 2018.

\bibitem[Bain \& Sammut(1995)Bain and Sammut]{bain1995framework}
Bain, M. and Sammut, C.
\newblock A framework for behavioural cloning.
\newblock In \emph{Machine Intelligence 15}, pp.\  103--129, 1995.

\bibitem[Baker et~al.(2022)Baker, Akkaya, Zhokhov, Huizinga, Tang, Ecoffet, Houghton, Sampedro, and Clune]{baker2022video}
Baker, B., Akkaya, I., Zhokhov, P., Huizinga, J., Tang, J., Ecoffet, A., Houghton, B., Sampedro, R., and Clune, J.
\newblock Video pretraining (vpt): Learning to act by watching unlabeled online videos.
\newblock \emph{arXiv preprint arXiv:2206.11795}, 2022.

\bibitem[Ball et~al.(2023)Ball, Smith, Kostrikov, and Levine]{ball2023efficient}
Ball, P.~J., Smith, L., Kostrikov, I., and Levine, S.
\newblock Efficient online reinforcement learning with offline data.
\newblock In \emph{International Conference on Machine Learning}, pp.\  1577--1594. PMLR, 2023.

\bibitem[Bellemare et~al.(2013)Bellemare, Naddaf, Veness, and Bowling]{bellemare2013arcade}
Bellemare, M.~G., Naddaf, Y., Veness, J., and Bowling, M.
\newblock The arcade learning environment: An evaluation platform for general agents.
\newblock \emph{Journal of Artificial Intelligence Research}, 47:\penalty0 253--279, 2013.

\bibitem[Ben-Iwhiwhu et~al.(2022)Ben-Iwhiwhu, Nath, Pilly, Kolouri, and Soltoggio]{ben2022lifelong}
Ben-Iwhiwhu, E., Nath, S., Pilly, P.~K., Kolouri, S., and Soltoggio, A.
\newblock Lifelong reinforcement learning with modulating masks.
\newblock \emph{arXiv preprint arXiv:2212.11110}, 2022.

\bibitem[Bommasani et~al.(2021)Bommasani, Hudson, Adeli, Altman, Arora, von Arx, Bernstein, Bohg, Bosselut, Brunskill, et~al.]{bommasani2021opportunities}
Bommasani, R., Hudson, D.~A., Adeli, E., Altman, R., Arora, S., von Arx, S., Bernstein, M.~S., Bohg, J., Bosselut, A., Brunskill, E., et~al.
\newblock On the opportunities and risks of foundation models.
\newblock \emph{arXiv preprint arXiv:2108.07258}, 2021.

\bibitem[Bornschein et~al.(2022)Bornschein, Galashov, Hemsley, Rannen-Triki, Chen, Chaudhry, He, Douillard, Caccia, Feng, et~al.]{bornschein2022nevis}
Bornschein, J., Galashov, A., Hemsley, R., Rannen-Triki, A., Chen, Y., Chaudhry, A., He, X.~O., Douillard, A., Caccia, M., Feng, Q., et~al.
\newblock Nevis'22: A stream of 100 tasks sampled from 30 years of computer vision research.
\newblock \emph{arXiv preprint arXiv:2211.11747}, 2022.

\bibitem[Burda et~al.(2018)Burda, Edwards, Storkey, and Klimov]{burda2018exploration}
Burda, Y., Edwards, H., Storkey, A., and Klimov, O.
\newblock Exploration by random network distillation.
\newblock \emph{International Conference On Learning Representations}, 2018.

\bibitem[Buzzega et~al.(2021)Buzzega, Boschini, Porrello, and Calderara]{buzzega2021rethinking}
Buzzega, P., Boschini, M., Porrello, A., and Calderara, S.
\newblock Rethinking experience replay: a bag of tricks for continual learning.
\newblock In \emph{2020 25th International Conference on Pattern Recognition (ICPR)}, pp.\  2180--2187. IEEE, 2021.

\bibitem[Ceron \& Castro(2021)Ceron and Castro]{ceron2021revisiting}
Ceron, J. S.~O. and Castro, P.~S.
\newblock Revisiting rainbow: Promoting more insightful and inclusive deep reinforcement learning research.
\newblock In \emph{International Conference on Machine Learning}, pp.\  1373--1383. PMLR, 2021.

\bibitem[Chaudhry et~al.(2019)Chaudhry, Rohrbach, Elhoseiny, Ajanthan, Dokania, Torr, and Ranzato]{chaudhry2019tiny}
Chaudhry, A., Rohrbach, M., Elhoseiny, M., Ajanthan, T., Dokania, P.~K., Torr, P.~H., and Ranzato, M.
\newblock On tiny episodic memories in continual learning.
\newblock \emph{arXiv preprint arXiv:1902.10486}, 2019.

\bibitem[Chithrananda et~al.(2020)Chithrananda, Grand, and Ramsundar]{chithrananda2020chemberta}
Chithrananda, S., Grand, G., and Ramsundar, B.
\newblock Chemberta: Large-scale self-supervised pretraining for molecular property prediction.
\newblock \emph{arXiv preprint arXiv:2010.09885}, 2020.

\bibitem[Chung et~al.(2022)Chung, Hou, Longpre, Zoph, Tay, Fedus, Li, Wang, Dehghani, Brahma, et~al.]{chung2022scaling}
Chung, H.~W., Hou, L., Longpre, S., Zoph, B., Tay, Y., Fedus, W., Li, E., Wang, X., Dehghani, M., Brahma, S., et~al.
\newblock Scaling instruction-finetuned language models.
\newblock \emph{arXiv preprint arXiv:2210.11416}, 2022.

\bibitem[De~Lange et~al.(2021)De~Lange, Aljundi, Masana, Parisot, Jia, Leonardis, Slabaugh, and Tuytelaars]{de2021continual}
De~Lange, M., Aljundi, R., Masana, M., Parisot, S., Jia, X., Leonardis, A., Slabaugh, G., and Tuytelaars, T.
\newblock A continual learning survey: Defying forgetting in classification tasks.
\newblock \emph{IEEE transactions on pattern analysis and machine intelligence}, 44\penalty0 (7):\penalty0 3366--3385, 2021.

\bibitem[Devlin et~al.(2019)Devlin, Chang, Lee, and Toutanova]{Devlin2019BERTPO}
Devlin, J., Chang, M.-W., Lee, K., and Toutanova, K.
\newblock Bert: Pre-training of deep bidirectional transformers for language understanding.
\newblock \emph{ArXiv}, abs/1810.04805, 2019.

\bibitem[Dosovitskiy et~al.(2020)Dosovitskiy, Beyer, Kolesnikov, Weissenborn, Zhai, Unterthiner, Dehghani, Minderer, Heigold, Gelly, et~al.]{dosovitskiy2020image}
Dosovitskiy, A., Beyer, L., Kolesnikov, A., Weissenborn, D., Zhai, X., Unterthiner, T., Dehghani, M., Minderer, M., Heigold, G., Gelly, S., et~al.
\newblock An image is worth 16x16 words: Transformers for image recognition at scale.
\newblock \emph{arXiv preprint arXiv:2010.11929}, 2020.

\bibitem[Evanson et~al.(2023)Evanson, Lakretz, and King]{evanson2023language}
Evanson, L., Lakretz, Y., and King, J.-R.
\newblock Language acquisition: do children and language models follow similar learning stages?
\newblock \emph{arXiv preprint arXiv:2306.03586}, 2023.

\bibitem[Evron et~al.(2022)Evron, Moroshko, Ward, Srebro, and Soudry]{evron2022catastrophic}
Evron, I., Moroshko, E., Ward, R., Srebro, N., and Soudry, D.
\newblock How catastrophic can catastrophic forgetting be in linear regression?
\newblock In \emph{Conference on Learning Theory}, pp.\  4028--4079. PMLR, 2022.

\bibitem[Evron et~al.(2024)Evron, Goldfarb, Weinberger, Soudry, and Hand]{evron2024joint}
Evron, I., Goldfarb, D., Weinberger, N., Soudry, D., and Hand, P.
\newblock The joint effect of task similarity and overparameterization on catastrophic forgetting--an analytical model.
\newblock \emph{arXiv preprint arXiv:2401.12617}, 2024.

\bibitem[Gaya et~al.(2022)Gaya, Doan, Caccia, Soulier, Denoyer, and Raileanu]{gaya2022building}
Gaya, J.-B., Doan, T., Caccia, L., Soulier, L., Denoyer, L., and Raileanu, R.
\newblock Building a subspace of policies for scalable continual learning.
\newblock \emph{arXiv preprint arXiv:2211.10445}, 2022.

\bibitem[Girshick et~al.(2014)Girshick, Donahue, Darrell, and Malik]{girshick2014rich}
Girshick, R., Donahue, J., Darrell, T., and Malik, J.
\newblock Rich feature hierarchies for accurate object detection and semantic segmentation.
\newblock In \emph{Proceedings of the IEEE conference on computer vision and pattern recognition}, pp.\  580--587, 2014.

\bibitem[Gretton et~al.(2005)Gretton, Bousquet, Smola, and Sch{\"o}lkopf]{gretton2005measuring}
Gretton, A., Bousquet, O., Smola, A., and Sch{\"o}lkopf, B.
\newblock Measuring statistical dependence with hilbert-schmidt norms.
\newblock In \emph{Algorithmic Learning Theory: 16th International Conference, ALT 2005, Singapore, October 8-11, 2005. Proceedings 16}, pp.\  63--77. Springer, 2005.

\bibitem[Haarnoja et~al.(2018{\natexlab{a}})Haarnoja, Zhou, Abbeel, and Levine]{haarnoja2018soft}
Haarnoja, T., Zhou, A., Abbeel, P., and Levine, S.
\newblock Soft actor-critic: Off-policy maximum entropy deep reinforcement learning with a stochastic actor.
\newblock In \emph{International conference on machine learning}, pp.\  1861--1870. PMLR, 2018{\natexlab{a}}.

\bibitem[Haarnoja et~al.(2018{\natexlab{b}})Haarnoja, Zhou, Hartikainen, Tucker, Ha, Tan, Kumar, Zhu, Gupta, Abbeel, et~al.]{haarnoja2018softapp}
Haarnoja, T., Zhou, A., Hartikainen, K., Tucker, G., Ha, S., Tan, J., Kumar, V., Zhu, H., Gupta, A., Abbeel, P., et~al.
\newblock Soft actor-critic algorithms and applications.
\newblock \emph{arXiv preprint arXiv:1812.05905}, 2018{\natexlab{b}}.

\bibitem[Hambro et~al.(2022{\natexlab{a}})Hambro, Mohanty, Babaev, Byeon, Chakraborty, Grefenstette, Jiang, Daejin, Kanervisto, Kim, et~al.]{hambro2022insights}
Hambro, E., Mohanty, S., Babaev, D., Byeon, M., Chakraborty, D., Grefenstette, E., Jiang, M., Daejin, J., Kanervisto, A., Kim, J., et~al.
\newblock Insights from the neurips 2021 nethack challenge.
\newblock In \emph{NeurIPS 2021 Competitions and Demonstrations Track}, pp.\  41--52. PMLR, 2022{\natexlab{a}}.

\bibitem[Hambro et~al.(2022{\natexlab{b}})Hambro, Raileanu, Rothermel, Mella, Rockt{\"a}schel, K{\"u}ttler, and Murray]{hambro2022dungeons}
Hambro, E., Raileanu, R., Rothermel, D., Mella, V., Rockt{\"a}schel, T., K{\"u}ttler, H., and Murray, N.
\newblock Dungeons and data: A large-scale nethack dataset.
\newblock \emph{Advances in Neural Information Processing Systems}, 35:\penalty0 24864--24878, 2022{\natexlab{b}}.

\bibitem[Hambro et~al.(2022{\natexlab{c}})Hambro, Raileanu, Rothermel, Mella, Rockt{\"a}schel, Kuttler, and Murray]{hambrodungeons}
Hambro, E., Raileanu, R., Rothermel, D., Mella, V., Rockt{\"a}schel, T., Kuttler, H., and Murray, N.
\newblock Dungeons and data: A large-scale nethack dataset.
\newblock In \emph{Thirty-sixth Conference on Neural Information Processing Systems Datasets and Benchmarks Track}, 2022{\natexlab{c}}.

\bibitem[Hu et~al.(2021)Hu, Shen, Wallis, Allen-Zhu, Li, Wang, Wang, and Chen]{hu2021lora}
Hu, E.~J., Shen, Y., Wallis, P., Allen-Zhu, Z., Li, Y., Wang, S., Wang, L., and Chen, W.
\newblock Lora: Low-rank adaptation of large language models.
\newblock \emph{arXiv preprint arXiv:2106.09685}, 2021.

\bibitem[Huang et~al.(2021)Huang, Xie, Bharadhwaj, and Shkurti]{huang2021continual}
Huang, Y., Xie, K., Bharadhwaj, H., and Shkurti, F.
\newblock Continual model-based reinforcement learning with hypernetworks.
\newblock In \emph{2021 IEEE International Conference on Robotics and Automation (ICRA)}, pp.\  799--805. IEEE, 2021.

\bibitem[Kemker et~al.(2018)Kemker, McClure, Abitino, Hayes, and Kanan]{kemker2018measuring}
Kemker, R., McClure, M., Abitino, A., Hayes, T., and Kanan, C.
\newblock Measuring catastrophic forgetting in neural networks.
\newblock In \emph{Proceedings of the AAAI conference on artificial intelligence}, volume~32, 2018.

\bibitem[Kessler et~al.(2022{\natexlab{a}})Kessler, Mi{\l}o{\'s}, Parker-Holder, and Roberts]{kessler2022surprising}
Kessler, S., Mi{\l}o{\'s}, P., Parker-Holder, J., and Roberts, S.~J.
\newblock The surprising effectiveness of latent world models for continual reinforcement learning.
\newblock \emph{arXiv preprint arXiv:2211.15944}, 2022{\natexlab{a}}.

\bibitem[Kessler et~al.(2022{\natexlab{b}})Kessler, Parker-Holder, Ball, Zohren, and Roberts]{kessler2022same}
Kessler, S., Parker-Holder, J., Ball, P., Zohren, S., and Roberts, S.~J.
\newblock Same state, different task: Continual reinforcement learning without interference.
\newblock In \emph{Proceedings of the AAAI Conference on Artificial Intelligence}, volume~36, pp.\  7143--7151, 2022{\natexlab{b}}.

\bibitem[Kessler et~al.(2023)Kessler, Ostaszewski, Bortkiewicz, Żarski, Wołczyk, Parker-Holder, Roberts, and Miłoś]{kessler2023effectiveness}
Kessler, S., Ostaszewski, M., Bortkiewicz, M., Żarski, M., Wołczyk, M., Parker-Holder, J., Roberts, S.~J., and Miłoś, P.
\newblock The effectiveness of world models for continual reinforcement learning, 2023.

\bibitem[Khetarpal et~al.(2022)Khetarpal, Riemer, Rish, and Precup]{khetarpal2022towards}
Khetarpal, K., Riemer, M., Rish, I., and Precup, D.
\newblock Towards continual reinforcement learning: A review and perspectives.
\newblock \emph{Journal of Artificial Intelligence Research}, 75:\penalty0 1401--1476, 2022.

\bibitem[Kingma \& Ba(2014)Kingma and Ba]{kingma2014adam}
Kingma, D.~P. and Ba, J.
\newblock Adam: A method for stochastic optimization.
\newblock \emph{arXiv preprint arXiv:1412.6980}, 2014.

\bibitem[Kirkpatrick et~al.(2017)Kirkpatrick, Pascanu, Rabinowitz, Veness, Desjardins, Rusu, Milan, Quan, Ramalho, Grabska-Barwinska, et~al.]{kirkpatrick2017overcoming}
Kirkpatrick, J., Pascanu, R., Rabinowitz, N., Veness, J., Desjardins, G., Rusu, A.~A., Milan, K., Quan, J., Ramalho, T., Grabska-Barwinska, A., et~al.
\newblock Overcoming catastrophic forgetting in neural networks.
\newblock \emph{Proceedings of the national academy of sciences}, 114\penalty0 (13):\penalty0 3521--3526, 2017.

\bibitem[Klissarov et~al.(2023)Klissarov, D'Oro, Sodhani, Raileanu, Bacon, Vincent, Zhang, and Henaff]{klissarov2023motif}
Klissarov, M., D'Oro, P., Sodhani, S., Raileanu, R., Bacon, P.-L., Vincent, P., Zhang, A., and Henaff, M.
\newblock Motif: Intrinsic motivation from artificial intelligence feedback.
\newblock \emph{arXiv preprint arXiv:2310.00166}, 2023.

\bibitem[Kornblith et~al.(2019)Kornblith, Norouzi, Lee, and Hinton]{kornblith2019similarity}
Kornblith, S., Norouzi, M., Lee, H., and Hinton, G.
\newblock Similarity of neural network representations revisited.
\newblock In \emph{International Conference on Machine Learning}, pp.\  3519--3529. PMLR, 2019.

\bibitem[Kornblith et~al.(2021)Kornblith, Chen, Lee, and Norouzi]{kornblith2021better}
Kornblith, S., Chen, T., Lee, H., and Norouzi, M.
\newblock Why do better loss functions lead to less transferable features?
\newblock \emph{Advances in Neural Information Processing Systems}, 34:\penalty0 28648--28662, 2021.

\bibitem[Kostrikov et~al.(2021)Kostrikov, Nair, and Levine]{kostrikov2021offline}
Kostrikov, I., Nair, A., and Levine, S.
\newblock Offline reinforcement learning with implicit q-learning.
\newblock \emph{arXiv preprint arXiv:2110.06169}, 2021.

\bibitem[Kumar et~al.(2022)Kumar, Singh, Ebert, Yang, Finn, and Levine]{kumar2022pre}
Kumar, A., Singh, A., Ebert, F., Yang, Y., Finn, C., and Levine, S.
\newblock Pre-training for robots: Offline rl enables learning new tasks from a handful of trials.
\newblock \emph{arXiv preprint arXiv:2210.05178}, 2022.

\bibitem[K{\"u}ttler et~al.(2020)K{\"u}ttler, Nardelli, Miller, Raileanu, Selvatici, Grefenstette, and Rockt{\"a}schel]{kuttler2020nethack}
K{\"u}ttler, H., Nardelli, N., Miller, A., Raileanu, R., Selvatici, M., Grefenstette, E., and Rockt{\"a}schel, T.
\newblock The nethack learning environment.
\newblock \emph{Advances in Neural Information Processing Systems}, 33:\penalty0 7671--7684, 2020.

\bibitem[Lee et~al.(2022{\natexlab{a}})Lee, Devin, Springenberg, Zhou, Lampe, Abdolmaleki, and Bousmalis]{lee2022spend}
Lee, A.~X., Devin, C., Springenberg, J.~T., Zhou, Y., Lampe, T., Abdolmaleki, A., and Bousmalis, K.
\newblock How to spend your robot time: Bridging kickstarting and offline reinforcement learning for vision-based robotic manipulation.
\newblock In \emph{2022 IEEE/RSJ International Conference on Intelligent Robots and Systems (IROS)}, pp.\  2468--2475. IEEE, 2022{\natexlab{a}}.

\bibitem[Lee et~al.(2022{\natexlab{b}})Lee, Nachum, Yang, Lee, Freeman, Xu, Guadarrama, Fischer, Jang, Michalewski, et~al.]{lee2022multi}
Lee, K.-H., Nachum, O., Yang, M., Lee, L., Freeman, D., Xu, W., Guadarrama, S., Fischer, I., Jang, E., Michalewski, H., et~al.
\newblock Multi-game decision transformers.
\newblock \emph{arXiv preprint arXiv:2205.15241}, 2022{\natexlab{b}}.

\bibitem[Lee et~al.(2021)Lee, Goldt, and Saxe]{lee2021continual}
Lee, S., Goldt, S., and Saxe, A.
\newblock Continual learning in the teacher-student setup: Impact of task similarity.
\newblock In \emph{International Conference on Machine Learning}, pp.\  6109--6119. PMLR, 2021.

\bibitem[Lee et~al.(2022{\natexlab{c}})Lee, Seo, Lee, Abbeel, and Shin]{lee2022offline}
Lee, S., Seo, Y., Lee, K., Abbeel, P., and Shin, J.
\newblock Offline-to-online reinforcement learning via balanced replay and pessimistic q-ensemble.
\newblock In \emph{Conference on Robot Learning}, pp.\  1702--1712. PMLR, 2022{\natexlab{c}}.

\bibitem[Lesort et~al.(2022)Lesort, Ostapenko, Misra, Arefin, Rodr{\'\i}guez, Charlin, and Rish]{lesort2022scaling}
Lesort, T., Ostapenko, O., Misra, D., Arefin, M.~R., Rodr{\'\i}guez, P., Charlin, L., and Rish, I.
\newblock Scaling the number of tasks in continual learning.
\newblock \emph{arXiv preprint arXiv:2207.04543}, 2022.

\bibitem[Lin(1992)]{lin1992reinforcement}
Lin, L.-J.
\newblock \emph{Reinforcement learning for robots using neural networks}.
\newblock Carnegie Mellon University, 1992.

\bibitem[Luo et~al.(2023)Luo, Yang, Meng, Li, Zhou, and Zhang]{luo2023empirical}
Luo, Y., Yang, Z., Meng, F., Li, Y., Zhou, J., and Zhang, Y.
\newblock An empirical study of catastrophic forgetting in large language models during continual fine-tuning.
\newblock \emph{arXiv preprint arXiv:2308.08747}, 2023.

\bibitem[Machado et~al.(2018{\natexlab{a}})Machado, Bellemare, Talvitie, Veness, Hausknecht, and Bowling]{machado2018revisiting}
Machado, M.~C., Bellemare, M.~G., Talvitie, E., Veness, J., Hausknecht, M., and Bowling, M.
\newblock Revisiting the arcade learning environment: Evaluation protocols and open problems for general agents.
\newblock \emph{Journal of Artificial Intelligence Research}, 61:\penalty0 523--562, 2018{\natexlab{a}}.

\bibitem[Machado et~al.(2018{\natexlab{b}})Machado, Bellemare, Talvitie, Veness, Hausknecht, and Bowling]{machado18arcade}
Machado, M.~C., Bellemare, M.~G., Talvitie, E., Veness, J., Hausknecht, M.~J., and Bowling, M.
\newblock Revisiting the arcade learning environment: Evaluation protocols and open problems for general agents.
\newblock \emph{Journal of Artificial Intelligence Research}, 61:\penalty0 523--562, 2018{\natexlab{b}}.

\bibitem[Mallya \& Lazebnik(2018)Mallya and Lazebnik]{mallya2018packnet}
Mallya, A. and Lazebnik, S.
\newblock Packnet: Adding multiple tasks to a single network by iterative pruning.
\newblock In \emph{Proceedings of the IEEE conference on Computer Vision and Pattern Recognition}, pp.\  7765--7773, 2018.

\bibitem[Mandi et~al.(2022)Mandi, Abbeel, and James]{mandi2022effectiveness}
Mandi, Z., Abbeel, P., and James, S.
\newblock On the effectiveness of fine-tuning versus meta-reinforcement learning.
\newblock \emph{arXiv preprint arXiv:2206.03271}, 2022.

\bibitem[Mendez et~al.(2022)Mendez, van Seijen, and Eaton]{mendez2022modular}
Mendez, J.~A., van Seijen, H., and Eaton, E.
\newblock Modular lifelong reinforcement learning via neural composition.
\newblock \emph{arXiv preprint arXiv:2207.00429}, 2022.

\bibitem[Mirzadeh et~al.(2022)Mirzadeh, Chaudhry, Yin, Nguyen, Pascanu, Gorur, and Farajtabar]{mirzadeh2022architecture}
Mirzadeh, S.~I., Chaudhry, A., Yin, D., Nguyen, T., Pascanu, R., Gorur, D., and Farajtabar, M.
\newblock Architecture matters in continual learning.
\newblock \emph{arXiv preprint arXiv:2202.00275}, 2022.

\bibitem[Mnih et~al.(2013)Mnih, Kavukcuoglu, Silver, Graves, Antonoglou, Wierstra, and Riedmiller]{mnih2013playing}
Mnih, V., Kavukcuoglu, K., Silver, D., Graves, A., Antonoglou, I., Wierstra, D., and Riedmiller, M.
\newblock Playing atari with deep reinforcement learning.
\newblock \emph{arXiv preprint arXiv:1312.5602}, 2013.

\bibitem[Mu et~al.(2022)Mu, Zhong, Raileanu, Jiang, Goodman, Rockt{\"a}schel, and Grefenstette]{mu2022improving}
Mu, J., Zhong, V., Raileanu, R., Jiang, M., Goodman, N., Rockt{\"a}schel, T., and Grefenstette, E.
\newblock Improving intrinsic exploration with language abstractions.
\newblock \emph{Advances in Neural Information Processing Systems}, 35:\penalty0 33947--33960, 2022.

\bibitem[Nair et~al.(2020)Nair, Gupta, Dalal, and Levine]{nair2020awac}
Nair, A., Gupta, A., Dalal, M., and Levine, S.
\newblock Awac: Accelerating online reinforcement learning with offline datasets.
\newblock \emph{arXiv preprint arXiv:2006.09359}, 2020.

\bibitem[Nakamoto et~al.(2024)Nakamoto, Zhai, Singh, Sobol~Mark, Ma, Finn, Kumar, and Levine]{nakamoto2024cal}
Nakamoto, M., Zhai, S., Singh, A., Sobol~Mark, M., Ma, Y., Finn, C., Kumar, A., and Levine, S.
\newblock Cal-ql: Calibrated offline rl pre-training for efficient online fine-tuning.
\newblock \emph{Advances in Neural Information Processing Systems}, 36, 2024.

\bibitem[Nekoei et~al.(2021)Nekoei, Badrinaaraayanan, Courville, and Chandar]{nekoei2021continuous}
Nekoei, H., Badrinaaraayanan, A., Courville, A., and Chandar, S.
\newblock Continuous coordination as a realistic scenario for lifelong learning.
\newblock In \emph{International Conference on Machine Learning}, pp.\  8016--8024. PMLR, 2021.

\bibitem[{NetHack DevTeam}(1987)]{NetHack}
{NetHack DevTeam}.
\newblock {NetHack Home Page}.
\newblock \url{https://nethackwiki.com/wiki/DevTeam}, 1987.
\newblock Accessed: 2023-05-04.

\bibitem[Neyshabur et~al.(2020)Neyshabur, Sedghi, and Zhang]{neyshabur2020being}
Neyshabur, B., Sedghi, H., and Zhang, C.
\newblock What is being transferred in transfer learning?
\newblock \emph{Advances in neural information processing systems}, 33:\penalty0 512--523, 2020.

\bibitem[Ostapenko et~al.(2021)Ostapenko, Rodriguez, Caccia, and Charlin]{ostapenko2021continual}
Ostapenko, O., Rodriguez, P., Caccia, M., and Charlin, L.
\newblock Continual learning via local module composition.
\newblock \emph{Advances in Neural Information Processing Systems}, 34:\penalty0 30298--30312, 2021.

\bibitem[Pardo et~al.(2017)Pardo, Tavakoli, Levdik, and Kormushev]{Pardo2017TimeLI}
Pardo, F., Tavakoli, A., Levdik, V., and Kormushev, P.
\newblock Time limits in reinforcement learning.
\newblock In \emph{International Conference on Machine Learning}, 2017.

\bibitem[Parisotto et~al.(2015)Parisotto, Ba, and Salakhutdinov]{parisotto2015actor}
Parisotto, E., Ba, J.~L., and Salakhutdinov, R.
\newblock Actor-mimic: Deep multitask and transfer reinforcement learning.
\newblock \emph{arXiv preprint arXiv:1511.06342}, 2015.

\bibitem[Pertsch et~al.(2021)Pertsch, Lee, Wu, and Lim]{pertsch2021guided}
Pertsch, K., Lee, Y., Wu, Y., and Lim, J.~J.
\newblock Guided reinforcement learning with learned skills.
\newblock \emph{arXiv preprint arXiv:2107.10253}, 2021.

\bibitem[Petrenko et~al.(2020)Petrenko, Huang, Kumar, Sukhatme, and Koltun]{Petrenko2020SampleFE}
Petrenko, A., Huang, Z., Kumar, T., Sukhatme, G.~S., and Koltun, V.
\newblock Sample factory: Egocentric 3d control from pixels at 100000 fps with asynchronous reinforcement learning.
\newblock \emph{ArXiv}, abs/2006.11751, 2020.

\bibitem[Piterbarg et~al.(2023)Piterbarg, Pinto, and Fergus]{piterbarg2023nethack}
Piterbarg, U., Pinto, L., and Fergus, R.
\newblock Nethack is hard to hack.
\newblock \emph{arXiv preprint arXiv:2305.19240}, 2023.

\bibitem[Powers et~al.(2022)Powers, Xing, Kolve, Mottaghi, and Gupta]{powers2022cora}
Powers, S., Xing, E., Kolve, E., Mottaghi, R., and Gupta, A.
\newblock Cora: Benchmarks, baselines, and metrics as a platform for continual reinforcement learning agents.
\newblock In \emph{Conference on Lifelong Learning Agents}, pp.\  705--743. PMLR, 2022.

\bibitem[Radford et~al.(2018)Radford, Narasimhan, Salimans, Sutskever, et~al.]{radford2018improving}
Radford, A., Narasimhan, K., Salimans, T., Sutskever, I., et~al.
\newblock Improving language understanding by generative pre-training.
\newblock 2018.

\bibitem[Ramasesh et~al.(2020)Ramasesh, Dyer, and Raghu]{ramasesh2020anatomy}
Ramasesh, V.~V., Dyer, E., and Raghu, M.
\newblock Anatomy of catastrophic forgetting: Hidden representations and task semantics.
\newblock \emph{arXiv preprint arXiv:2007.07400}, 2020.

\bibitem[Ramasesh et~al.(2022)Ramasesh, Lewkowycz, and Dyer]{ramasesh2022effect}
Ramasesh, V.~V., Lewkowycz, A., and Dyer, E.
\newblock Effect of scale on catastrophic forgetting in neural networks.
\newblock In \emph{International Conference on Learning Representations}, 2022.

\bibitem[Rebuffi et~al.(2017)Rebuffi, Kolesnikov, Sperl, and Lampert]{rebuffi2017icarl}
Rebuffi, S.-A., Kolesnikov, A., Sperl, G., and Lampert, C.~H.
\newblock icarl: Incremental classifier and representation learning.
\newblock In \emph{Proceedings of the IEEE conference on Computer Vision and Pattern Recognition}, pp.\  2001--2010, 2017.

\bibitem[Rolnick et~al.(2019)Rolnick, Ahuja, Schwarz, Lillicrap, and Wayne]{rolnick2019experience}
Rolnick, D., Ahuja, A., Schwarz, J., Lillicrap, T., and Wayne, G.
\newblock Experience replay for continual learning.
\newblock \emph{Advances in Neural Information Processing Systems}, 32, 2019.

\bibitem[Ross \& Bagnell(2010)Ross and Bagnell]{ross2010efficient}
Ross, S. and Bagnell, D.
\newblock Efficient reductions for imitation learning.
\newblock In \emph{Proceedings of the thirteenth international conference on artificial intelligence and statistics}, pp.\  661--668. JMLR Workshop and Conference Proceedings, 2010.

\bibitem[Rusu et~al.(2016)Rusu, Rabinowitz, Desjardins, Soyer, Kirkpatrick, Kavukcuoglu, Pascanu, and Hadsell]{rusu2016progressive}
Rusu, A.~A., Rabinowitz, N.~C., Desjardins, G., Soyer, H., Kirkpatrick, J., Kavukcuoglu, K., Pascanu, R., and Hadsell, R.
\newblock Progressive neural networks.
\newblock \emph{arXiv preprint arXiv:1606.04671}, 2016.

\bibitem[Rusu et~al.(2022)Rusu, Flennerhag, Rao, Pascanu, and Hadsell]{rusu2022probing}
Rusu, A.~A., Flennerhag, S., Rao, D., Pascanu, R., and Hadsell, R.
\newblock Probing transfer in deep reinforcement learning without task engineering.
\newblock In \emph{Conference on Lifelong Learning Agents}, pp.\  1231--1254. PMLR, 2022.

\bibitem[Sandler et~al.(2022)Sandler, Zhmoginov, Vladymyrov, and Jackson]{sandler2022fine}
Sandler, M., Zhmoginov, A., Vladymyrov, M., and Jackson, A.
\newblock Fine-tuning image transformers using learnable memory.
\newblock In \emph{Proceedings of the IEEE/CVF Conference on Computer Vision and Pattern Recognition}, pp.\  12155--12164, 2022.

\bibitem[Schaul et~al.(2019)Schaul, Borsa, Modayil, and Pascanu]{schaul2019ray}
Schaul, T., Borsa, D., Modayil, J., and Pascanu, R.
\newblock Ray interference: a source of plateaus in deep reinforcement learning, 2019.

\bibitem[Schmitt et~al.(2018)Schmitt, Hudson, Zidek, Osindero, Doersch, Czarnecki, Leibo, Kuttler, Zisserman, Simonyan, et~al.]{schmitt2018kickstarting}
Schmitt, S., Hudson, J.~J., Zidek, A., Osindero, S., Doersch, C., Czarnecki, W.~M., Leibo, J.~Z., Kuttler, H., Zisserman, A., Simonyan, K., et~al.
\newblock Kickstarting deep reinforcement learning.
\newblock \emph{arXiv preprint arXiv:1803.03835}, 2018.

\bibitem[Schulman et~al.(2017)Schulman, Wolski, Dhariwal, Radford, and Klimov]{schulman2017proximal}
Schulman, J., Wolski, F., Dhariwal, P., Radford, A., and Klimov, O.
\newblock Proximal policy optimization algorithms.
\newblock \emph{arXiv preprint arXiv:1707.06347}, 2017.

\bibitem[Schwarzer et~al.(2021)Schwarzer, Rajkumar, Noukhovitch, Anand, Charlin, Hjelm, Bachman, and Courville]{schwarzer2021pretraining}
Schwarzer, M., Rajkumar, N., Noukhovitch, M., Anand, A., Charlin, L., Hjelm, R.~D., Bachman, P., and Courville, A.~C.
\newblock Pretraining representations for data-efficient reinforcement learning.
\newblock \emph{Advances in Neural Information Processing Systems}, 34:\penalty0 12686--12699, 2021.

\bibitem[Seo et~al.(2022)Seo, Lee, James, and Abbeel]{seo2022reinforcement}
Seo, Y., Lee, K., James, S.~L., and Abbeel, P.
\newblock Reinforcement learning with action-free pre-training from videos.
\newblock In \emph{International Conference on Machine Learning}, pp.\  19561--19579. PMLR, 2022.

\bibitem[Stooke et~al.(2021)Stooke, Lee, Abbeel, and Laskin]{stooke2021decoupling}
Stooke, A., Lee, K., Abbeel, P., and Laskin, M.
\newblock Decoupling representation learning from reinforcement learning.
\newblock In \emph{International Conference on Machine Learning}, pp.\  9870--9879. PMLR, 2021.

\bibitem[Sutton \& Barto(2018)Sutton and Barto]{sutton2018reinforcement}
Sutton, R.~S. and Barto, A.~G.
\newblock \emph{Reinforcement learning: An introduction}.
\newblock MIT press, 2018.

\bibitem[Traor{\'e} et~al.(2019)Traor{\'e}, Caselles-Dupr{\'e}, Lesort, Sun, Cai, D{\'\i}az-Rodr{\'\i}guez, and Filliat]{traore2019discorl}
Traor{\'e}, R., Caselles-Dupr{\'e}, H., Lesort, T., Sun, T., Cai, G., D{\'\i}az-Rodr{\'\i}guez, N., and Filliat, D.
\newblock Discorl: Continual reinforcement learning via policy distillation.
\newblock \emph{arXiv preprint arXiv:1907.05855}, 2019.

\bibitem[Tuyls et~al.(2023)Tuyls, Madeka, Torkkola, Foster, Narasimhan, and Kakade]{tuyls2023scaling}
Tuyls, J., Madeka, D., Torkkola, K., Foster, D., Narasimhan, K., and Kakade, S.
\newblock Scaling laws for imitation learning in nethack.
\newblock \emph{arXiv preprint arXiv:2307.09423}, 2023.

\bibitem[Veniat et~al.(2021)Veniat, Denoyer, and Ranzato]{veniat2021efficient}
Veniat, T., Denoyer, L., and Ranzato, M.
\newblock Efficient continual learning with modular networks and task-driven priors.
\newblock In \emph{9th International Conference on Learning Representations, ICLR 2021}, 2021.

\bibitem[Vinyals et~al.(2019)Vinyals, Babuschkin, Czarnecki, Mathieu, Dudzik, Chung, Choi, Powell, Ewalds, Georgiev, et~al.]{vinyals2019grandmaster}
Vinyals, O., Babuschkin, I., Czarnecki, W.~M., Mathieu, M., Dudzik, A., Chung, J., Choi, D.~H., Powell, R., Ewalds, T., Georgiev, P., et~al.
\newblock Grandmaster level in starcraft ii using multi-agent reinforcement learning.
\newblock \emph{Nature}, 575\penalty0 (7782):\penalty0 350--354, 2019.

\bibitem[Williams(1992)]{williams1992simple}
Williams, R.~J.
\newblock Simple statistical gradient-following algorithms for connectionist reinforcement learning.
\newblock \emph{Reinforcement learning}, pp.\  5--32, 1992.

\bibitem[Wo{\l}czyk et~al.(2021)Wo{\l}czyk, Zaj{\k{a}}c, Pascanu, Kuci{\'n}ski, and Mi{\l}o{\'s}]{wolczyk2021continual}
Wo{\l}czyk, M., Zaj{\k{a}}c, M., Pascanu, R., Kuci{\'n}ski, {\L}., and Mi{\l}o{\'s}, P.
\newblock Continual world: A robotic benchmark for continual reinforcement learning.
\newblock \emph{Advances in Neural Information Processing Systems}, 34:\penalty0 28496--28510, 2021.

\bibitem[Wolczyk et~al.(2022)Wolczyk, Zaj{\k{a}}c, Pascanu, Kuci{\'n}ski, and Mi{\l}o{\'s}]{wolczykdisentangling}
Wolczyk, M., Zaj{\k{a}}c, M., Pascanu, R., Kuci{\'n}ski, {\L}., and Mi{\l}o{\'s}, P.
\newblock Disentangling transfer in continual reinforcement learning.
\newblock In \emph{Advances in Neural Information Processing Systems}, 2022.

\bibitem[Wulfmeier et~al.(2023)Wulfmeier, Byravan, Bechtle, Hausman, and Heess]{wulfmeier2023foundations}
Wulfmeier, M., Byravan, A., Bechtle, S., Hausman, K., and Heess, N.
\newblock Foundations for transfer in reinforcement learning: A taxonomy of knowledge modalities, 2023.

\bibitem[Xu et~al.(2023)Xu, Xie, Qin, Tao, and Wang]{xu2023parameter}
Xu, L., Xie, H., Qin, S.-Z.~J., Tao, X., and Wang, F.~L.
\newblock Parameter-efficient fine-tuning methods for pretrained language models: A critical review and assessment.
\newblock \emph{arXiv preprint arXiv:2312.12148}, 2023.

\bibitem[Yang et~al.(2023)Yang, Yong, Ma, Hu, Zhang, and Zhang]{yang2023essential}
Yang, R., Yong, L., Ma, X., Hu, H., Zhang, C., and Zhang, T.
\newblock What is essential for unseen goal generalization of offline goal-conditioned rl?
\newblock In \emph{International Conference on Machine Learning}, pp.\  39543--39571. PMLR, 2023.

\bibitem[Yosinski et~al.(2014)Yosinski, Clune, Bengio, and Lipson]{yosinski2014transferable}
Yosinski, J., Clune, J., Bengio, Y., and Lipson, H.
\newblock How transferable are features in deep neural networks?
\newblock \emph{Advances in neural information processing systems}, 27, 2014.

\bibitem[Yu et~al.(2020)Yu, Quillen, He, Julian, Hausman, Finn, and Levine]{yu2020meta}
Yu, T., Quillen, D., He, Z., Julian, R., Hausman, K., Finn, C., and Levine, S.
\newblock Meta-world: A benchmark and evaluation for multi-task and meta reinforcement learning.
\newblock In \emph{Conference on robot learning}, pp.\  1094--1100. PMLR, 2020.

\bibitem[Zhang et~al.(2022)Zhang, Park, Han, Qin, Gulati, Shor, Jansen, Xu, Huang, Wang, et~al.]{zhang2022bigssl}
Zhang, Y., Park, D.~S., Han, W., Qin, J., Gulati, A., Shor, J., Jansen, A., Xu, Y., Huang, Y., Wang, S., et~al.
\newblock Bigssl: Exploring the frontier of large-scale semi-supervised learning for automatic speech recognition.
\newblock \emph{IEEE Journal of Selected Topics in Signal Processing}, 16\penalty0 (6):\penalty0 1519--1532, 2022.

\bibitem[Zheng et~al.(2023)Zheng, Luo, Wei, Song, Li, and Jiang]{zheng2023adaptive}
Zheng, H., Luo, X., Wei, P., Song, X., Li, D., and Jiang, J.
\newblock Adaptive policy learning for offline-to-online reinforcement learning.
\newblock \emph{arXiv preprint arXiv:2303.07693}, 2023.

\end{thebibliography}
\bibliographystyle{icml2024}

\newpage
\appendix
\onecolumn
\newpage{}
\newcommand{\mute}[1]{#1}
\section{Toy Examples -- MDP and AppleRetrieval}\label{app:toy_mdp}

In the main text, we showed empirically that \problem{} appears in standard RL scenarios. Here, we additionally provide two toy environments: two-state MDPs and a simple grid-world called \textsc{AppleRetrieval}. We find these environments to be helpful for understanding the core of the problem and for building intuition.

\renewcommand\thesubfigure{(\alph{subfigure})}
\captionsetup[sub]{
  labelformat=simple
}

\begin{figure}[t]
  \centering
  \begin{subfigure}[t]{0.4\textwidth} 
    \centering
    \resizebox{\textwidth}{!}{
        \begin{tikzpicture}[node distance=1.5cm,auto]
        \node[state] (s_0) {$s_0$};
        \node[state] (s_1) [right of=s_0] {$s_1$};
        
        \path[->] (s_0) edge [loop left] node {$r_0, (1-\theta)$} ()
                        edge [bend left] node {$1, \theta$} (s_1);
        \path[->] (s_1) edge [loop right] node {$1, f_\theta$} ()
                        edge [bend left] node {$r_1, (1-f_\theta$)} (s_0);
        \end{tikzpicture}
    }
    \caption{\small{MDP with two states with $\gamma=0.9$.} 
    }
    \label{fig:mdp_graph}
  \end{subfigure} \\
  \hfill
  \begin{subfigure}[t]{0.4\textwidth}z
   \mute{\includegraphics[width=\textwidth]{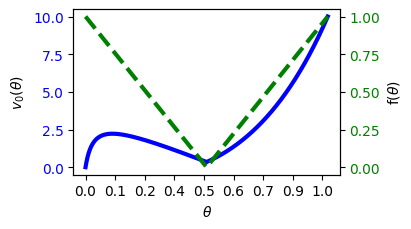}}
    \caption{\small{Value of $s_0$ (blue) and $f_\theta$ (green) 
    for $r_0=0$, $r_1=-1$, 
    \diff{ and $f_\theta=2|\theta-0.5|$.}}
    }
    \label{fig:mdp_dist_shift}
  \end{subfigure}
  \hfill
  \begin{subfigure}[t]{0.4\textwidth}
    \includegraphics[width=\textwidth]{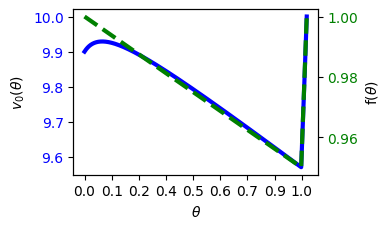}
    \caption{\small{Value of $s_0$ (blue) and $f_\theta$ (green)
    for $r_0=0.99$, $r_1=0$, $\epsilon = 0.05$, 
    \diff{$f_\theta = 1-\frac{\epsilon\theta}{1-\epsilon/2}$ for $\theta \le 1 - \epsilon/2$ 
    and $f_\theta= 2\theta - 1$ for $\theta > 1 - \epsilon/2$.}}
    }
    \label{fig:mdp_e_close}
  \end{subfigure}
  \hfill
    
  %
  %
  \caption{(a) A toy two-state MDP. Each arrow depicts a transition between states, and the annotation encodes the reward and the probability of transition from the policy. (b,c) A policy with its corresponding value function $v_0(\theta)$, for two variants of parameterization and reward functions.}
  \label{fig:example_figure}
\end{figure}
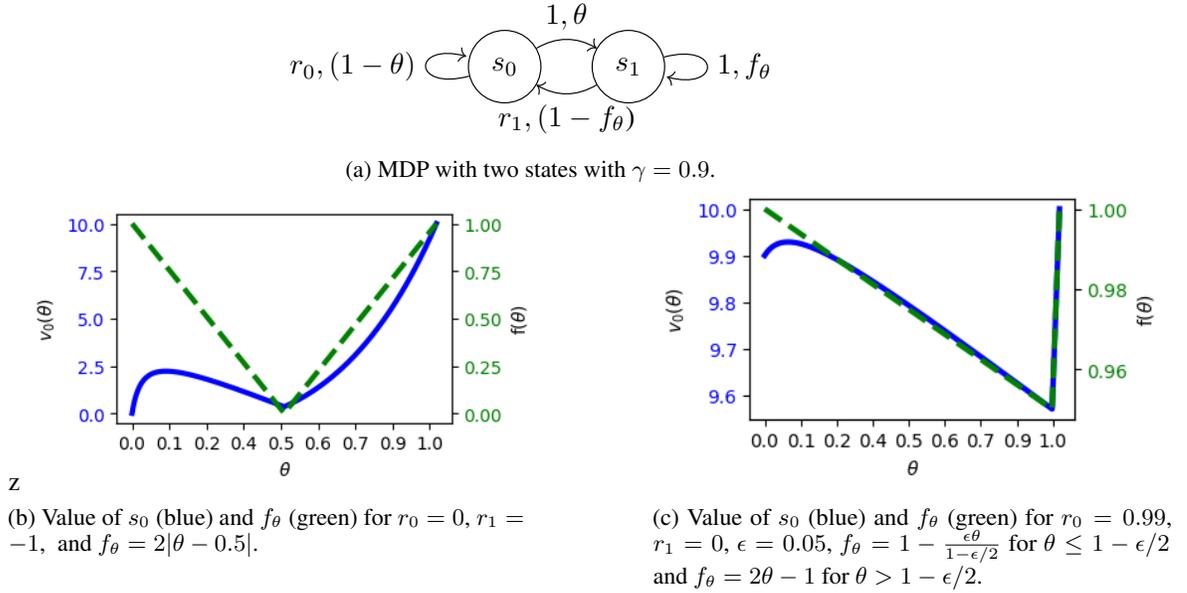

\subsection{Two-state MDPs}

In this subsection, we show that the two scenarios of \problem{}, \envshift{} and \imitationgap{} can happen even in a very simple $2$-state MDP.
This observation fits well into the RL tradition of showing counterexamples on small MDPs \citep{sutton2018reinforcement}.
The MDP, shown in Figure \ref{fig:mdp_graph}, consists of two states, labeled as $s_0$ and $s_1$.
The transition between states is stochastic and is indicated by an arrow annotated by a reward and transition probability. For example, a transition from $s_1$ to $s_0$ happens with probability $1-f_\theta$ and grants a reward $r_1$.
The value of state $s_0$, visualized as a blue line in Figure \ref{fig:mdp_dist_shift} and \ref{fig:mdp_e_close}, equals
\[
v_0(\theta) = \frac{1}{1-\gamma}\frac{\theta + r_0(1-\theta)(1-\gamma f_\theta) + \gamma\theta r_1(1-f_\theta)}{1 - \gamma f_\theta + \gamma \theta}.
\]
In each case, we treat fine-tuning as the process of adjusting $\theta$ towards the gradient direction of $v_0(\theta)$ until a local extremum is encountered. We now consider two parameterizations of this MDP that represent \envshift{} and \imitationgap{}.

\paragraph{\envshiftUpper{}} In Figure~\ref{fig:mdp_dist_shift}, we present a \envshift{} scenario, where we fine-tune a policy that was pre-trained on a subset of downstream states and we show that it can lead to divergence.  We parameterize the policy as:

\begin{equation*}
    f_\theta = \left(\frac{-\epsilon}{1-\epsilon/2}\theta + 1\right) \mathbf{1}_{\theta \le 1 - \epsilon/2} + \left(2\theta - 1\right)\mathbf{1}_{\theta > 1 - \epsilon/2}.
\end{equation*}

Here, we have an MDP where the initial policy $\theta=0$ was trained only on state $s_1$. Since $f_0=1$, such a policy stays in $s_1$ once it starts from $s_1$. 

If we now try to fine-tune this policy where the starting state is $s_0$, 
the agent will forget the behavior in $s_1$ due to the interference caused by the parametrization of the policy. This in turn will lead the system to converge to a suboptimal policy $\theta=0.11$ with a value of $2.22$. In this case, the environment has changed by introducing new states that need to be traversed to reach states on which we know how to behave.  
Learning on these new states that are visited early on will lead to forgetting of the pre-trained behavior.

\paragraph{\imitationgapUpper{}} Subsequently, in Figure~\ref{fig:mdp_e_close}, we provide an example of \imitationgap{}. The policy is parametrized as 

\begin{equation*}
    f_\theta=2|\theta-0.5|.
\end{equation*}

In this scenario, $\theta=1$ (with $f_1=1$) represents the optimal behavior of staying in $s_1$ and achieving maximum total discounted returns equal to $10$. However, for a given parametrization of $f_\theta$, this maximum can be unstable, and adding a small noise $\epsilon$ to $\theta$ before fine-tuning will lead to divergence towards a local maximum at $\theta=0.08$ with the corresponding value $9.93$. 
Perturbing $\theta$ by $\epsilon$ will make the system visit $s_0$ more often, and learning on $s_0$ with further push $\theta$ away from 1, forgetting the skill of \emph{moving to and staying in $s_1$}.

\subsection{Synthetic example: \textsc{AppleRetrieval}} \label{sec:toy_example}

\begin{wrapfigure}{R}{0.45\linewidth}
    \vspace{-4.75em}
    \centering
    \mute{\includegraphics[width=0.95\linewidth]{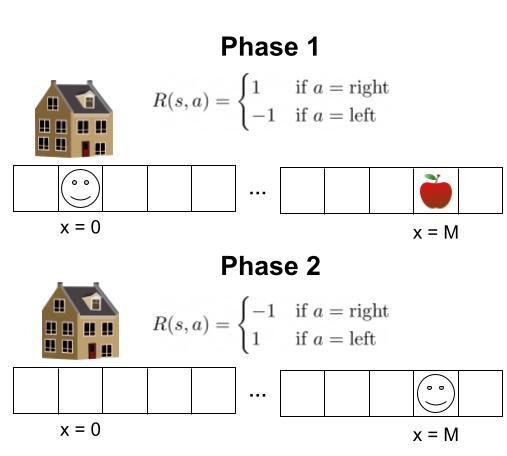}}
    \caption*{\small \textsc{AppleRetrieval} environment.}
    \label{fig:appleenv}
    \vspace{-1.5em}
\end{wrapfigure}

Additionally, we introduce a synthetic example of an environment exhibiting \envshift{}, dubbed $\textsc{AppleRetrieval}$. We will show that even a vanilla RL algorithm with linear function approximators shows \problem{}. 

$\textsc{AppleRetrieval}$ is a $1$D gridworld, consisting of two phases. In Phase 1, starting at home: $x=0$, the agent has to go to $x=M$ and retrieve an apple, $M \in \mathbb{N}$. In Phase 2, the agent has to go back to $x=0$. In each phase, the reward is $1$ for going in the correct direction and $-1$ otherwise. The observation is $o = [-c]$ in Phase 1 and $o = [c]$ in Phase 2, for some $c\in \mathbb{R}$; i.e. it encodes the information about the current phase. {Given this observation}, it is now trivial to encode the optimal policy: go right in Phase 1 and go left in Phase 2. Episodes are terminated if the solution is reached or after $100$ timesteps. Since we can only get to Phase 2 by completing Phase 1, this corresponds to dividing the states to sets \textsc{Close} and \textsc{Far}, as described in Section~\ref{sec:preliminaries}.


We run experiments in \textsc{AppleRetrieval} using the REINFORCE algorithm~\citep{williams1992simple} and assume a simple model in which the probability to move right is given by: $\pi_{w,b}(o) = \sigma(w\cdot o + b), w, b\in \mathbb{R}$. Importantly, we initialize $w,b$  with the weights trained in Phase 2. 

\begin{figure}
    \centering
    \mute{\includegraphics[width=\textwidth]{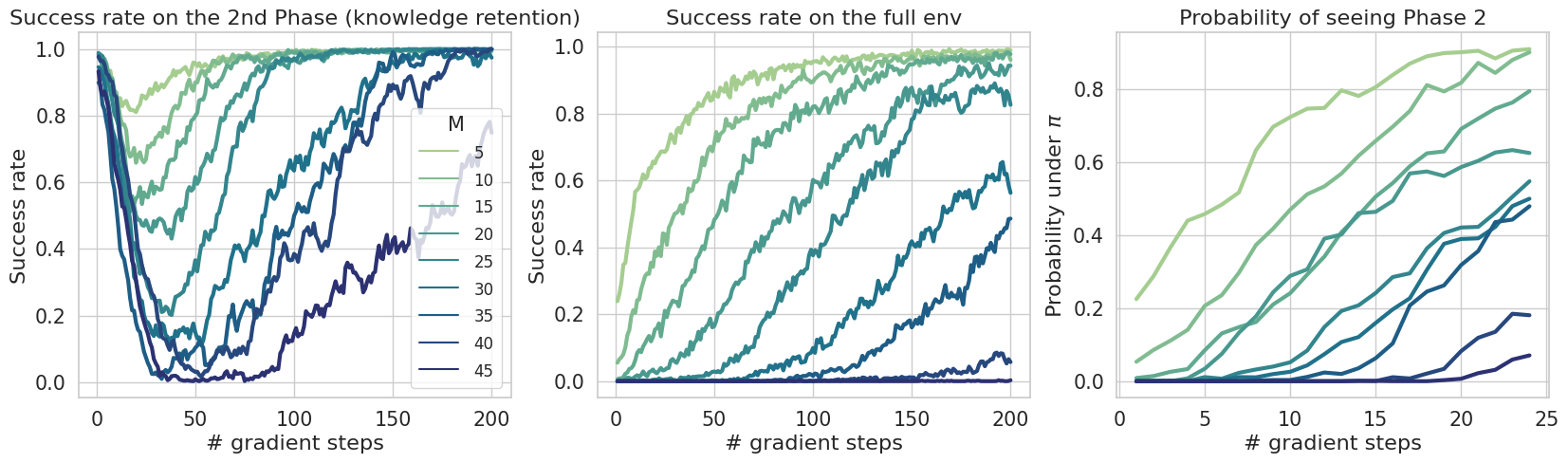}}
    \caption{\small \problemUpper{} in \textsc{AppleRetrieval}. (Left) Forgetting becomes more problematic as $M$ (the distance from the house to the apple) increases and (center) hinders the overall performance. (Right, note x-scale change) This happens since the probability of reaching Phase 2 in early training decreases. 
    }
    \label{fig:apple_results}
    
\end{figure}

\begin{figure}
    \centering
    \centering
    \mute{\includegraphics[width=\textwidth]{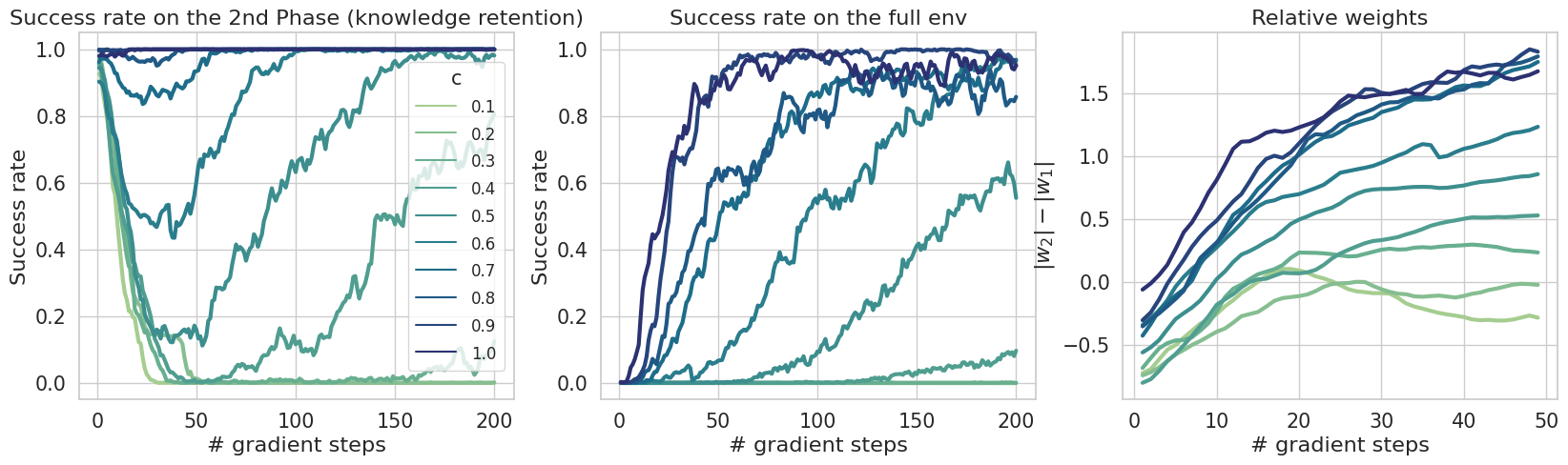}}

    \caption{Impact of $c$ on the results for set $M=30$. For smaller $c$ forgetting (left) is greater and the overall success rate is smaller (center) since it encourages the pre-trained model to find solutions with a high $\frac{|b|}{|w|}$ ratio, as confirmed by looking at weight difference early in fine-tuning (right).} 
    \label{fig:apple_results_appendix}
\end{figure}

We show experimentally, see Figure~\ref{fig:apple_results}, that for high enough distance $M$, the \problem{} problem appears. Intuitively, the probability of concluding Phase 1 becomes small enough that the pre-trained Phase 2 policy is forgotten, leading to overall poor performance. In this simple case, we can mechanically analyze this process of forgetting.

Since the linear model in \textsc{AppleRetrieval} has only two parameters (weight $w$, bias $b$) we can analyze and understand what parameter sets lead to forgetting. If the pre-trained policy mostly relies on weight (i.e. $|w| \gg |b|$) then the interference will be limited. However, if the model relies on bias (i.e. $|b| \gg |w|$) then interference will occur as bias will impact the output in the same way in both phases. We can guide the model towards focusing on one or the other by setting the $c$ parameter since the linear model trained with gradient descent will tend towards a solution with a low weight norm. The results presented in Figure~\ref{fig:apple_results_appendix} confirm our hypothesis, as lower values of $c$ encourage models to rely more on $b$ which leads to forgetting. Such a low-level analysis is infeasible for deep neural networks, but experimental results confirm that interference occurs in practice~\citep{kirkpatrick2017overcoming,kemker2018measuring,ramasesh2022effect}.

\section{Technical details}
\label{app:technical}

\subsection{NetHack}
\label{app:nethack}

\paragraph{Environment} NetHack~\cite{NetHack} is a classic and highly complex terminal roguelike game that immerses players in a procedurally generated dungeon crawling experience, navigating through a labyrinth in a world filled with monsters, treasures, and challenges. The NetHack Learning Environment (NLE) introduced in~\cite{kuttler2020nethack} is a scalable, procedurally generated, stochastic, rich, and challenging environment aimed to drive long-term research on problems such as exploration, planning, skill acquisition, and language-conditioned RL.

The NLE is characterized by a state space that includes a 2D grid representing the game map and additional information like the player's inventory, health, and other statistics. Thus, the NLE is multimodal and consists of an image, the main map screen, and text.
The action space in NLE consists of a set of 120 discrete actions.
At the same time, the NLE presents a challenge for RL agents due to its action-chaining behavior. For instance, the player must press three distinct keys in a specific sequence to throw an item, which creates additional complexity to the RL problem. 
The environmental reward in \verb|score| task, used in this paper, is based on the increase in the in-game score between two-time steps. A complex calculation determines the in-game score. However, during the game's early stages, the score is primarily influenced by factors such as killing monsters and the number of dungeon levels the agent explores. The in-game score is a sensible proxy for incremental progress on NLE. Still, training agents to maximize it is likely not perfectly aligned with solving the game, as expert human players can solve NetHack while keeping the score low. In each run, the dungeon is generated anew, so the agent does not ever see a specific level twice. Consequently, we can't expect the agent to remember solutions to specific levels, but rather, we aim for it to recall general behavioral patterns for different levels.

It is important to note that during training, the agent may not follow levels in a linear sequence due to NetHack's allowance for backtracking or branching to different dungeon parts (as described in \url{https://nethackwiki.com/wiki/Branch}). This highlights the issue of forgetting, even in the absence of strictly defined linear tasks or stages, contrary to the continual learning literature.


\paragraph{Architecture} 
We fine-tune the model pre-trained by~\citet{tuyls2023scaling}, which scales up (from 6M to 33M parameters) and modifies the solution proposed by the 'Chaotic Dwarven GPT-5' team, which is based on Sample Factory~\cite{Petrenko2020SampleFE} that was also used in~\cite{hambrodungeons}. 
This model utilizes an LSTM architecture that incorporates representations from three encoders, which take observations as inputs. The LSTM network's output is then fed into two separate heads: a policy head and a baseline head. The model architecture used both in online and offline settings consists of a joint backbone for both actor and critic. It takes as an input three components: main observation of the dungeon screen, \verb|blstats|, and \verb|message|. \verb|blstats| refers to the player's status information, such as health and hunger, and \verb|message| refers to the textual information displayed to the player, such as notifications and warnings. \verb|blstats| and \verb|message| are processed using two layer MLP. The main observation of the dungeon screen is processed by embedding each character and color in an embedding lookup table which is later put into a grid processed by ResNet. For more details about Main screen encoder refer to~\cite{tuyls2023scaling}. The components are encoded, and are merged before passing to LSTM. This baseline allows for fast training
but struggles with learning complex behaviours required for certain roles in the game. 
More details about the architecture can be found in~\cite{tuyls2023scaling,Petrenko2020SampleFE}. The model hyperparameters are shown in Table~\ref{tab:hparams_nle} -- analogical to Table 6 from~\cite{Petrenko2020SampleFE}.

\paragraph{Dataset} The knowledge retention methods presented in this paper use a subset of the NetHack Learning Dataset (NLD) collected by~\cite{hambrodungeons} called NLD-AA. It contains over 3 billion state-action-score transitions and metadata from 100,000 games collected from the winning bot of the NetHack Challenge~\cite{hambro2022insights}.
In particular, we use about 8000 games of Human Monk. This character was chosen because it was extensively evaluated in the previous work \cite{hambrodungeons} and because the game setup for the Human Monk is relatively straightforward, as it does not require the agent to manage the inventory. 
The bot is based on the 'AutoAscend' team solution, a symbolic agent that leverages human knowledge and hand-crafted heuristics to progress in the game. Its decision-making module is based on a behavior tree model.

The checkpoint we use as the pre-trained policy $\pi_*$ was trained by~\citet{tuyls2023scaling} on a larger set of trajectories from the AutoAscend agent, containing over $115B$ transitions.

\paragraph{Pre-training}
As for the offline pre-training phase, we used a model trained with Behavioral Cloning (BC)~\cite{bain1995framework,ross2010efficient} by~\cite{tuyls2023scaling}, an imitation learning approach that utilizes a supervised learning objective to train the policy to mimic the actions present in the dataset. To be more specific, it utilizes a cross-entropy loss function between the policy action distribution and the actions from the NLD-AA dataset. 
For more details on hyperparameters, please refer to the original article~\cite{tuyls2023scaling}. It should be noted that BC does not include a critic. To improve stability during the beginning of the fine-tuning we additionally pre-train the baseline head by freezing the rest of the model for 500M environment steps.

\paragraph{Fine-tuning}
In the online training phase, we employed a highly parallelizable architecture called Asynchronous Proximal Policy Optimization (APPO)~\cite{schulman2017proximal, Petrenko2020SampleFE}. In this setup, we can run over $500$ million environment steps under $24$ hours of training on A100 Nvidia GPU.
%
 %
Within the main manuscript, we examined vanilla fine-tuning and fine-tuning with a behavioral cloning loss, kickstarting and EWC, explained in more detail in Appendix~\ref{app:cl_methods}. 

In Fine-tuning + KS we compute the auxiliary loss on data generated by the online policy. We scaled the loss by a factor of $0.5$ and used exponential decay $0.99998$, where the coefficient was decayed every train step. 
In Fine-tuning + BC we compute the auxiliary loss by utilizing the trajectories generated by the expert (i.e. the AutoAscend algorithm), note that no decay was used here. 
We scaled the auxiliary loss by a factor of $2.0$. 
To improve the stability of the models we froze the encoders during the course of the training. Additionally, we turn off entropy when employing knowledge retention methods in similar fashion to \cite{baker2022video}. For EWC we use a regularization coefficient of $2 \cdot 10^6$.

\diff{\paragraph{Evaluation}
During the evaluation phase, we provide the in-game score achieved and the number of filled pits for Sokoban levels at specific checkpoints during training. Models were evaluated every 25 million environment steps for Figure~\ref{fig:nethack_per_level}.
To perform the per-level evaluation in Figure~\ref{fig:nethack_per_level}, we employ the AutoAscend expert, used for behavioral cloning in pre-training. We use AutoAscend to play the game and save the state when it reaches the desired level. We generate 200 game saves for each level and evaluate our agents on each save by loading the game, running our agent where the expert finished, and reporting the score our agent achieved on top of the expert's score.
}

\begin{table}[htbp]
\centering
\caption{Hyperparameters of the model used in NLE. For the most part, we use hyperparameters values from \cite{hambrodungeons}.}
\label{tab:hparams_nle}
\begin{tabular}{|c|c|}
\hline
Hyperparameter Name & Value \\
\hline
activation\_function & relu \\
adam\_beta1 & 0.9 \\
adam\_beta2 & 0.999 \\
adam\_eps & 0.0000001 \\
adam\_learning\_rate & 0.0001 \\
weight\_decay & 0.0001 \\
appo\_clip\_policy & 0.1 \\
appo\_clip\_baseline & 1.0 \\
baseline\_cost & 1 \\
discounting & 0.999999 \\
entropy\_cost & 0.001 \\
grad\_norm\_clipping & 4 \\
hidden\_dim & 1738 \\
batch\_size & 128 \\
penalty\_step & 0.0 \\
penalty\_time & 0.0 \\
reward\_clip & 10 \\
reward\_scale & 1 \\
unroll\_length & 32 \\
\hline
\end{tabular}
\end{table}

\subsection{Montezuma's Revenge}
\label{app:montezuma}

\paragraph{Environment} In this section, we provide further details on our experiments with Montezuma's Revenge from Atari Learning Environment (ALE) \cite{machado18arcade}. 
Montezuma's Revenge, released in 1984, presents a challenging platformer scenario where players control the adventurer Panama Joe as he navigates a labyrinthine Aztec temple, solving puzzles and avoiding a variety of deadly obstacles and enemies. What makes Montezuma's Revenge particularly interesting for research purposes is its extreme sparsity of rewards, where meaningful positive feedback is rare and often delayed, posing a significant challenge. 

\begin{figure}[H]
    \centering
    \mute{\includegraphics[width=1\textwidth]{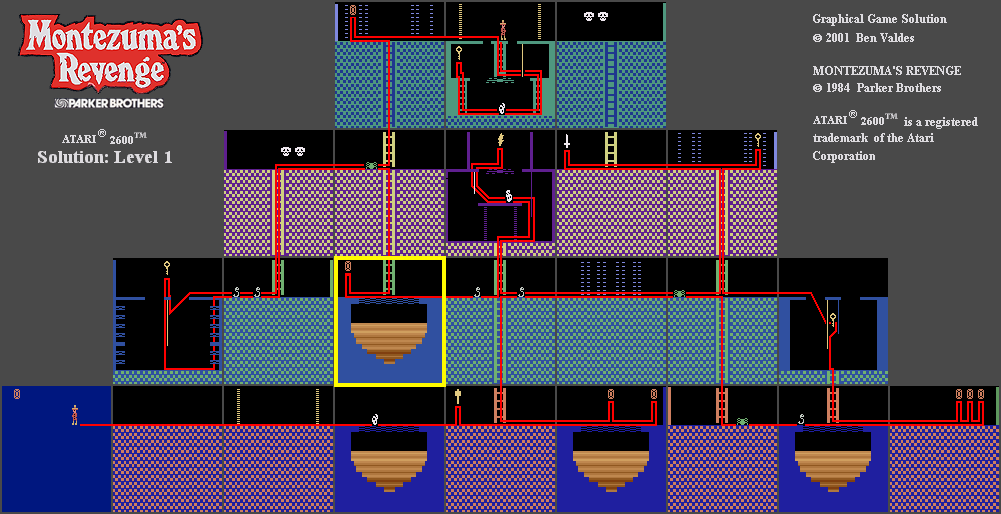}}
    \caption {
    The order in which rooms are visited to complete the first level of Montezuma's Revenge is presented with the red line. We highlight Room 7, which we use for experiments in the mani text, with a yellow border. Source: https://pitfallharry.tripod.com/MapRoom/MontezumasRevengeLvl1.html
    } 
    \label{fig:montezuma_layout}
\end{figure}

We enumerate rooms according to the progression shown in Figure \ref{fig:montezuma_layout}, starting from Room 1, where the player begins gameplay. As a successful completion of the room in Figure \ref{fig:montezuma_room10}, we consider achieving at least one of the following options: either earn a coin as a reward, acquire a new item, or exit the room through a different passage than the one we entered through.

\paragraph{Architecture} In our experiments, we use a PPO agent with a Random Network Distillation (RND) mechanism \cite{burda2018exploration} for exploration boost. It achieves this by employing two neural networks: a randomly initialized target network and a prediction network. Both networks receive observation as an input and return a vector with size 512. The prediction network is trained to predict the random outputs generated by the target network. During interaction with the environment, the prediction network assesses the novelty of states, prioritizing exploration in less predictable regions. States for which the prediction network's predictions deviate significantly from the random targets are considered novel and are prioritized for exploration. Detailed hyperparameter values can be found in Table \ref{tab:hparams_montezuma}.

\begin{figure}[H]
    \centering
    \mute{\includegraphics[width=0.6\textwidth]{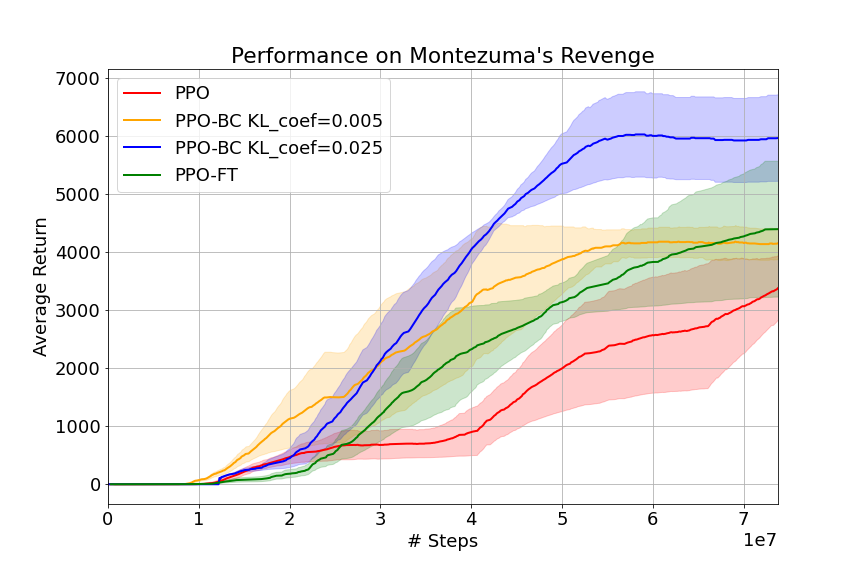}}
    \caption {Average return in Montezuma's Revenge for PPO (trained from scratch), fine-tuned PPO and two different coefficients for fine-tuned PPO + BC.
    } 
    \label{fig:montezuma_2kl}
\end{figure}

\textbf{Dataset} For behavioral cloning purposes, we collected more than 500 trajectories sampled from a pre-trained PPO agent with RND that achieved an episode cumulative reward of around $7000$.
In Figure \ref{fig:montezuma_2kl} we show the impact of different values of the Kullback–Leibler weight coefficient on agent performance.

\begin{table}[htbp]
\centering
\caption{Hyperparameters of the model used in Montezuma's Revenge. For the most part, we use hyperparameter values from \cite{burda2018exploration}. We used PyTorch implementation by \textit{jcwleo} from https://github.com/jcwleo/random-network-distillation-pytorch}
\label{tab:hparams_montezuma}
\begin{tabular}{|c|c|}
\hline
Hyperparameter Name & Value \\
\hline
MaxStepPerEpisode & 4500 \\
ExtCoef & 2.0 \\
LearningRate & 1e-4 \\
NumEnv & 128 \\
NumStep & 128 \\
Gamma & 0.999 \\
IntGamma & 0.99 \\
Lambda & 0.95 \\
StableEps & 1e-8 \\
StateStackSize & 4 \\
PreProcHeight & 84 \\
ProProcWidth & 84 \\
UseGAE & True \\
UseGPU & True \\
UseNorm & False \\
UseNoisyNet & False \\
ClipGradNorm & 0.5 \\
Entropy & 0.001 \\
Epoch & 4 \\
MiniBatch & 4 \\
PPOEps & 0.1 \\
IntCoef & 1.0 \\
StickyAction & True \\
ActionProb & 0.25 \\
UpdateProportion & 0.25 \\
LifeDone & False \\
ObsNormStep & 50 \\
\hline
\end{tabular}
\end{table}

\subsection{Meta World}
\label{app:robotic_details}

In this section, we describe the \roboticsequence{} setting, and we provide more details about its construction. The algorithm representing \roboticsequence{} construction is presented in Algorithm~\ref{alg:stitchedenv}.

We use multi-layer perceptrons (4 hidden layers, 256 neurons each) as function approximators for the policy and $Q$-value function. For all experiments in this section, we use the Soft Actor-Critic (SAC) algorithm \citep{haarnoja2018soft}. The observation space consists of information about the current robot configuration, see ~\citep{yu2020meta} for details, and the stage ID encoded as a one-hot vector. 
In our experiments, we use a pre-trained model that we trained with SAC on the last two stages (\texttt{peg-unplug-side} and \texttt{push-wall}) until convergence (i.e. $100\%$ success rate).
All experiments on Meta-World are run with at least $20$ seeds and we present the results with $90\%$ confidence intervals. The codebase is available in the supplementary materials. 

    \begin{algorithm}[H]
    \caption{\roboticsequence{}}
    \label{alg:stitchedenv}
    \begin{algorithmic}
    \STATE \textbf{Input:} list of $N$ environments $E_k$, policy $\pi$, time limit $T$.
    \STATE \textbf{Returns:} number of solved environments.
    \STATE $i = 1; t = 1$ \COMMENT{Initialize env idx, timestep counter}
    \WHILE{$i \leq N$ \AND $t \leq T$} 
        \STATE Take a step in $E_i$ using $\pi$
        \IF{$E_i$ is solved}
            \STATE $i = i + 1; t = 1$
            \COMMENT{Move to the next env, reset timestep counter }
        \ENDIF
    \ENDWHILE
    \STATE \textbf{Return} $i-1$
    \end{algorithmic}
    \end{algorithm}
    \vspace{-1em}

In order to make the problem more challenging, we randomly sample the start and goal conditions, similarly as in~\cite{wolczyk2021continual}. Additionally, we change the behavior of the terminal states. In the original paper and codebase, the environments are defined to run indefinitely, but during the training, finite trajectories are sampled (i.e. $200$ steps). On the $200$th step even though the trajectory ends, SAC receives information that the environment is still going. Effectively, it means that we still bootstrap our Q-value target as if this state was not terminal. This is a common approach for environments with infinite trajectories~\cite{Pardo2017TimeLI}.

However, this approach is unintuitive from the perspective of \roboticsequence{}. We would like to go from a given stage to the next one at the moment when the success signal appears, without waiting for an arbitrary number of steps. As such, we introduce a change to the environments and terminate the episode in two cases: when the agent succeeds or when the time limit is reached. In both cases, SAC receives a signal that the state was terminal, which means we do not apply bootstrapping in the target Q-value. In order for the MDP to be fully observable, we append the normalized timestep (i.e. the timestep divided by the maximal number of steps in the environment, $T=200$ in our case) to the state vector. Additionally, when the episode ends with success, we provide the agent with the "remaining" reward it would get until the end of the episode. That is, if the last reward was originally $r_t$, the augmented reward is given by $r'_t = \beta r_t(T - t)$. $\beta = 1.5$ is a coefficient to encourage the agent to succeed. Without the augmented reward there is a risk that the policy would avoid succeeding and terminating the episode, in order to get rewards for a longer period of time. 

\paragraph{SAC}

We use the Soft Actor-Critic~\citep{haarnoja2018soft} algorithm for all the experiments on Meta-World and by default use the same architecture as in the Continual World~\citep{wolczyk2021continual} paper, which is a $4$-layer MLP with $256$ neurons each and Leaky-ReLU activations. We apply layer normalization after the first layer. The entropy coefficient is tuned automatically~\citep{haarnoja2018softapp}. We create a separate output head for each stage in the neural networks and then we use the stage ID information to choose the correct head. We found that this approach works better than adding the stage ID to the observation vector.

For the base SAC, we started with the hyperparameters listed in~\cite{wolczyk2021continual} and then performed additional hyperparameter tuning. We set the learning rate to $10^{-3}$ and use the Adam~\cite{kingma2014adam} optimizer. The batch size is $128$ in all experiments. We use EWC, and BC as described in~\cite{wolczyk2021continual,wolczykdisentangling}. For episodic memory, we sample $10$k state-action-reward tuples from the pre-trained stages using the pre-trained policy and we keep them in SAC's replay buffer throughout the training on the downstream task. Since replay buffer is of size $100$k, $10\%$ of the buffer is filled with samples from the prior stages.  For each method, we perform a hyperparameter search on method-specific coefficients. Following~\cite{wolczyk2021continual,wolczykdisentangling} we do not regularize the critic. 
The final hyperparameters are listed in Table~\ref{tab:hyperparams}.

%



\begin{table}[t]
    \centering
    \caption{Hyperparameters of knowledge retention methods in Meta-World experiments.}
    \label{tab:hyperparams}
    \begin{tabular}{c|cccc}
    \toprule
         Method & actor reg. coef. & critic reg. coef. & memory \\
    \midrule
         EWC & 100 & 0 & - \\
         BC & 1 & 0 & 10000 \\ 
         \newdiff{EM} & - & - & 10000 \\
     \bottomrule
    \end{tabular}
\end{table}

\paragraph{CKA} We use Central Kernel Alignment~\cite{kornblith2019similarity} to study similarity of representations. CKA is computed between a pair of matrices, $X \in \mathbb{R}^{n\times p_1}, Y \in \mathbb{R}^{n\times p_2}$, which record, respectively, activations for $p_1$ and $p_2$ neurons for the same $n$ examples. The formula is then given as follows:
\begin{align}
    \text{CKA}(K, L) = \frac{\text{HSIC}(K, L)}{\sqrt{\text{HSIC}(K, K)\text{HSIC}(L, L)}},
\end{align}
where HSIC is the Hilbert-Schmidt Independence Criterion~\cite{gretton2005measuring}, $K_{ij} = k(\mathbf{x}_i, \mathbf{x}_j)$ and $L_{ij} = l(\mathbf{y}_i, \mathbf{y}_j)$, and $k$ and $l$ are two kernels. In our experiments, we simply use a linear kernel in both cases.

\paragraph{Compute} For the experiments based on Meta-World, we use CPU acceleration, as the observations and the networks are relatively small and the gains from GPUs are marginal~\cite{wolczyk2021continual}. For each experiment, we use 8 CPU cores and 30GB RAM. The average length of an experiment is 48 hours. During our research for this paper, we ran over 20,000 experiments on Contiual World.

\section{Knowledge retention methods}
\label{app:cl_methods}

In this section, we provide more details about the knowledge retention methods used in the experiments, and we briefly describe different types of possible approaches.

In this paper, we mostly focus on fine-tuning only on a single stationary task. However, in continual learning literature that often focuses on the problem of mitigating forgetting, the goal is to usually deal with a sequence of tasks (up to several hundred~\cite{lesort2022scaling}) and efficiently accumulate knowledge over the whole sequence. As such, although here we will describe knowledge retention methods with two tasks (corresponding to pre-training and fine-tuning), in practice dealing with a longer sequence of tasks might require more careful considerations.


\subsection{Regularization-based methods}
Regularization-based methods in CL aim to limit forgetting by penalizing changes in parameters that are relevant to the current task. In particular, a few regularization methods~\cite{kirkpatrick2017overcoming,aljundi2018memory} add an auxiliary loss of the following form:
\begin{equation}
    \mathcal L_{aux}(\theta) = \sum_i F^i (\theta_{\text{pre}}^i  - \theta^i )^2,
\end{equation}
where $\theta$ are the weights of the current model, $\theta_{\text{pre}}$ are the weights of a prior model, and $F^i$ are weighting coefficients. 
 In \textbf{Elastic Weight Consolidation (EWC)}~\citep{kirkpatrick2017overcoming} we use in our experiments, $F$ is the diagonal of the Fisher Information Matrix, see~\citep{wolczyk2021continual} for details about its implementation in Soft Actor-Critic. 

\subsection{Distillation-based methods}
\label{app:cl_distillation}
In this work, we use the behavioral cloning approach used previously in continual reinforcement learning setup~\cite{wolczykdisentangling,rolnick2019experience} 
This approach is based on minimizing the Kullback-Leibler of action distributions under particular states $D_{KL}^s\infdivx{p}{q} = \mathbb{E}_{a \sim p(\cdot \vert s)} \left [ \log( \frac{p(a \vert s)}{q(a \vert s)}) \right ]$. Assume that $\pi_\theta$ is the current policy parameterized by $\theta$ (student) and $\pi_*$ is the pre-trained policy (teacher).

In behavioral cloning, we apply the following loss:
\begin{equation}
    \mathcal L_{BC}(\theta) = \mathbb{E}_{s \sim \mathcal{B}} [ D_{KL}^s\infdivx{\pi_*(\cdot \vert s)}{\pi_\theta(\cdot \vert s)} ],
\end{equation}
where $\mathcal{B}$ is a buffer of data containing states from pre-training. 

In \textbf{Kickstarting} (KS)~\cite{schmitt2018kickstarting}, we use a very similar loss, but now we apply KL on the data gathered online by the student. More formally:
\begin{equation}
    \mathcal L_{KS}(\theta) = \mathbb{E}_{s \sim \mathcal{B}_\theta} [ D_{KL}^s\infdivx{\pi_*(\cdot \vert s)}{\pi_\theta(\cdot \vert s)} ],
\end{equation}
where $\mathcal{B}_\theta$ denotes a buffer of data gathered by the online policy $\pi_\theta$.




\subsection{Replay-based methods}

\newdiff{A simple way to mitigate forgetting is to add the prior data to the training dataset for the current dataset (in supervised learning~\citep{chaudhry2019tiny,buzzega2021rethinking}) or to the replay buffer (in off-policy RL~\citep{rolnick2019experience,kessler2022same}). By mixing the data from the previous and the current task, one approximates the perfectly mixed i.i.d. data distribution, thus going closer to stationary learning. }

In our experiments, we use a simple episodic memory (EM) approach along with the off-policy SAC algorithm. At the start of the training, we gather a set of trajectories from the pre-trained environment and we use them to populate SAC's replay buffer. In our experiments, old samples take $10\%$ of the whole buffer size. Then, throughout the training we protect that part of the buffer, i.e. we do not allow the data from the pre-trained task to be overridden.

Although episodic memory performs well in our experiments, it is difficult to use this strategy in settings with on-policy algorithms. In particular, we cannot trivially use it with PPO in Montezuma's Revenge and with APPO in NetHack as these methods do not use a replay buffer and might become unstable when trained with off-policy data. Additionally, we note that episodic memory seems to work poorly with SAC in traditional continual learning settings~\cite{wolczyk2021continual,wolczykdisentangling}. As such, we focus on the distillation approaches instead.

\subsection{Parameter-isolation methods}
Standard taxonomies of continual learning~\cite{de2021continual} also consider parameter isolation-based (or modularity-based) method. Such methods assign a subset of parameters to each task and preserve the performance by keeping these weights frozen. For example, Progressive Networks~\cite{rusu2016progressive} introduces a new set of parameters with each introduced task, and PackNet~\cite{mallya2018packnet} freezes a subset of existing weights after each task. Recent works showed that by carefully combining the modules, one can achieve a significant knowledge transfer without any forgetting~\cite{veniat2021efficient,ostapenko2021continual}. However, in most cases, methods in this family require access to the task ID. Although we provide the stage ID in our controlled \roboticsequence{} environments, most realistic problems, such as NetHack, do not have clearly separable tasks and as such application of such methods to the general fine-tuning problem might be non-trivial.

\subsection{Note on critic regularization}
\label{app:critic_details}
In actor-critic architectures popular in reinforcement learning, one can decide whether to apply knowledge retention methods only to the actor and only to the critic. If all we care about is the policy being able to correctly execute the policies for the previous tasks, then it is enough to force the actor to not forget. Since the critic is only used for training, forgetting in the critic will not directly impact the performance. On the other hand, in principle preserving knowledge in the critic might allow us to efficiently re-train on any of the prior tasks. In this paper, following~\cite{wolczykdisentangling} we focus on regularizing only the actor, i.e. we do not apply any distillation loss on the critic in distillation-based methods and we do not minimize the distance on the L2 norm on the critic-specific parameters.

\newpage

\section{Additional NetHack results}
\label{app:nethack_results_big}

\begin{figure*}
    \centering
    \includegraphics[width=\textwidth]{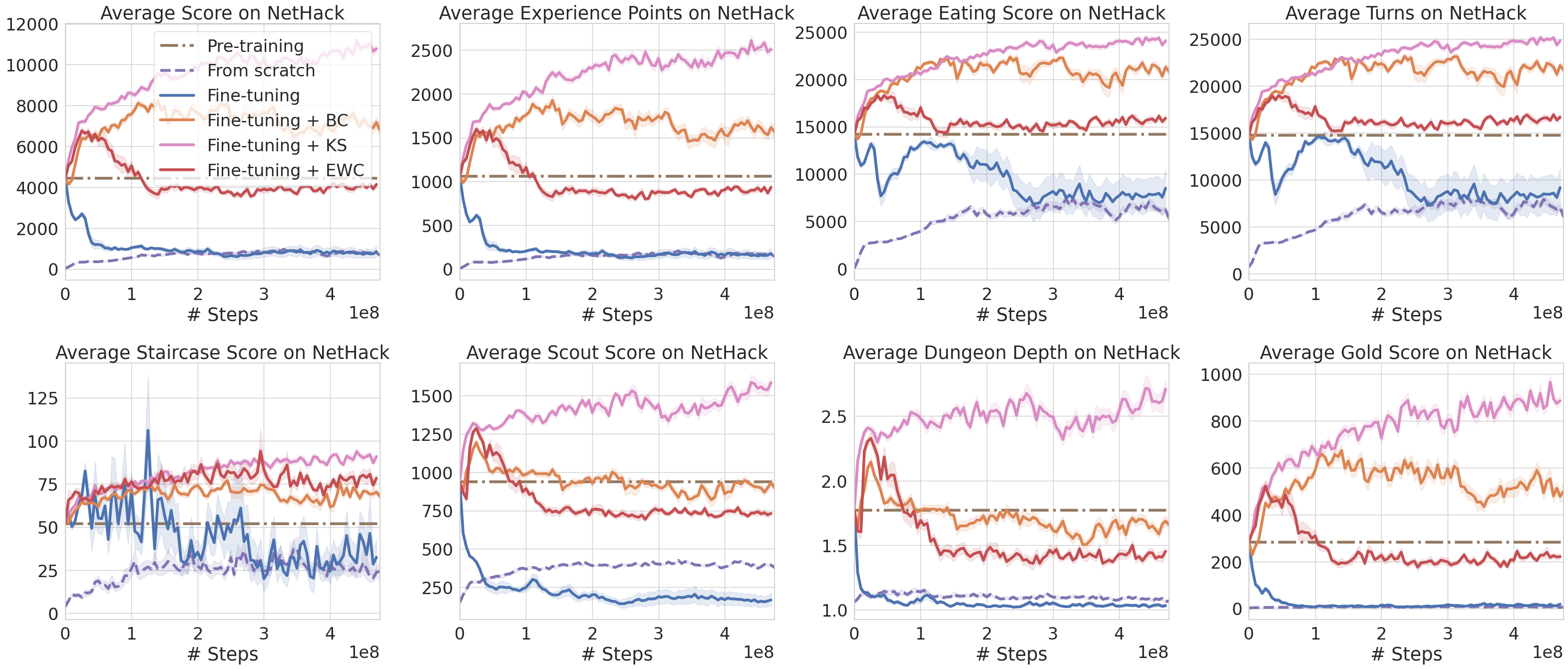}
    \caption{
    Performance on NetHack on additional metrics. Gold Score, Eating Score, Staircase Score and Scout Score are measured in the same way as additional tasks defined in NLE~\cite{kuttler2020nethack}. Score, Turns, Experience Points and Dungeon Depth are taken from blstats. All metrics are measured throughout the training.}
    \label{fig:nethack_auxiliary_results}
\end{figure*}

\paragraph{Additional metrics of NetHack performance}
In Figure~\ref{fig:nethack_auxiliary_results}, we visualize additional metrics. Some of them were originally introduced as tasks in NLE~\cite{kuttler2020nethack} (Gold Score, Eating Score, Staircase Score, and Scout Score), while the others are displayed at the bottom of the screen as statistics (Score, Turns, Experience Points, and Dungeon Depth). These metrics were measured throughout the training, providing a detailed perspective on the behavior of agents. Indeed, it is evident that knowledge retention methods are crucial for making progress in the game, as fine-tuning + KS achieves the highest score while also being the best in all other metrics that measure progress in the actual game. This observation confirms the importance of score as a reliable proxy for game progress, as methods achieving higher scores almost always outperform others in all additional metrics.

While the previous results were gathered during the training process, in Table~\ref{tab:nethack_last_checkpoint_eval} we provide different metrics for the full evaluation. Following the community standards~\cite{kuttler2020nethack}, we take the last checkpoints of each run and we generate $1000$ trajectories from it. The results again show that fine-tuning with knowledge retention methods helps and, in particular, fine-tuning + KS achieves state-of-the-art results throughout all metrics.

Additionally, in Table~\ref{tab:nethack_others} we position our score results against the prior work.

\begin{table}[H]
\caption{NetHack full evaluation results on last checkpoint of each run for 1000 episodes.}
\label{tab:nethack_last_checkpoint_eval}
\centering
\small

\begin{tabular}{lrrrrrrrrrr}
\toprule
method & score & turns & steps & dlvl & xplvl & eating & gold & scout & sokoban & staircase \\
\midrule
From scratch & 776 & 6696 & 13539 & 1.06 & 4.07 & 5862.56 & 5.34 & 370.62 & 0.00 & 25.17 \\
Fine-tuning & 647 & 7756 & 13352 & 1.02 & 2.73 & 7161.20 & 9.26 & 149.70 & 0.00 & 19.94 \\
Fine-tuning + EWC & 3976 & 16725 & 35018 & 1.41 & 6.29 & 15896.45 & 217.12 & 719.70 & 0.00 & 81.74 \\
Fine-tuning + BC & 7610 & 22895 & 34560 & 1.7 & 7.30 & 21995.63 & 582.33 & 959.34 & 0.00 & 69.89 \\
Fine-tuning + KS & 10588 & 24436 & 38635 & 2.66 & 7.73 & 23705.56 & 857.20 & 1551.18 & 0.04 & 90.10 \\

\bottomrule
\end{tabular}
\end{table}

\begin{table}[H]
\caption{Score comparison of methods from prior work and our best performing method (denoted as Fine-tuning + KS in the main text, here as "Scaled-BC + Fine-tuning + KS" to differentiate the pre-trained model).}
\label{tab:nethack_others}
\centering

\begin{tabular}{lr}
\toprule
\textbf{Models} & \textbf{Human Monk} \\
\midrule
\textbf{Offline only} & \\
DQN-Offline~\cite{hambrodungeons} & 0.0 ± 0.0 \\
CQL~\cite{hambrodungeons} & 366 ± 35 \\
IQL~\cite{hambrodungeons} & 267 ± 28 \\
BC (CDGPT5)~\cite{hambrodungeons,hambro2022insights}  & 1059 ± 159 \\
Scaled-BC~\cite{tuyls2023scaling} & 5218 ± - \\
\midrule
\textbf{Offline + Online} & \\
From Scratch + KS~\cite{hambrodungeons} & 2090 ± 123 \\
From Scratch + BC~\cite{hambrodungeons} & 2809 ± 103 \\
LDD$^{*}$~\cite{mu2022improving}  & 2100 ± - \\
\textbf{Scaled-BC + Fine-tuning + KS (ours)}  & \textbf{10588 ± 672} \\
\bottomrule
\end{tabular}

\end{table}

\begin{figure}[H]
    \centering
    \includegraphics[width=\textwidth]{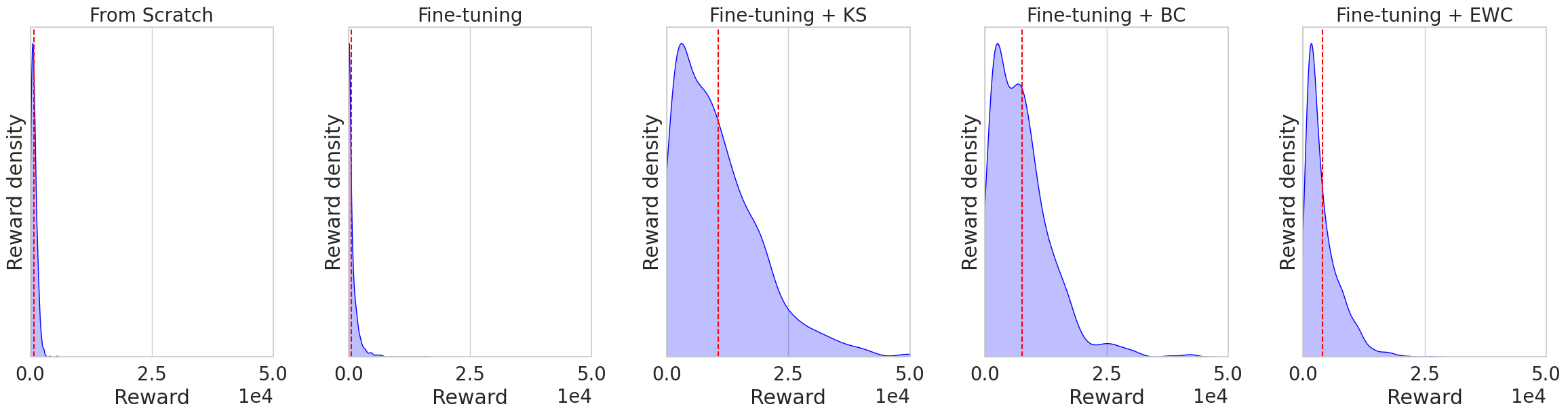}
    \caption{Return distribution for each of the tested methods. The mean return is denoted by the dashed red line.}
    \label{fig:nethack_return_distribution}
\end{figure}

\paragraph{Return density} In previous sections we looked at the mean return. Here, to better understand the behavior of the tested methods, we also look at the whole distribution of returns. This way, we can understand whether e.g., the score of a given method relies on a few lucky high-reward runs. The results presented in Figure~\ref{fig:nethack_return_distribution} show that while from scratch and fine-tuning achieve consistently poor returns, the variance in scores is much higher for fine-tuning with knowledge retention. In particular, we observe that there are occurrences of fine-tuning + KS achieving returns as high as $50000$. At the same time, there is a significant time of unlucky runs that end with a return of $1000$. We can attribute this variance to the high stochasticity of the game, e.g., if the first level happens to contain many monsters that are difficult to defeat, that episode may end earlier than expected.

\begin{figure}[H]
    \centering
    \includegraphics[width=\textwidth]{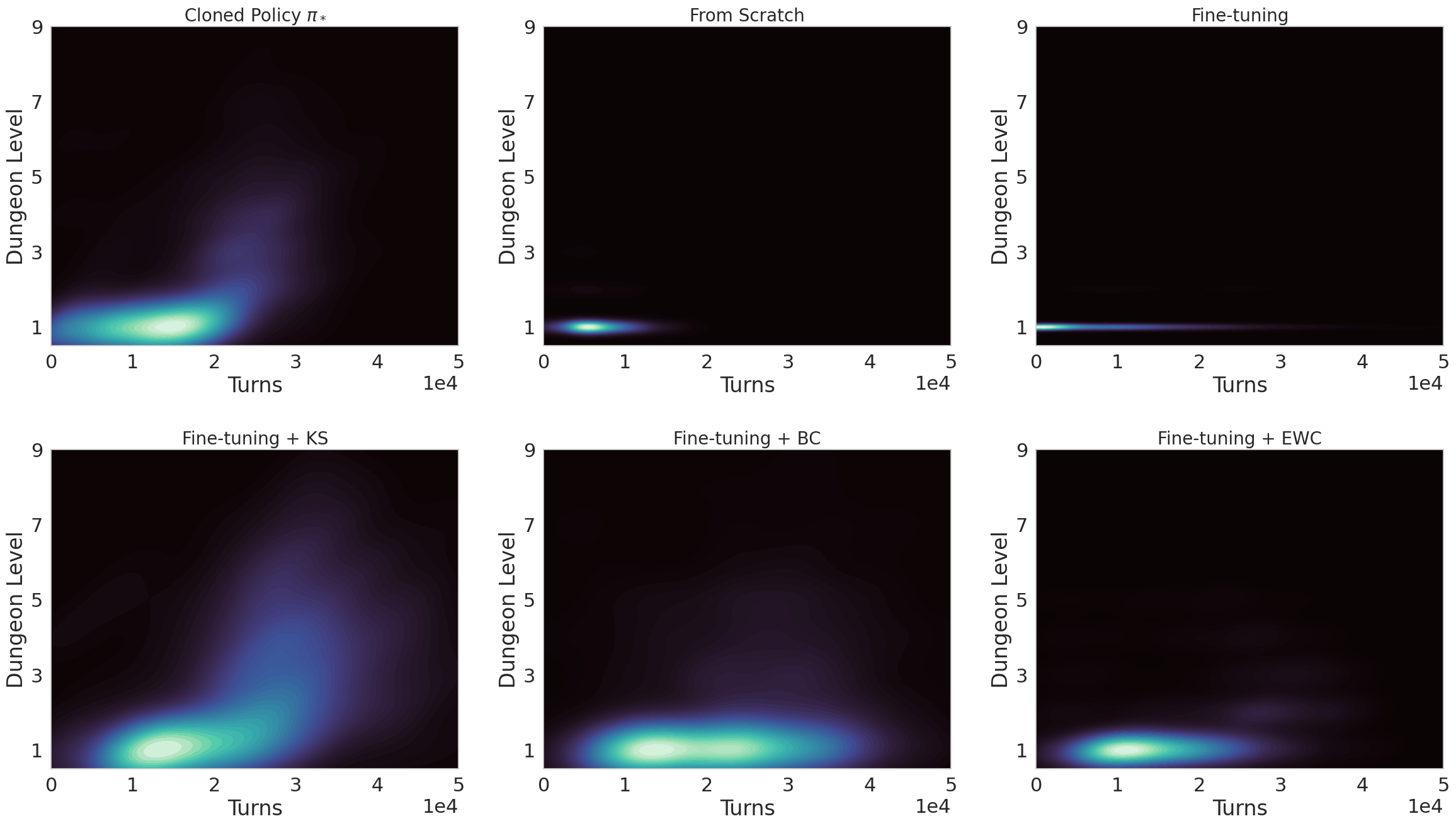}
    \caption{Density plots showing maximum dungeon level achieved compared to the total number of turns (units of in-game time). Brighter colors indicate higher visitation density. }
    \label{fig:nethack_density_appendix}
\end{figure}

\paragraph{Level visitation density} In Figure~\ref{fig:nethack_density_appendix} we show the level density plots from Figure~\ref{fig:nethack_level_visitation} for all methods. In particular, we observe that fine-tuning and training from scratch almost never manage to leave the first level, confirming their poor performance with respect to score.

\newpage
\diff{\section{Additional Montezuma's Revenge results}}
\label{app:montezuma_results}

\begin{figure}[H]
    \centering
    \begin{subfigure}[t]{0.49\textwidth}
        \centering
        \mute{\includegraphics[width=\textwidth]{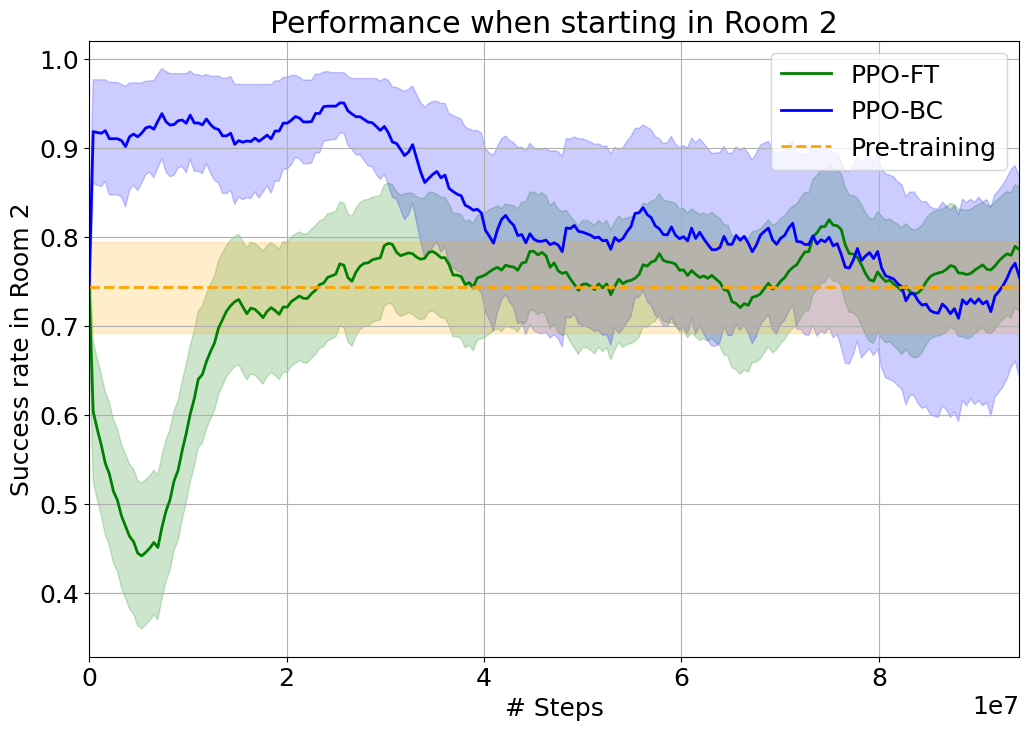}}
    \end{subfigure}
    \begin{subfigure}[t]{0.49\textwidth}
        \centering
        \mute{\includegraphics[width=\textwidth]{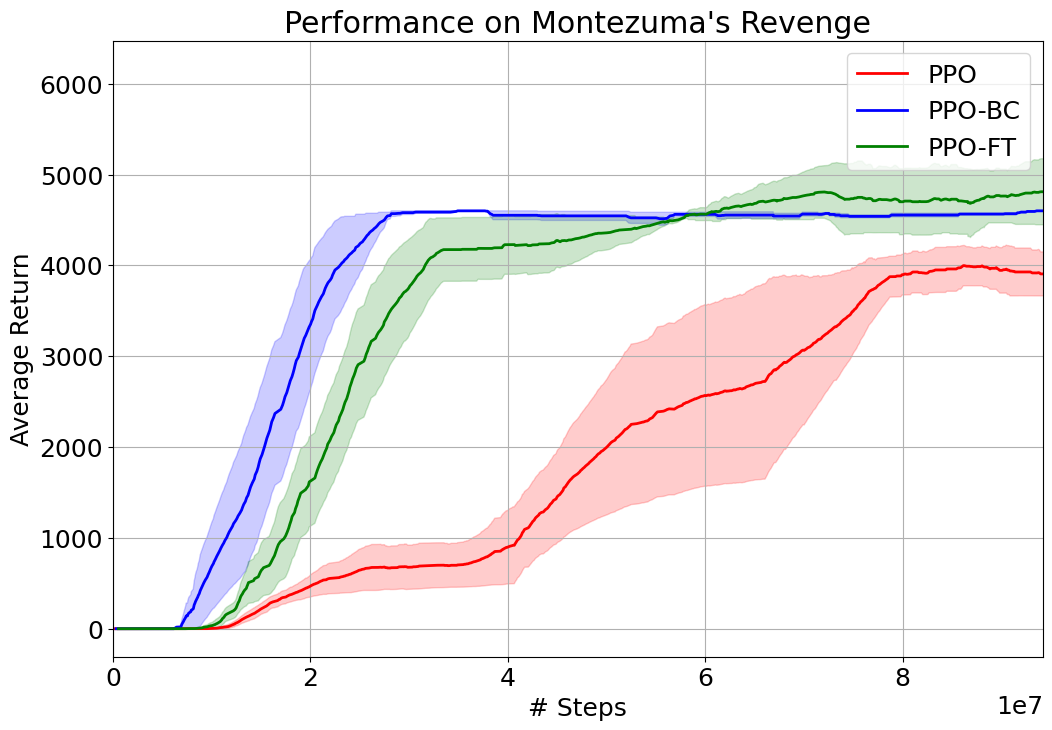}}
    \end{subfigure}
    \begin{subfigure}[t]{0.49\textwidth}
        \centering
        \mute{\includegraphics[width=\textwidth]{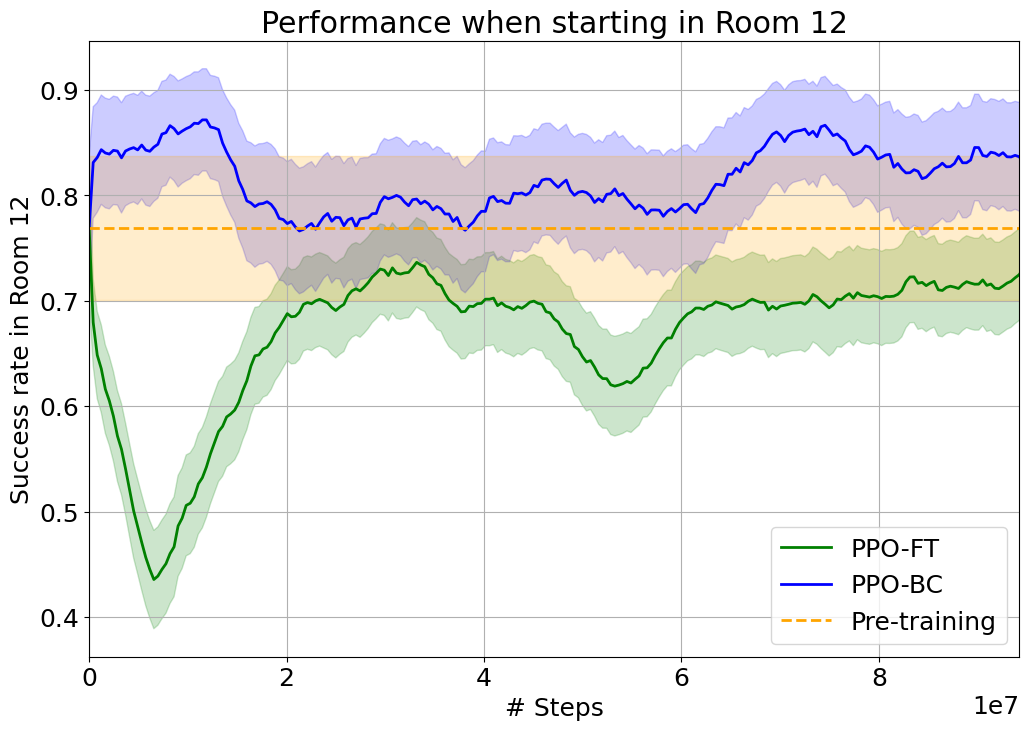}}
    \end{subfigure}
    \begin{subfigure}[t]{0.49\textwidth}
        \centering
        \mute{\includegraphics[width=\textwidth]{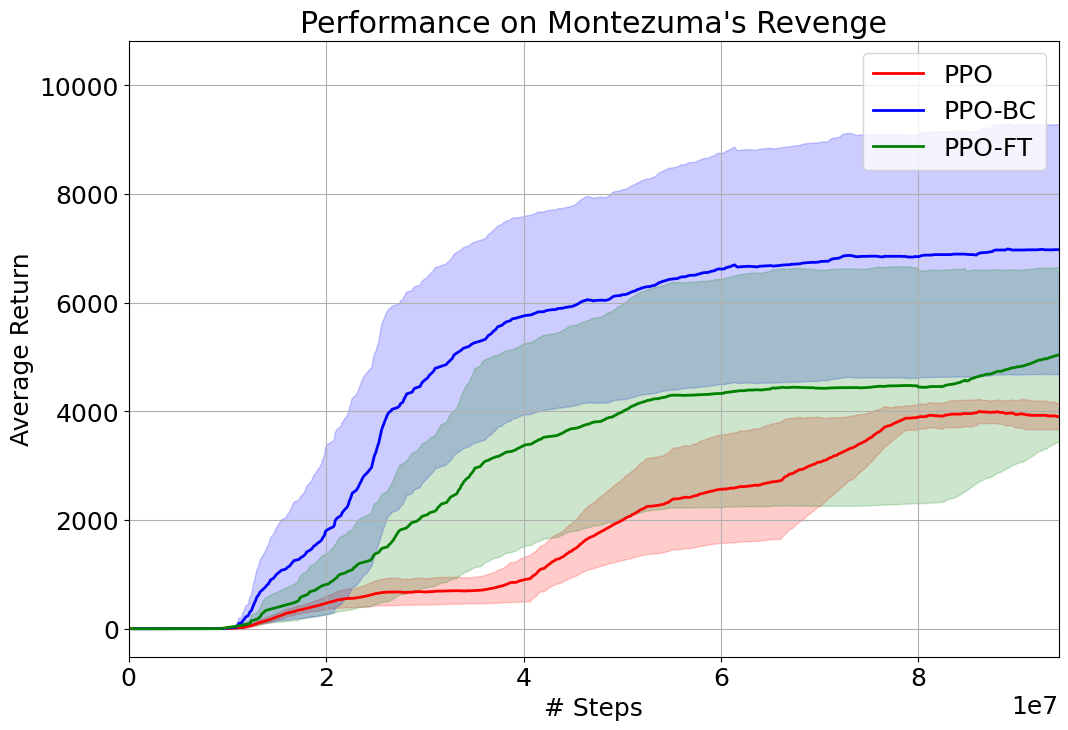}}
    \end{subfigure}
    \begin{subfigure}[t]{0.49\textwidth}
        \centering
        \mute{\includegraphics[width=\textwidth]{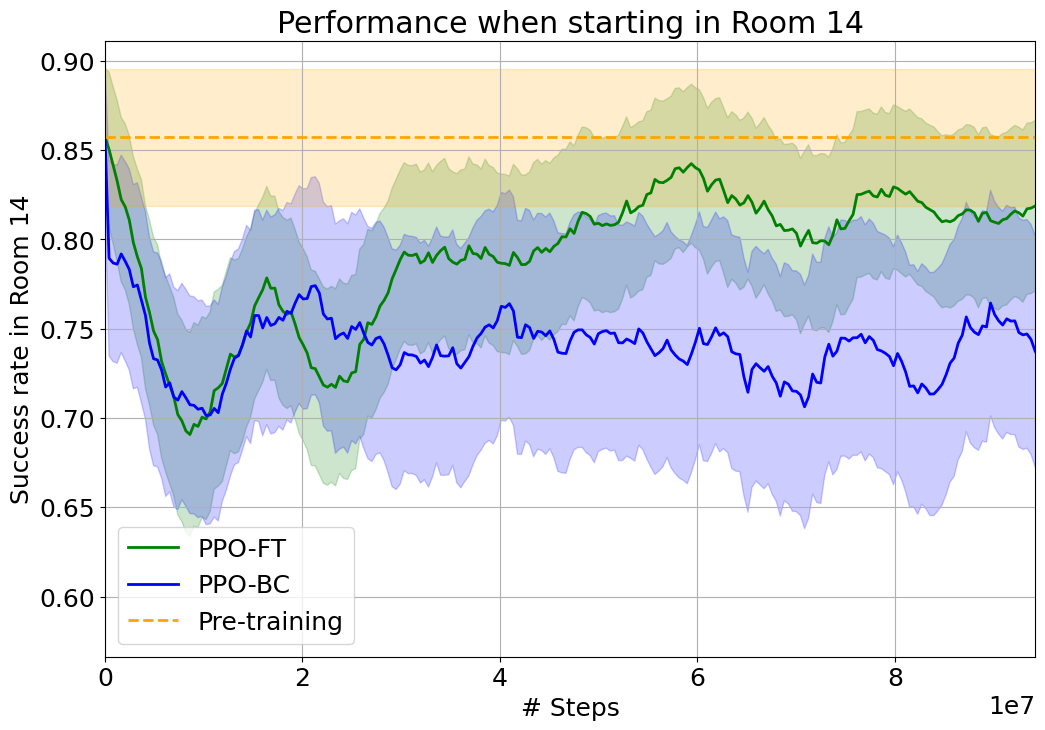}}
        \caption{Success rate in rooms during fine-tuning when initialized in that room.}
        \label{fig:montezuma_room_app}
    \end{subfigure}
    \begin{subfigure}[t]{0.49\textwidth}
        \centering
        \mute{\includegraphics[width=\textwidth]{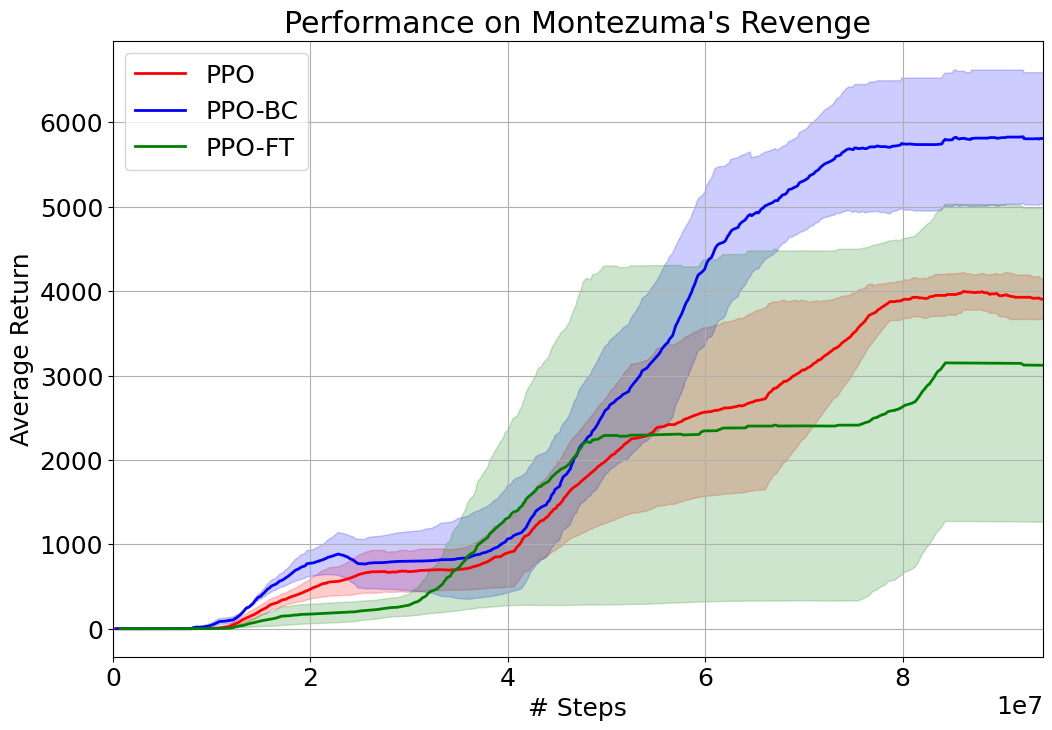}}
        \caption{Average return throughout the training. PPO represents training from scratch, PPO-FT is vanilla fine-tuning and PPO-BC is fine-tuning with the behavioral cloning loss.}
        \label{fig:montezuma_training14_app}
    \end{subfigure}
    
    \caption{\diff{\small \envshiftUpper{} in Montezuma's Revenge.}}
    \label{fig:montezuma_app}
\end{figure}

\begin{figure}[H]
    \centering
    \begin{subfigure}[t]{0.7\textwidth}
        \centering
        \mute{\includegraphics[width=\textwidth]{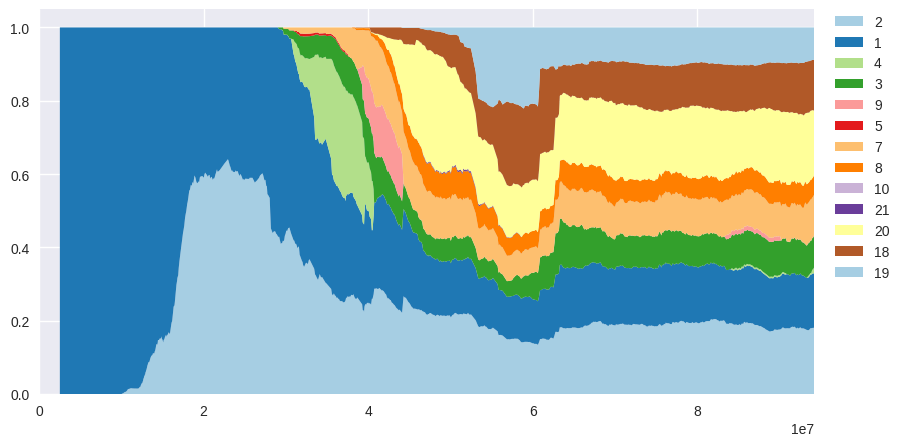}}
        \caption{Room visitation for training from scratch}
    \end{subfigure}
    \begin{subfigure}[t]{0.7\textwidth}
        \centering
        \mute{\includegraphics[width=\textwidth]{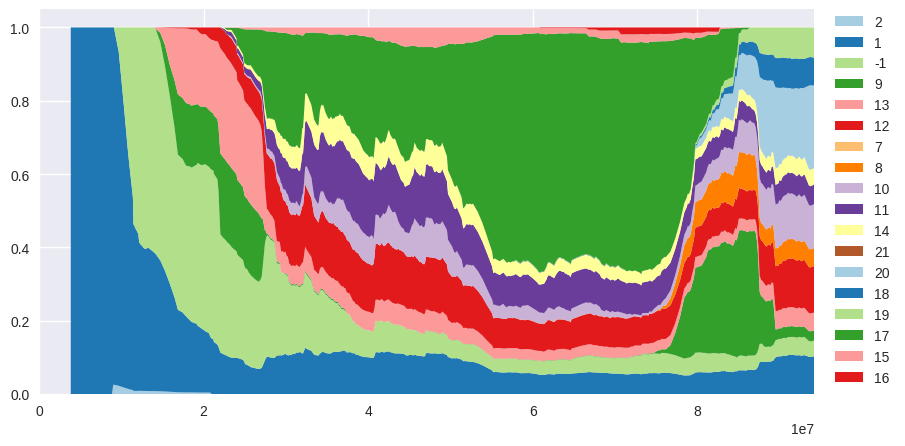}}
        \caption{Room visitation for fine-tuning}
    \end{subfigure}
    \begin{subfigure}[t]{0.7\textwidth}
        \centering
        \mute{\includegraphics[width=\textwidth]{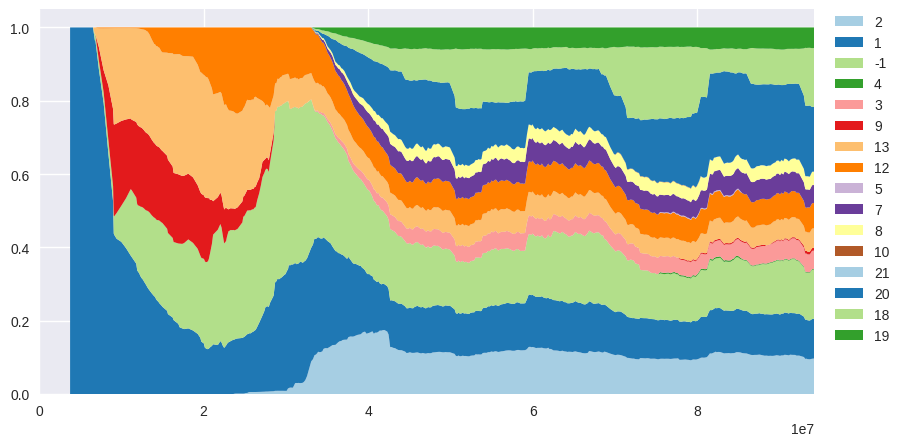}}
        \caption{Room visitation for fine-tuning + BC}
    \end{subfigure}
    
    \caption{Time spent in different rooms across training for training from scratch (top), fine-tuning (middle), and fine-tuning + BC (bottom). The agent trained from scratch struggles to explore rooms at the beginning of the training and eventually visits fewer of them than fine-tuned agents.}
    \label{fig:montezuma_visits}
\end{figure}

\begin{figure}[H]
    \centering
    \begin{subfigure}[t]{0.49\textwidth}
        \centering
        \mute{\includegraphics[width=\textwidth]{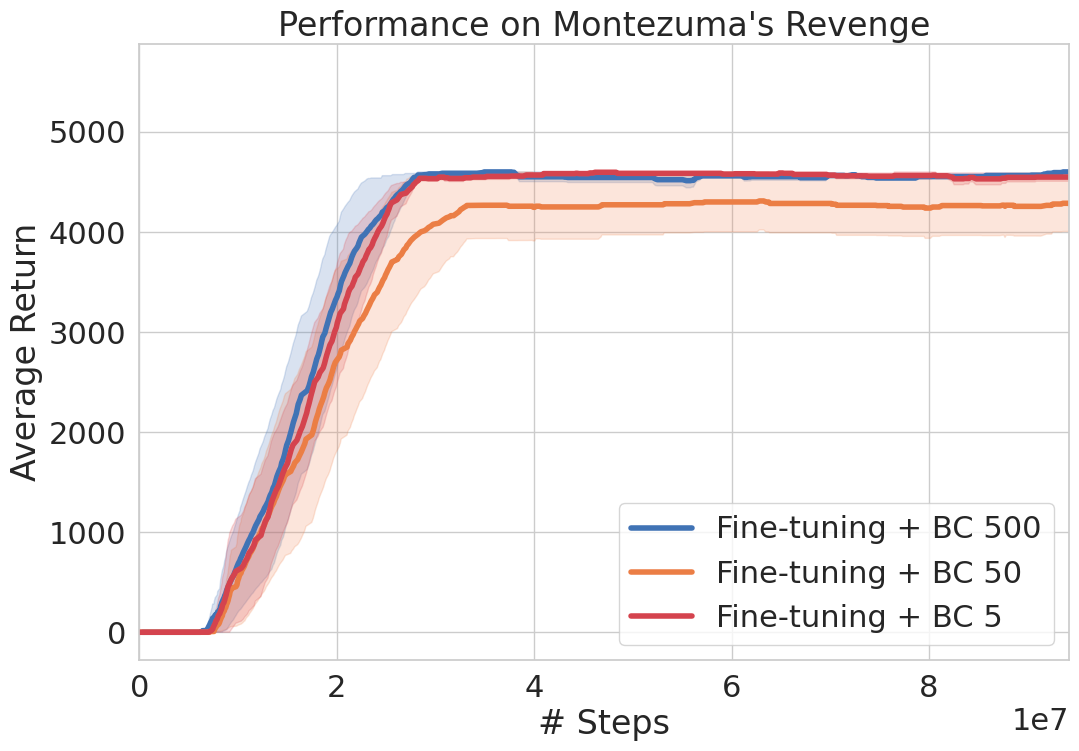}}
        \caption{Room 2}
    \end{subfigure}
    \begin{subfigure}[t]{0.49\textwidth}
        \centering
        \mute{\includegraphics[width=\textwidth]{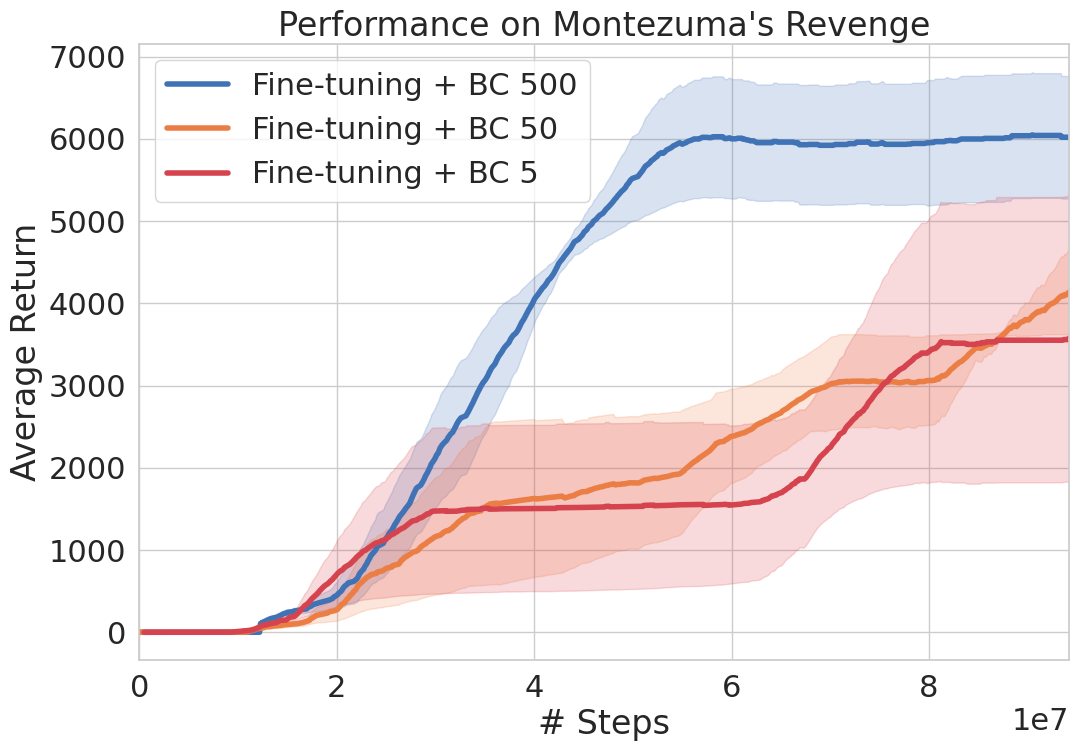}}
        \caption{Room 7}
    \end{subfigure}
    \begin{subfigure}[t]{0.49\textwidth}
        \centering
        \mute{\includegraphics[width=\textwidth]{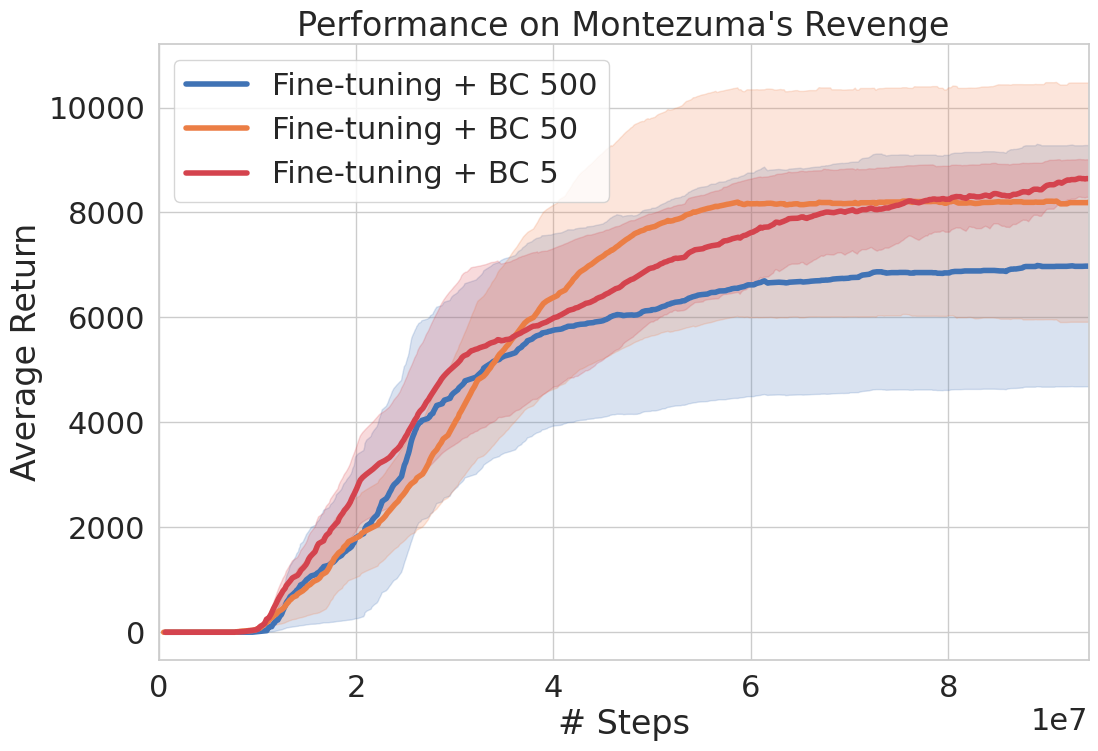}}
        \caption{Room 12}

    \end{subfigure}
    \begin{subfigure}[t]{0.49\textwidth}
        \centering
        \mute{\includegraphics[width=\textwidth]{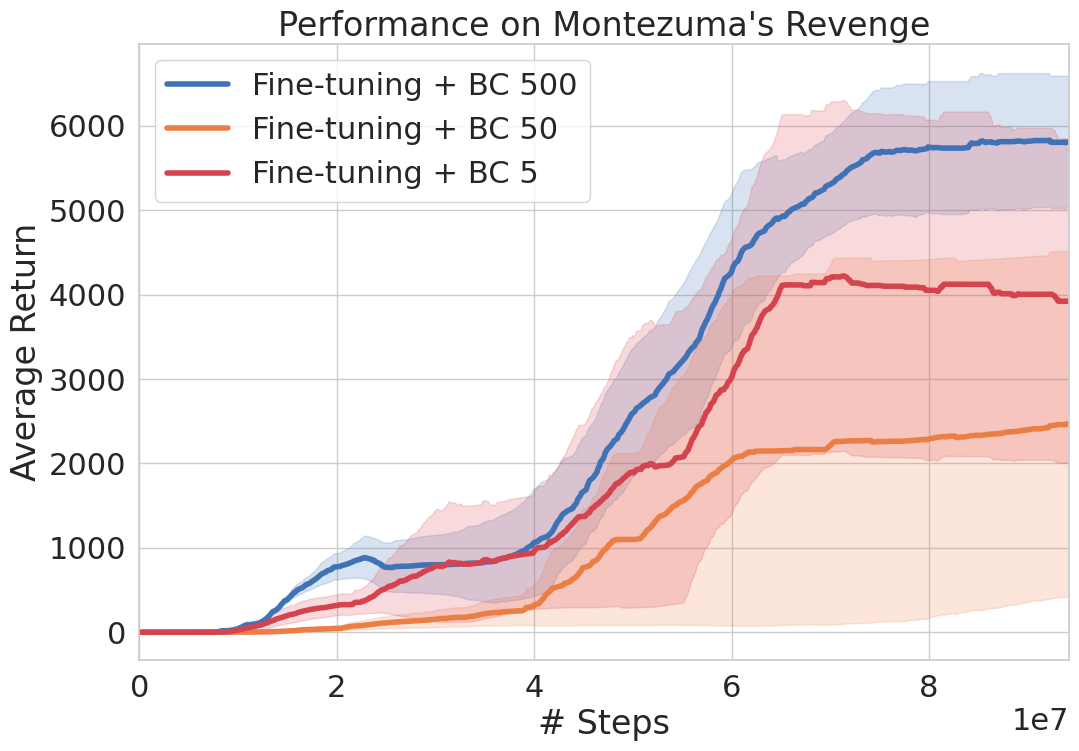}}
        \caption{Room 14}
    \end{subfigure}
    
    \caption{Results for different buffer sizes in Montezuma's Revenge.}
    \label{fig:montezuma_buffer_size}
\end{figure}

\paragraph{Analysis of forgetting with different pre-training schemes}
We perform additional experiments on three different rooms in a setting analogous to the one from the main paper (see Section~\ref{sec:exp_setup} for details). In particular, we are interested in the behavior of the pre-trained model from a specific room while fine-tuned. Figure \ref{fig:montezuma_app} shows a significant drop in performance for vanilla fine-tuned models without additional knowledge retention methods (PPO-FT) just after fine-tuning starts. In contrast, PPO-BC (i.e. fine-tuning + BC) mitigates this effect except for Room 14. For all pre-training types, PPO-BC outperforms PPO-FT with respect to the score.

\paragraph{Room visitation analysis} Since exploration is a crucial problem in Montezuma's Revenge, we check how well different types of agents manage to explore the maze throughout the game. In Figure \ref{fig:montezuma_visits}, we show how the time spent in different rooms changes across the training for an agent trained from scratch, the fine-tuned agent, and the fine-tuned agent with BC loss. For simplicity, we focus on our primary setting, i.e. the one where pre-training starts from Room 7. 

The agent trained from scratch spends a significant amount of time learning to escape the first two rooms and navigate the maze. Interestingly, both vanilla fine-tuning and fine-tuning + BC retain the capability for exploration obtained in pre-training, as they exit the first room quickly, even though it was not seen at all during pre-training. However, in the later phase of fine-tuning, the agent with knowledge retention manages to see a wider variety of rooms than the one without it, which spends a significant amount of time in e.g. Room 9. This suggests that \problem{} also applies to exploration capabilities and knowledge retention methods can mitigate their loss.

\paragraph{Impact of the buffer size} Finally, we check how the size of the replay buffer for Fine-tuning + BC impacts the results. Results presented in Figure~\ref{fig:montezuma_buffer_size} show that indeed having a larger buffer is always the best option, although the performance gap vanishes in some settings. 



\newpage
\section{Analysis of forgetting in robotic manipulation tasks}
\label{app:robotic_analysis}

In this section, we present additional results for our robotic manipulation experiments based on Meta-World.

Unless specified otherwise, we use the experimental setting from Section~\ref{sec:exp_setup}. We adopt the forward transfer metric used previously in~\cite{wolczyk2021continual,bornschein2022nevis} to measure how much pre-trained knowledge helps during fine-tuning:
\begin{align*}
    \text{Forward Transfer}    & := \frac{\text{AUC}-\text{AUC}^b}{1-\text{AUC}^b},\quad                        
    \text{AUC}   := \frac{1}{T}\int_{0}^{T} p(t) \text{d}t, \quad 
    \text{AUC}^b  :=  \frac{1}{T}\int_{0}^{T} p^b(t) \text{d}t,
\end{align*}
where $p(t)$ is the success rate of the pre-trained model at time $t$, $p^b$ denotes the success rate of a network trained from scratch, and $T$ is the training length. Intuitively, it measures how much faster the fine-tuned model learns than the one trained from scratch. 

\begin{figure}[t!]
    \centering
    \mute{\includegraphics[width=0.99\textwidth]{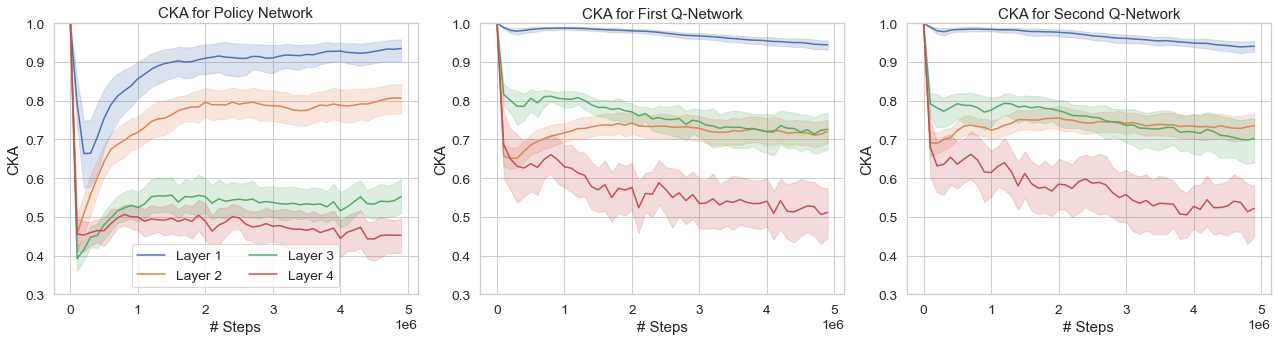}}
    \caption{The CKA values throughout vanilla fine-tuning (without knowledge retention methods), computed between the activations of the pre-trained model and the activations of the current model. The higher the values, the more similar the representations.}
    \label{fig:cka_results}
\end{figure}

\textbf{Analysis of internal representations}
We examine how activations of the actor and critic networks in SAC change throughout fine-tuning when we do not use any knowledge retention methods, with the goal of pinpointing the structure of forgetting. To measure the representation shift in the network, we use the Central Kernel Alignment (CKA)~\citep{kornblith2019similarity} metric, which was previously used in studying forgetting in the supervised learning paradigm~\citep{ramasesh2020anatomy,mirzadeh2022architecture}. 
Before starting the fine-tuning process, we collect optimal trajectories from the pre-trained model along with the activations of the networks after each layer. Then, at multiple points throughout the training process, we feed the same trajectories through the fine-tuned network and compare its activations to the prior activations using CKA. Figure~\ref{fig:cka_results} shows that, in general, later layers change more than the early layers, which is consistent with previous studies~\citep{ramasesh2020anatomy}. This is particularly visible in the policy network, while the tendency is not as strong for the critic networks, suggesting that the TD-learning guiding the critic leads to different representation learning dynamics.

In the policy network, representations in the early layers change rapidly at the beginning of the fine-tuning process. Then, interestingly, as we solve the new tasks and revisit the tasks from pre-training, CKA increases and the activations become more similar to the pre-trained ones. 
As such, the re-learning visible in per-task success rates in Figure~\ref{fig:analysis_robotic} is also reflected in the CKA here.
However, this phenomenon does not hold for the later layers in the policy network or the $Q$-networks. This suggests that 
the solution we find is significantly different.

\paragraph{Impact of the network size} Previous studies in supervised continual learning showed that forgetting might start disappearing as we increase the size of the neural network~\cite{ramasesh2022effect,mirzadeh2022architecture}, and here we investigate the same point in RL using our \roboticsequence{} setting.
We run a grid of experiments with hidden dimensions in $\{256, 512, 1024\}$ and number of layers in $\{2, 3, 4\}$. For each of these combinations, we repeat the experiment from the main text, namely, we measure how fine-tuning from a pre-trained solution compares to starting from random initialization and how the results change when we apply continual learning methods. The results are presented in Figure~\ref{fig:full_arch_results}. 


The results do not show any clear correlations between the network size and forgetting, hinting at more complex interactions than these previously showed in continual supervised learning literature~\cite{ramasesh2022effect}. The fine-tuning approach fails to achieve a significant positive transfer for two or four layers, but it does show signs of knowledge retention with three layers. 
Inspection of the detailed results for the three-layer case shows that the fine-tuning performance on the known tasks still falls to zero at the beginning, but it can regain performance relatively quickly.
As for the CL methods, we observe that behavioral cloning performs well independently of the size of the network. On the other hand, EWC tends to fail with two layers. Since EWC directly penalizes changes in the parameters, we hypothesize that with a small, two-layer network, the resulting loss of plasticity makes it especially difficult to learn. 

\paragraph{Impact of the number of unknown tasks}  In our $\textsc{AppleRetrieval}$ experiments, we showed that \problem{} is more visible as we increase the amount of time spent before visiting the known part of the state space. We investigate the same question in the context of robotic manipulation tasks by changing the number of new tasks the agent has to solve prior to reaching the ones it was pre-trained on. 
That is, we study \roboticsequence{}s where the last two tasks are \texttt{peg-unplug-side} and \texttt{push-wall}, as previously, but the first tasks are taken as different length suffixes of \texttt{window-close}, \texttt{faucet-close}, \texttt{hammer}, \texttt{push} 
We call the tasks preceding the pre-trained tasks the \emph{prefix tasks}.

%
%
%
 
We investigate how the number of the prefix tasks impacts the performance on the known tasks during the fine-tuning process. Table~\ref{tab:prefix_len_results_appendix} shows the forward transfer metric computed on the pre-trained tasks for fine-tuning, EWC and BC. 
As the number of prefix tasks grows, the forward transfer values for fine-tuning become smaller, which means that the gains offered by the prior knowledge vanish. Interestingly, even with a single prefix task the forward transfer is relatively low.
On the other hand, continual learning methods do not suffer as much from this issue.  BC achieves high forward transfer regardless of the setting and EWC experiences only small deterioration as we increase the number of prefix tasks. 

\paragraph{Impact of representation vs policy on transfer}

\begin{figure}
    \centering
    \mute{\includegraphics[width=0.7\textwidth]{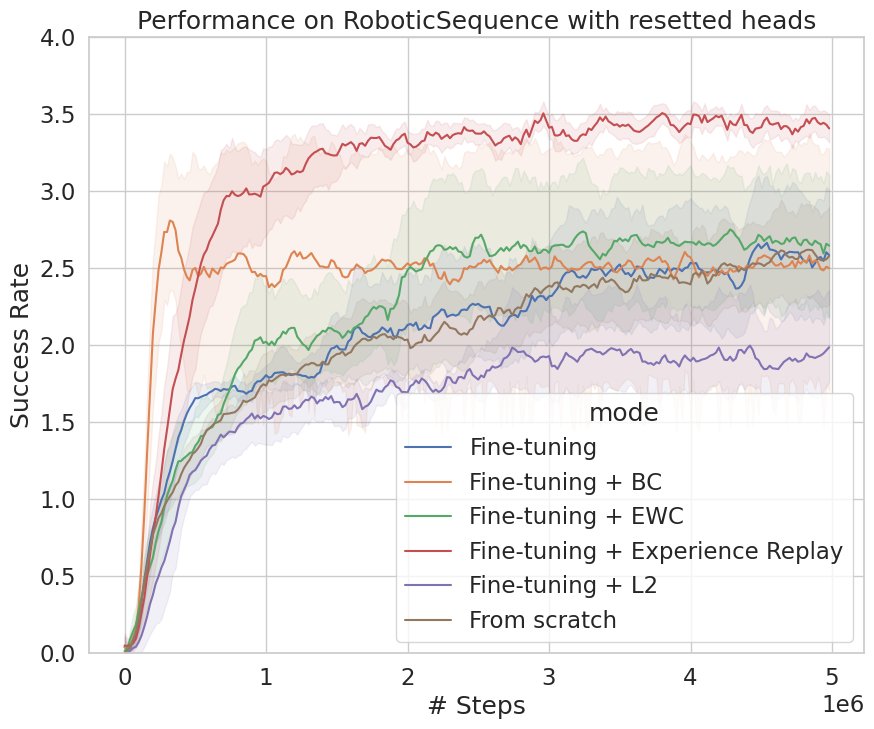}}
    \caption{Performance of different methods on the RoboticSequence where we reset the last layer of the policy and critic networks. The results are worse than in the standard case, but there is still some positive transfer, suggesting that benefits come from reusing both the representations as well as the policy.}
    \label{fig:cw_resethead}
\end{figure}

Although we see significant positive transfer once the forgetting problem is addressed, it remains an open question where this impact comes from. 
Although there are several studies on the impact of representation learning on transfer in supervised learning~\citep{neyshabur2020being,kornblith2021better}, the same question in RL remains relatively understudied. Here, we try to understand the impact of representation and policy on transfer by resetting the last layer of the network before starting the training. As such, the policy at the beginning is random even on the tasks known from pre-training, but has features relevant to solving these tasks. The improvements should then only come from the transfer of representation. 

The results for these experiments are presented in Figure~\ref{fig:cw_resethead}. First of all, we observe that, expectedly, this setting is significantly harder, as all methods perform worse than without resetting the head. However, we still observe significant transfer for BC and EWC as they train faster than a randomly initialized model. At the same time, fine-tuning in the end manages to match the performance of BC and EWC, however at a much slower pace. We hypothesize that the gap between knowledge retention methods and fine-tuning is smaller, since now the methods have to re-learn a new policy rather than maintain the old one. This preliminary experiment suggests that the benefits of fine-tuning come from both the policy and the representation since we can still observe a significant, although reduced, transfer after resetting the heads. Maximizing transfer from the representation remains an interesting open question.

\begin{figure}[t!]
    \centering
    \begin{subfigure}[b]{0.49\textwidth}
        \centering
        \mute{\includegraphics[width=\textwidth]{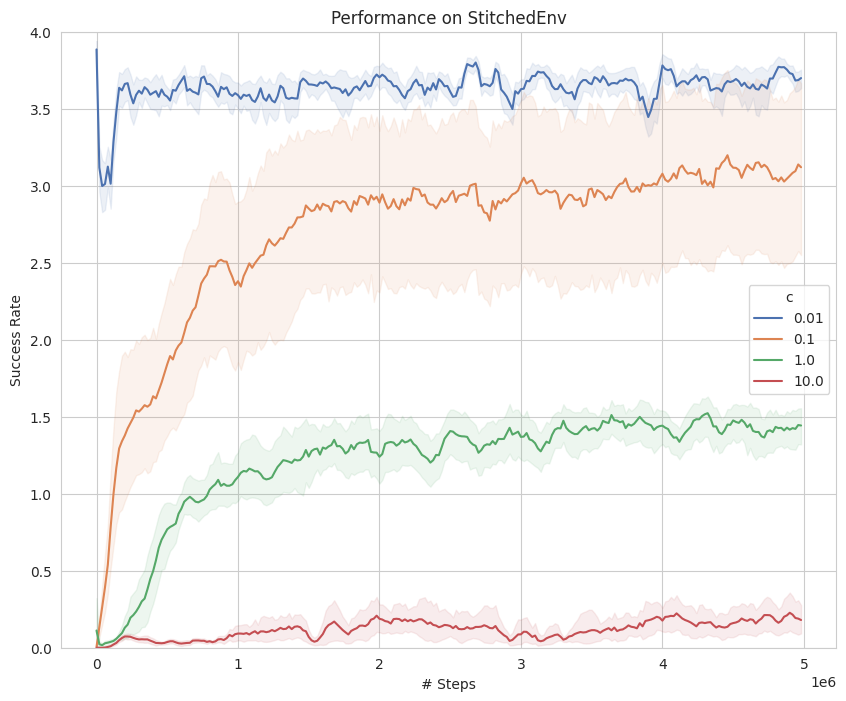}}
    \end{subfigure}
    \begin{subfigure}[b]{0.49\textwidth}
        \centering
        \mute{\includegraphics[width=\textwidth]{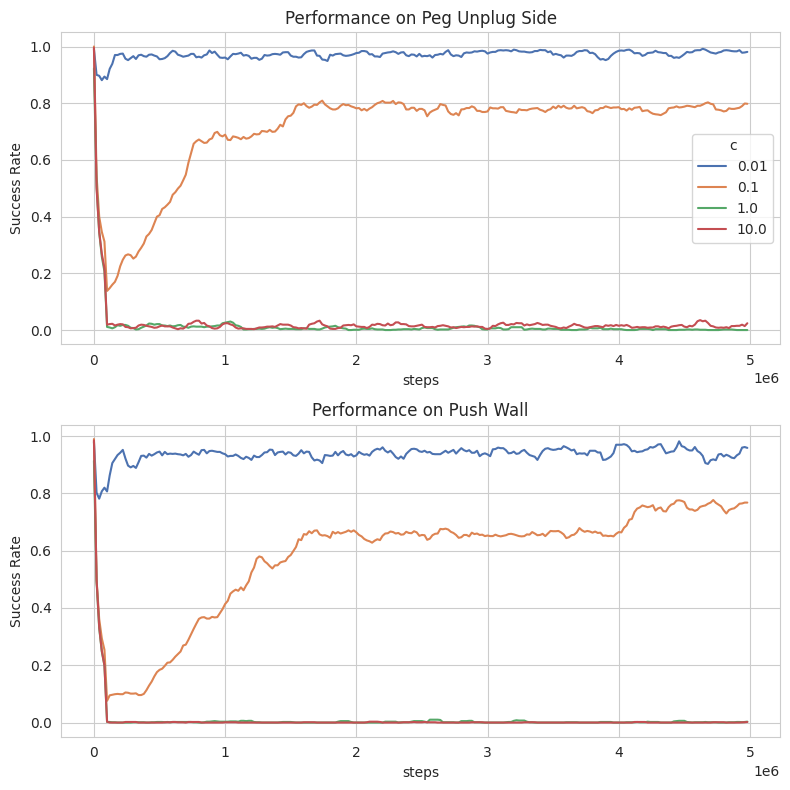}}
    \end{subfigure}

    \caption{The performance on a robotic sequence where the sequence consists of the same tasks, but with observations translated by a constant $c$. We can observe forgetting even for small perturbations ($c=0.1$).}
    \label{fig:stitchedenv_translated}
\end{figure}

\paragraph{Impact of task difference} \camera{The severity of forgetting is deeply connected to how different \textsc{Far} and \textsc{Close} tasks are to each other. We refer the reader to Section~\ref{sec:related_work} for a short description of prior continual learning papers on this problem, and here we perform a simple experiment on this issue.
We construct a RoboticSequence consisting of tasks \texttt{peg unplug (translated), push wall (translated), peg unplug, push wall} and use a model pre-trained on the last two tasks. (Translated) means that the observation vectors are shifted by a constant $c$. This is a very basic form of state perturbation. In this case, the non-translated (translated resp.) stages correspond to \textsc{Far} (\textsc{Close} resp.) states. We run vanilla fine-tuning experiments with values of $c \in (0.01, 0.1, 1, 10)$. We observe no forgetting for $c=0.01$, partial forgetting for $c=0.1$, and total forgetting for $c=1$, and $c=10$. We treat this result as initial evidence supporting the claim that even small discrepancies between far and close states might lead to forgetting.}

\paragraph{Other sequences}
%
%

%

\begin{figure}[t!]
    \centering
    \begin{subfigure}[b]{0.49\textwidth}
        \centering
        \mute{\includegraphics[width=\textwidth]{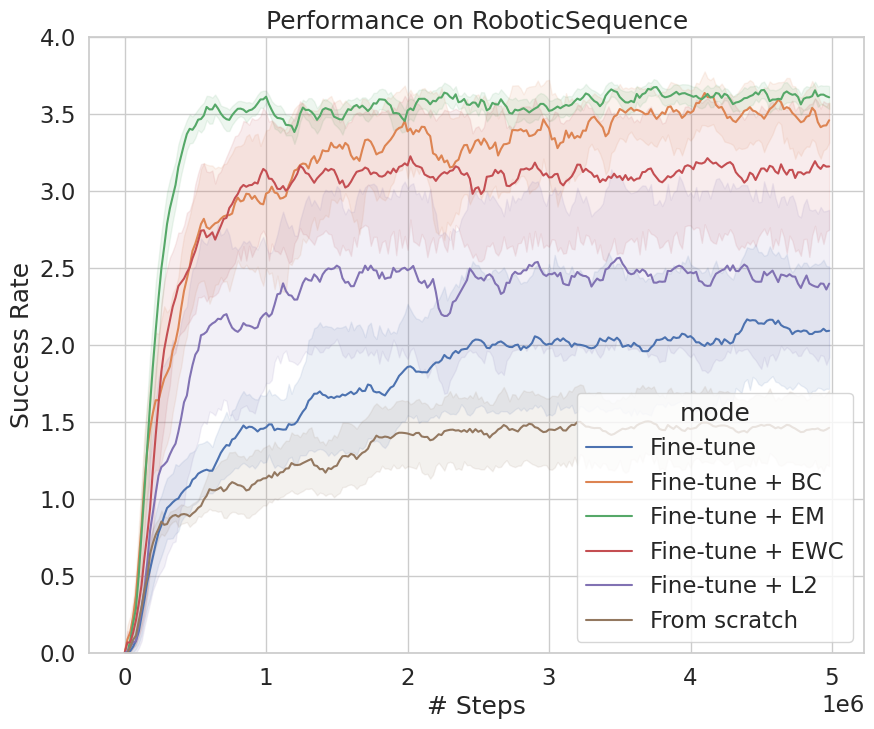}}
    \end{subfigure}
    \begin{subfigure}[b]{0.49\textwidth}
        \centering
        \mute{\includegraphics[width=\textwidth]{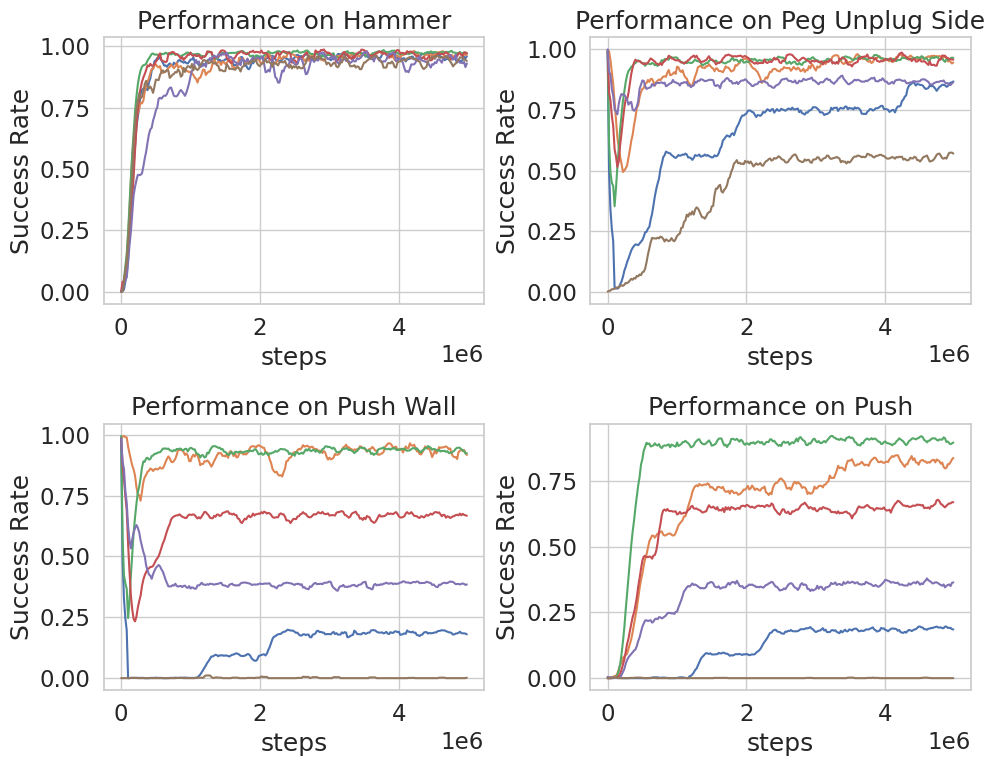}}
    \end{subfigure}

    \caption{The performance on a robotic sequence where the known tasks are in the middle.}
    \label{fig:stitchedenv_middle_seq}
\end{figure}

\begin{figure}[H]
    \centering
    \begin{subfigure}[b]{0.48\textwidth}
        \centering
        \mute{\includegraphics[width=\textwidth]{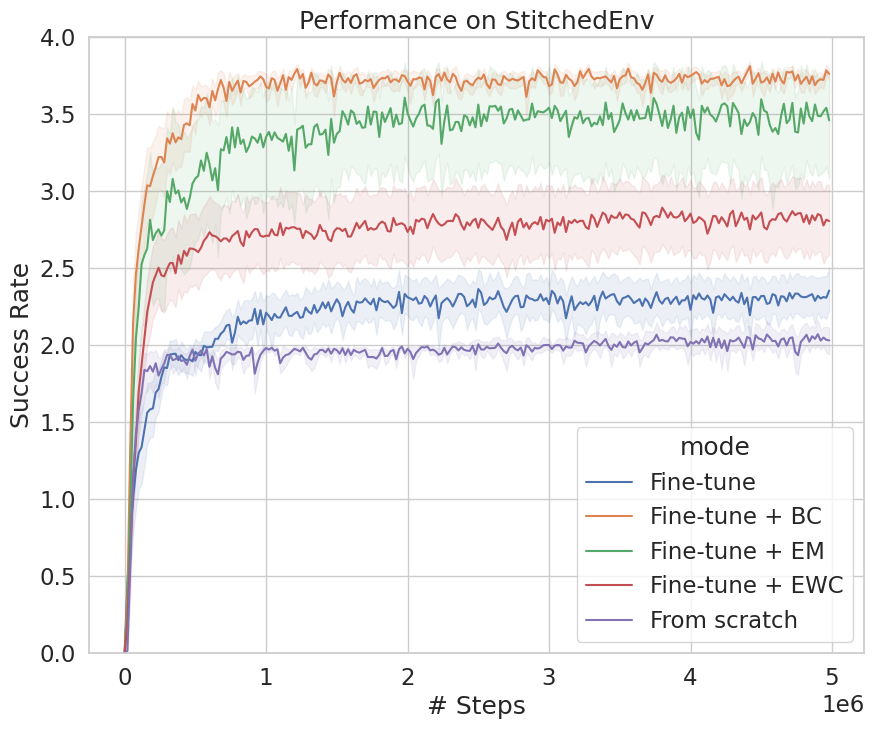}}
    \end{subfigure}
    \begin{subfigure}[b]{0.48\textwidth}
        \centering
        \mute{\includegraphics[width=\textwidth]{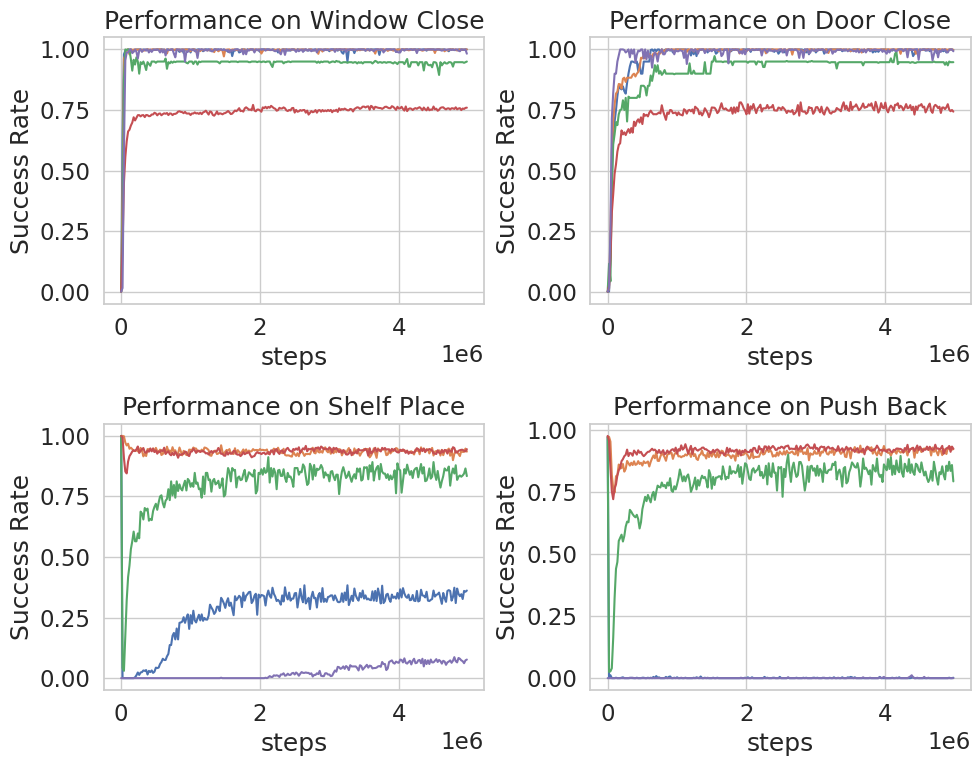}}
    \end{subfigure}

    \caption{The performance of the fine-tuned model on \roboticsequence{} compared to a model trained from scratch and knowledge retention methods on the sequence \texttt{shelf-place}, \texttt{push-back}, \texttt{window-close}, \texttt{door-close}.  
    }
    \label{fig:additional_sequence}
\end{figure}

In order to provide another testbed for our investigations, we repeat the main experiments on another sequence of tasks, namely \texttt{shelf-place}, \texttt{push-back}, \texttt{window-close}, \texttt{door-close}, where again we fine-tune a model that was pre-trained on the last two tasks. The results are presented in Figure~\ref{fig:additional_sequence}. We find that the main conclusions from the other sequence hold here, although, interestingly, the performance of EWC is significantly better.  Additionally, we run experiments on a simple, two task \roboticsequence{} with \texttt{drawer-open} and \texttt{pick-place}, showcased in Figure~\ref{fig:intro}. We used behavioral cloning as an example of a method that mitigates forgetting.

Additionally, we check what happens when the known tasks are "in the middle" of two known tasks. That is, we use the environment consisting of the following sequence of goals: \texttt{hammer, peg-unplug-side, push-wall, push} with a model pre-trained on \texttt{peg-unplug-side, push-wall}. With this setup, we are especially interested in the impact of different methods on the performance on the last task, i.e. can we still learn new things after visiting a known part of the state space?

The results presented in Figure~\ref{fig:stitchedenv_middle_seq} show that the relative performance of all methods is the same as in our original ordering, however, we observe that EWC almost matches the score of BC. The learning benefits on the last task, \texttt{push}, is somewhat difficult to estimate. That is since BC manages to maintain good performance on tasks \texttt{peg-unplug-side} and \texttt{push-wall}, it sees data from \texttt{push} much sooner than approaches that have to re-learn tasks 2 and 3. However, we observe that even after encountering the later tasks, knowledge retention methods perform much better on \texttt{push} than vanilla fine-tuning, which in turn is better than a model trained from scratch.

\begin{figure}[t!]
    \centering
    \begin{subfigure}[b]{0.49\textwidth}
        \centering
        \mute{\includegraphics[width=\textwidth]{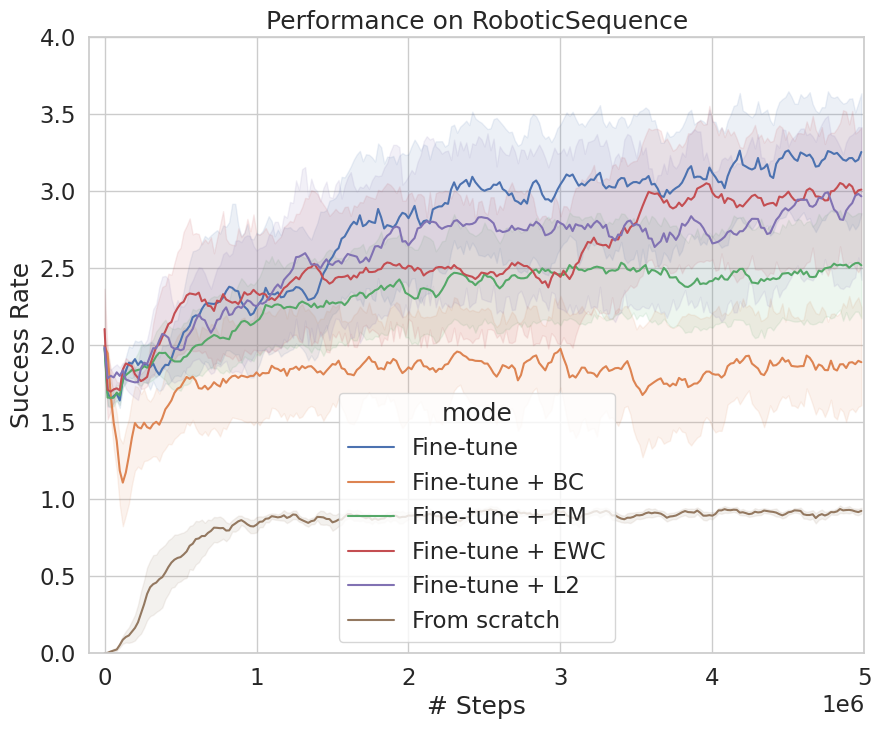}}
    \end{subfigure}
    \begin{subfigure}[b]{0.49\textwidth}
        \centering
        \mute{\includegraphics[width=\textwidth]{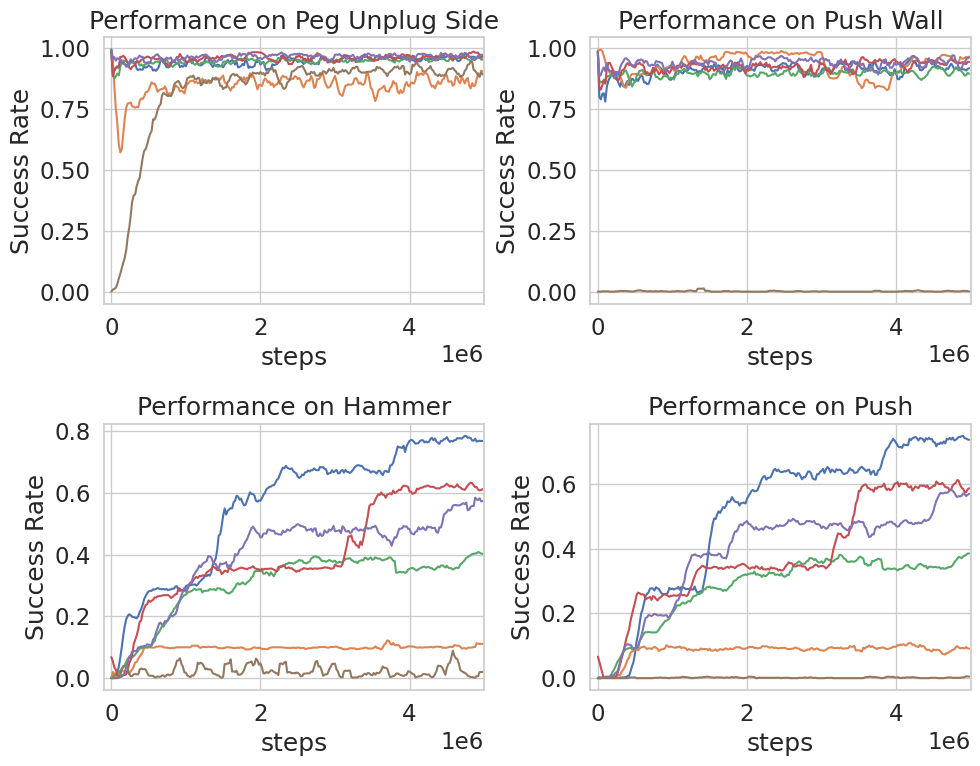}}
    \end{subfigure}

    \caption{The performance on a robotic sequence where the known tasks are positioned at the beginning.}
    \label{fig:stitchedenv_start_seq}
\end{figure}

Finally, we verify that the gap between vanilla fine-tuning and knowledge retention methods does not appear when the relevant skills are only needed at the start of the downstream task. To do this, we use the following sequence of goals: \texttt{peg-unplug-side, push-wall, hammer, push} with a model pre-trained on \texttt{peg-unplug-side, push-wall}. Results in Figure~\ref{fig:stitchedenv_start_seq} show that indeed in this scenario there is no forgetting and fine-tuning manages just as well or sometimes even slightly better than knowledge retention methods. 

\paragraph{Impact of the memory size on the results}

\begin{figure}[H]
    \centering
    \begin{subfigure}[b]{0.48\textwidth}
        \centering
        \mute{\includegraphics[width=\textwidth]{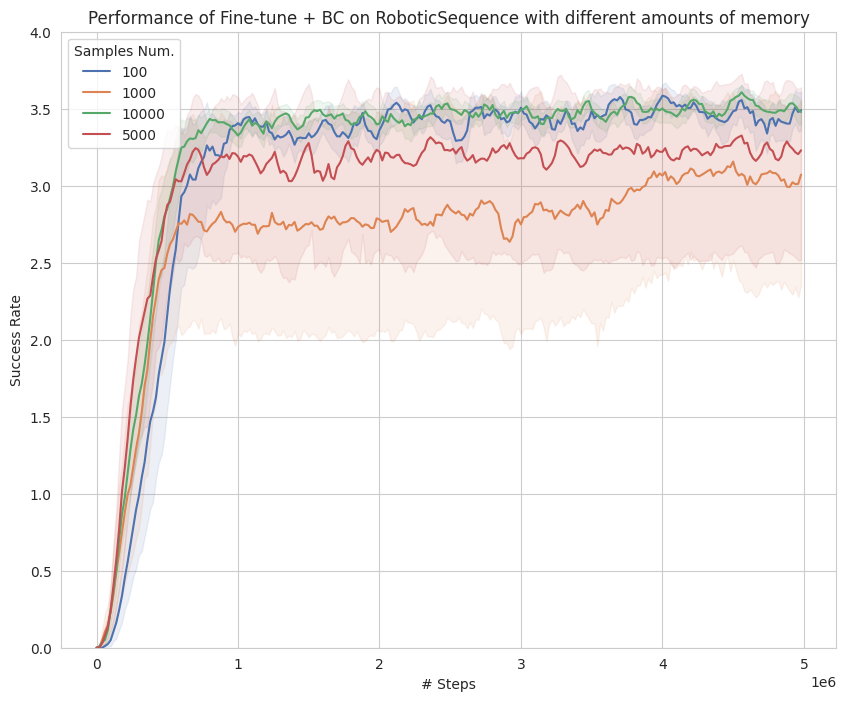}}
    \end{subfigure}
    \begin{subfigure}[b]{0.48\textwidth}
        \centering
        \mute{\includegraphics[width=\textwidth]{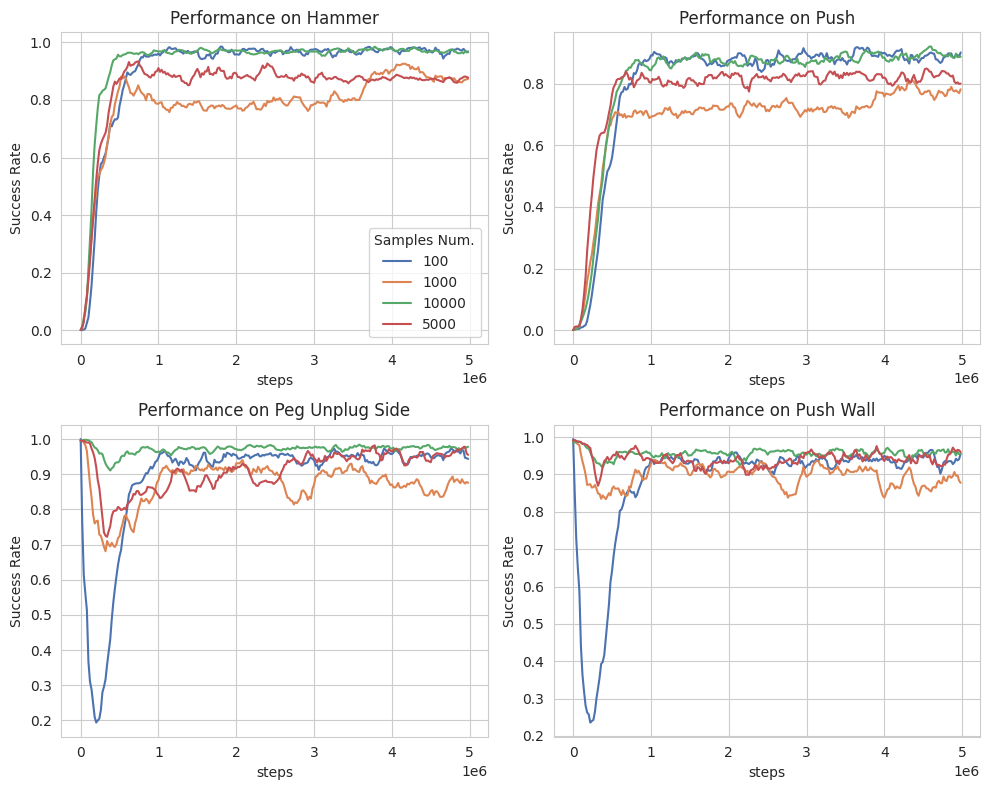}}
    \end{subfigure}

    \caption{\diff{The performance of Fine-tune + BC with different memory sizes. Even with $100$ samples we are able to retain the knowledge required to make progress in the training.}
    }
    \label{fig:memory_size}
\end{figure}

The memory overhead is an important consideration in fine-tuning with a behavioral cloning loss. We run experiments to check how many samples we actually need to protect knowledge of the previous tasks. Results presented in Figure~\ref{fig:memory_size} show that even with $100$ samples we are able to keep good performance, at the cost of a higher performance drop on the pre-trained tasks at the beginning of the fine-tuning process.

\begin{table}[H]
\centering
\caption{Forward transfer on the pre-trained tasks depending on the number of prefix tasks in \roboticsequence{}.}
\label{tab:prefix_len_results_appendix}

\begin{tabular}{crrrrrr}

\toprule
\multirow{ 2}{*}{\makecell{Prefix \\ Len}}  & \multicolumn{3}{c}{\texttt{push-wall}} & \multicolumn{3}{c}{\texttt{peg-unplug-side}}\\
\cmidrule(lr){2-4}
\cmidrule(lr){5-7}
&          FT &   EWC &    BC &          FT &   EWC &    BC \\
\midrule
1                 &                    0.18 \tiny{[-0.19, 0.43}] &  0.88 \tiny{[0.84, 0.91}] &  0.93 \tiny{[0.89, 0.96}] &               0.28 \tiny{[0.01, 0.46}] &  0.77 \tiny{[0.58, 0.88}] &  0.92 \tiny{[0.88, 0.94}] \\
2                 &                    0.17 \tiny{[-0.21, 0.44}] &  0.65 \tiny{[0.44, 0.82}] &  0.97 \tiny{[0.97, 0.98}] &              0.15 \tiny{[-0.08, 0.35}] &  0.55 \tiny{[0.37, 0.70}] &  0.95 \tiny{[0.94, 0.96}] \\
3                 &                    0.10 \tiny{[-0.03, 0.23}] &  0.64 \tiny{[0.50, 0.75}] &  0.98 \tiny{[0.98, 0.98}] &               0.03 \tiny{[0.00, 0.06}] &  0.41 \tiny{[0.28, 0.54}] &  0.95 \tiny{[0.95, 0.95}] \\
4                 &                   -0.00 \tiny{[-0.16, 0.10}] &  0.62 \tiny{[0.48, 0.75}] &  0.97 \tiny{[0.97, 0.98}] &              0.03 \tiny{[-0.00, 0.08}] &  0.46 \tiny{[0.33, 0.59}] &  0.94 \tiny{[0.94, 0.95}] \\
\bottomrule
\end{tabular}
\end{table}

\begin{figure}[H]
\caption{Training performance for different architecture choices.}
\label{fig:full_arch_results}
    \centering

    \begin{subfigure}[b]{0.85\textwidth}
        \centering
        \mute{\includegraphics[width=\textwidth]{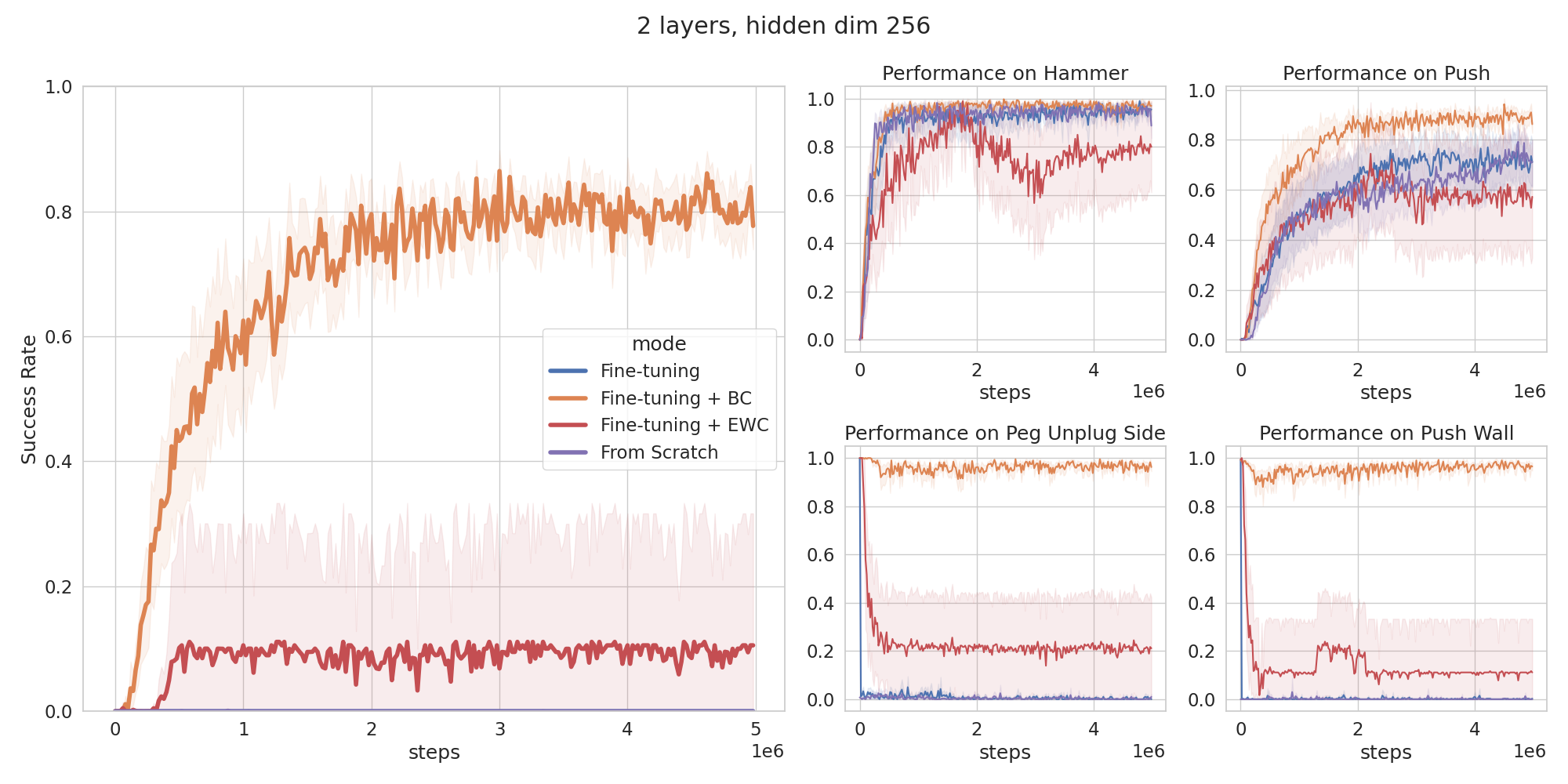}}
    \end{subfigure}
    \begin{subfigure}[b]{0.85\textwidth}
        \centering
        \mute{\includegraphics[width=\textwidth]{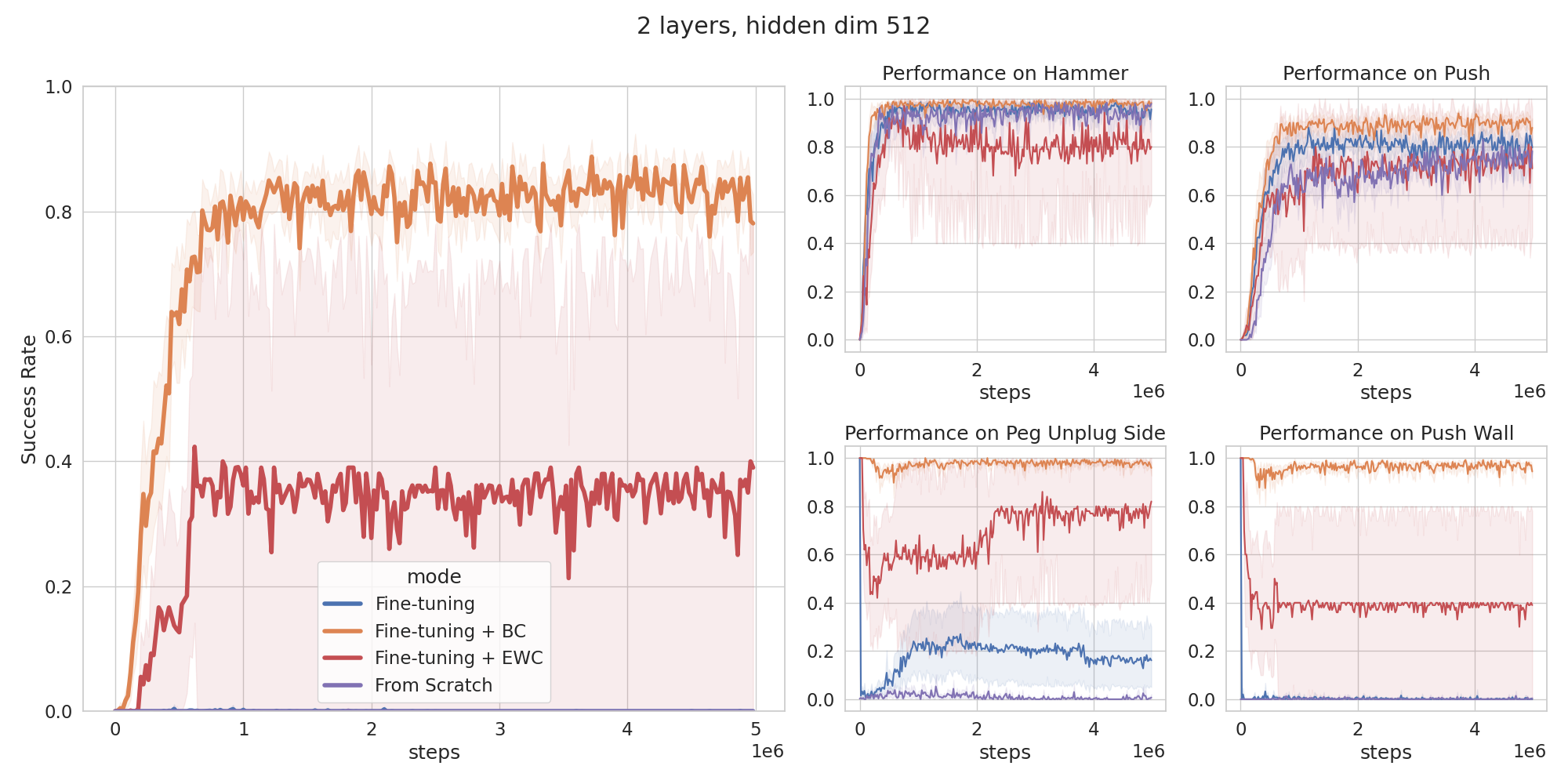}}
    \end{subfigure}
    \begin{subfigure}[b]{0.85\textwidth}
        \centering
        \mute{\includegraphics[width=\textwidth]{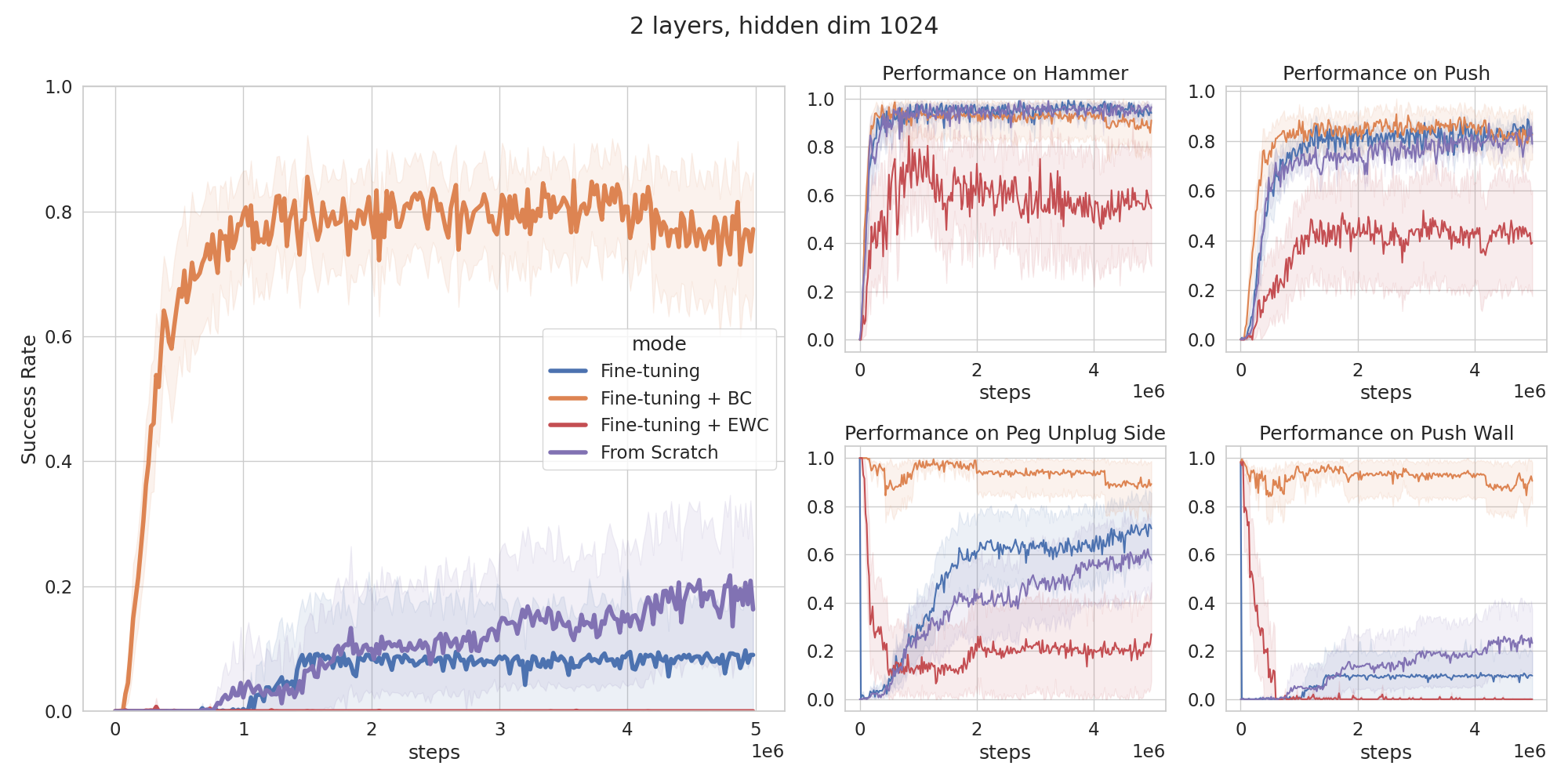}}
    \end{subfigure}

\end{figure}%
\begin{figure}[H]\ContinuedFloat
    \centering

    
    \begin{subfigure}[b]{0.85\textwidth}
        \centering
        \mute{\includegraphics[width=\textwidth]{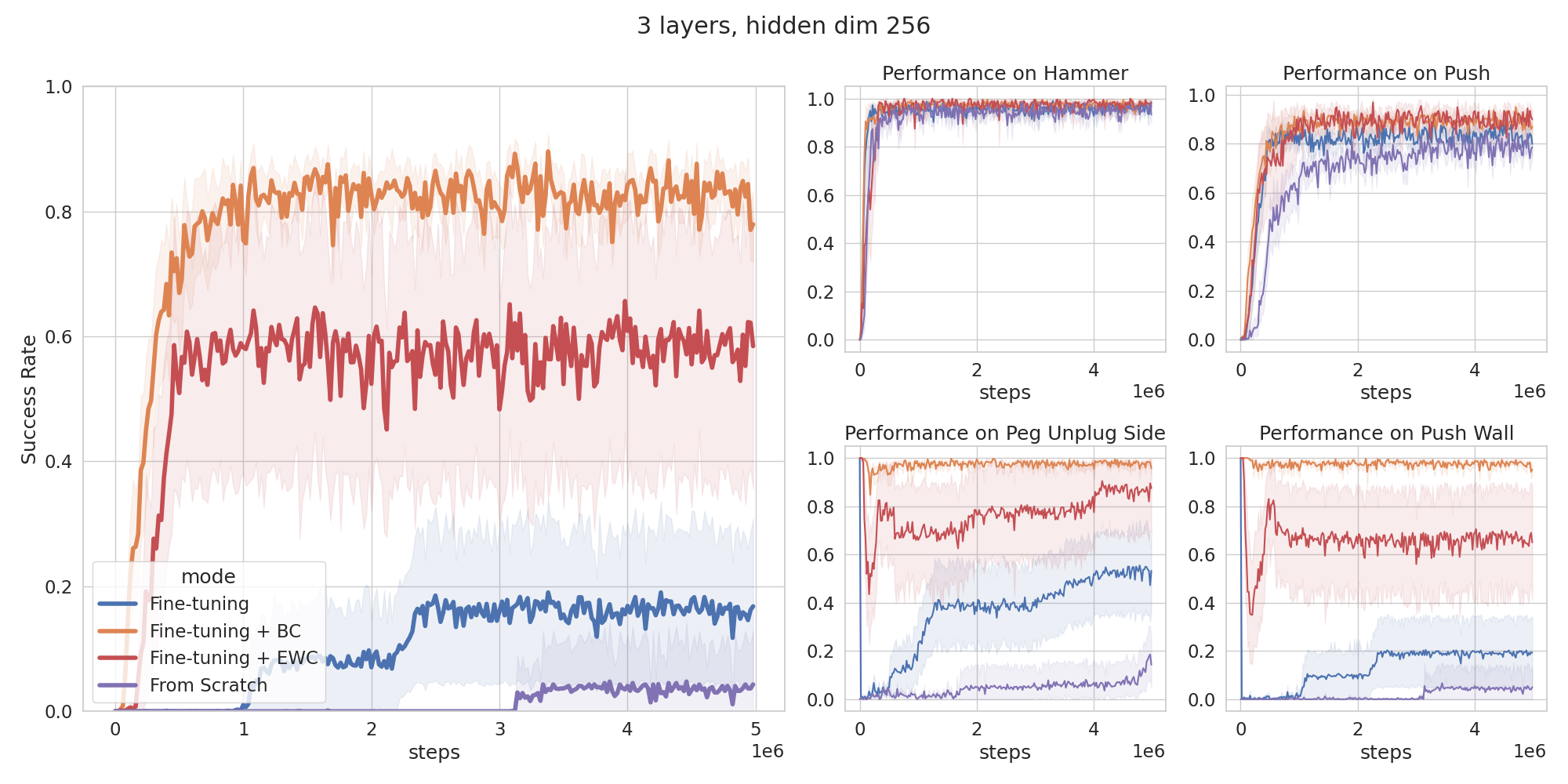}}
    \end{subfigure}
    \begin{subfigure}[b]{0.85\textwidth}
        \centering
        \mute{\includegraphics[width=\textwidth]{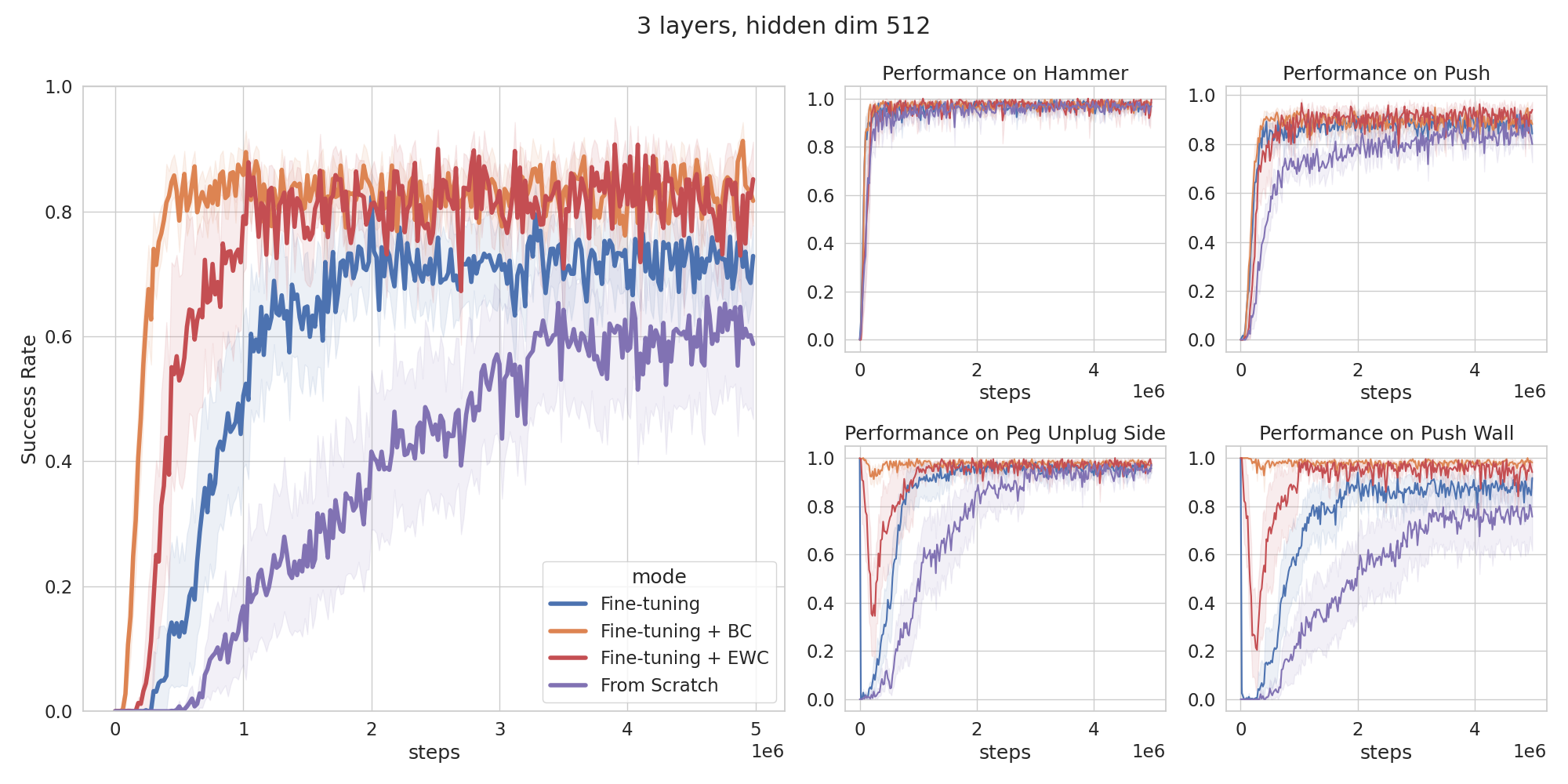}}
    \end{subfigure}
    \begin{subfigure}[b]{0.85\textwidth}
        \centering
        \mute{\includegraphics[width=\textwidth]{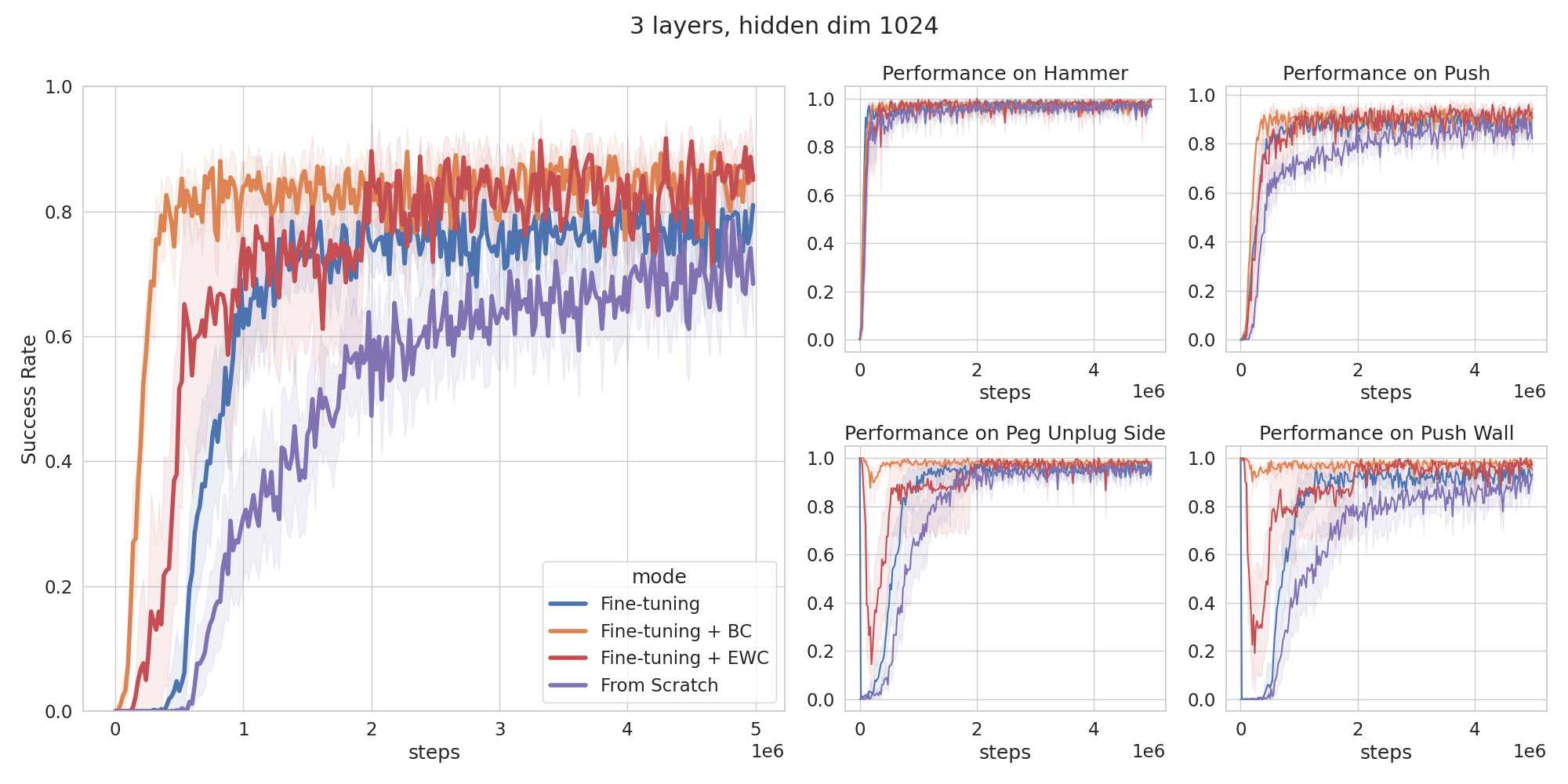}}
    \end{subfigure}
\end{figure}%
\begin{figure}[H]\ContinuedFloat
    \centering

    
    \begin{subfigure}[b]{0.85\textwidth}
        \centering
        \mute{\includegraphics[width=\textwidth]{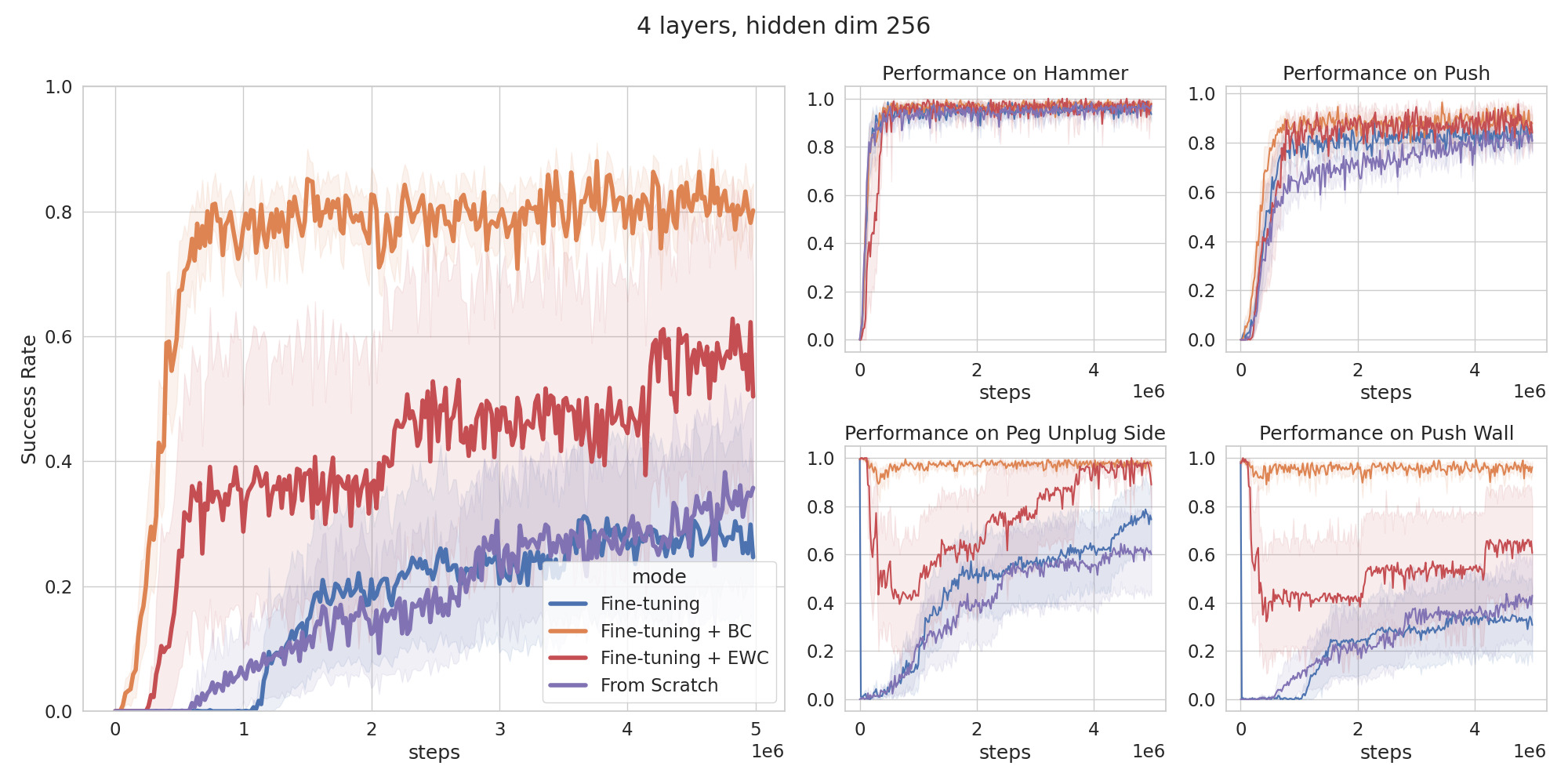}}
    \end{subfigure}
    \begin{subfigure}[b]{0.85\textwidth}
        \centering
        \mute{\includegraphics[width=\textwidth]{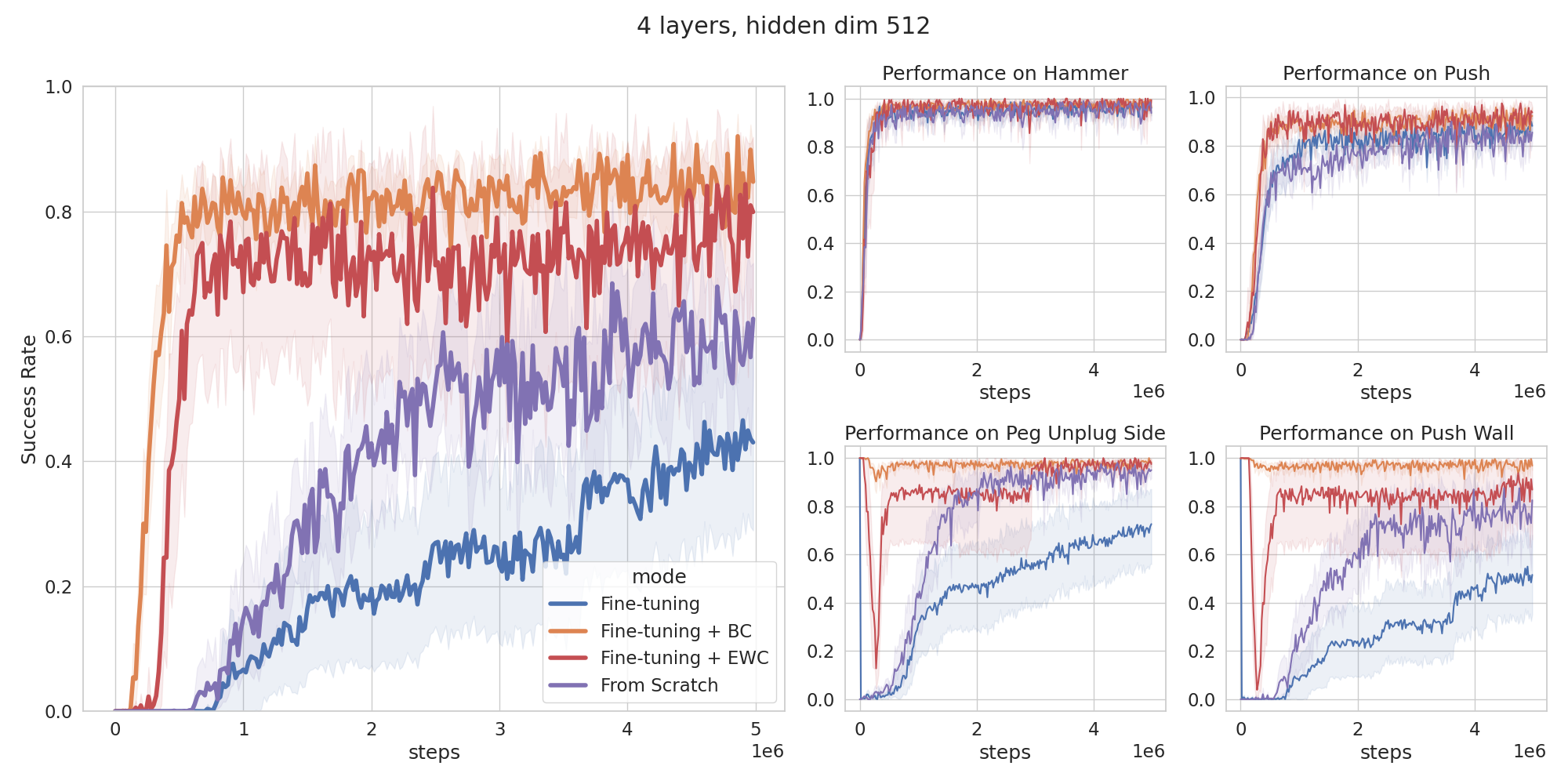}}
    \end{subfigure}
    \begin{subfigure}[b]{0.85\textwidth}
        \centering
        \mute{\includegraphics[width=\textwidth]{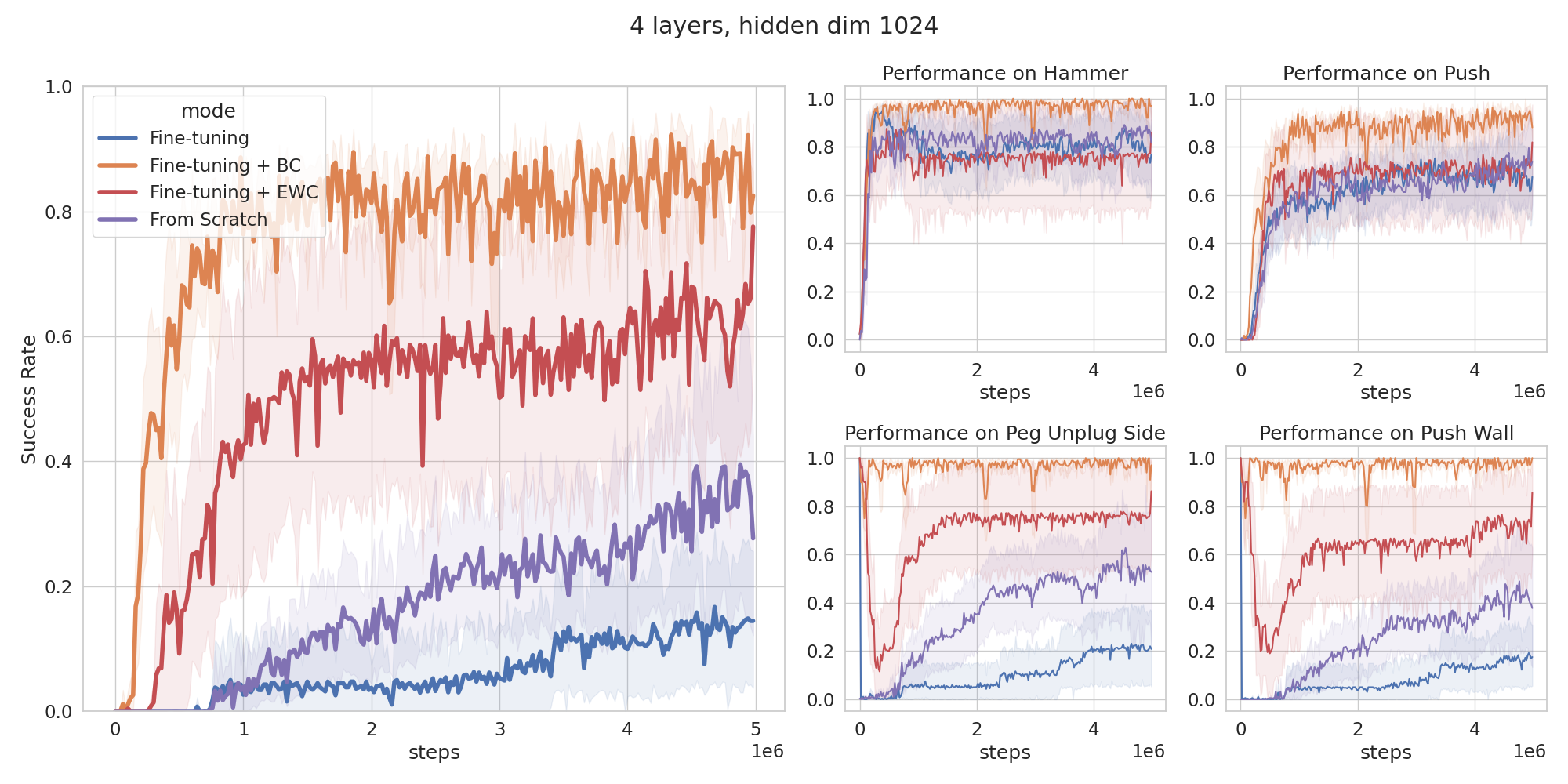}}
    \end{subfigure}

\end{figure}


\end{document}